\documentclass{article}

 \usepackage[preprint]{neurips_2026}

\usepackage[utf8]{inputenc} 
\usepackage[T1]{fontenc}    
\usepackage{hyperref}       
\usepackage{url}            
\usepackage{booktabs}       
\usepackage{amsfonts}       
\usepackage{nicefrac}       
\usepackage{microtype}      
\usepackage{xcolor}         
\usepackage{microtype}
\usepackage{graphicx}
\usepackage{subcaption}
\usepackage{booktabs} 
\usepackage{tabularx}
\usepackage{hyperref}
\usepackage{multicol, multirow}
\usepackage{titletoc}

\usepackage{amsmath}
\usepackage{amssymb}
\usepackage{mathtools}
\usepackage{amsthm}
\usepackage{subcaption}
\usepackage{graphicx}
\usepackage{longtable}
\usepackage{enumitem}
\usepackage{tcolorbox}
\newsavebox{\codebox}

\definecolor{rosso}{RGB}{220,57,18}
\definecolor{giallo}{RGB}{255,153,0}
\definecolor{blu}{RGB}{102,140,217}
\definecolor{verde}{RGB}{16,150,24}
\definecolor{viola}{RGB}{153,0,153}
\definecolor{yellow}{RGB}{255,255,168}
\definecolor{lightblue}{RGB}{0, 255, 255}
\definecolor{green2}{RGB}{64, 224, 208}
\definecolor{orange}{RGB}{255, 195, 0}
\definecolor{grey}{RGB}{220,220,220}
\definecolor{lightgreen}{RGB}{144,238,144}
\definecolor{pink}{RGB}{255, 192, 203}
\definecolor{red}{RGB}{255,71,77}
\definecolor{ytred}{HTML}{FF0000}
\definecolor{fbblue}{HTML}{1877F2}
\definecolor{tiktokblack}{HTML}{010101}
\usepackage{xcolor}

\definecolor{policeblue}{RGB}{0, 0, 205}      
\definecolor{civilianred}{RGB}{255, 69, 0}    
\definecolor{dispatchgray}{RGB}{255, 20, 147} 
\definecolor{othergreen}{RGB}{34, 139, 34}    

\usepackage[capitalize,noabbrev]{cleveref}

\theoremstyle{plain}

\theoremstyle{definition}

\theoremstyle{remark}

\newcommand{\tool}{\textbf{DeEscalWild}~}
\newcommand{\toolnospace}{\textbf{DeEscalWild}}

\usepackage[textsize=tiny]{todonotes}

\title{\textbf{\toolnospace: A Real-World Benchmark for Automated De-Escalation Training with SLMs}}

\author{%
  Md Hasebul Hasan$^{1}$\thanks{Corresponding author: \texttt{mdhasebul.hasan@uta.edu}} \quad
  Krity Haque Charu$^{1}$ \quad
  Eshwara Prasad Sridhar$^{2}$ \\
  \textbf{Shuchisnigdha Deb}$^{2}$ \quad
  \textbf{Mohammad A. Islam}$^{1}$ \\
  \\
  $^{1}$Department of Computer Science and Engineering \\
  $^{2}$Department of Industrial, Manufacturing, and Systems Engineering \\
  University of Texas at Arlington, USA
}

\begin{document}

\maketitle

\begin{abstract}
Effective de-escalation is critical for law enforcement safety and community trust, yet traditional training methods lack scalability and realism. While Large Language Models (LLMs) enable dynamic, open-ended simulations, their substantial computational footprint renders them impractical for deployment on the lightweight, portable hardware required for immersive field training. Small Language Models (SLMs) offer a viable real-time alternative but suffer from a critical scarcity of high-quality, domain-specific training data. To bridge this gap, we present \toolnospace, a novel benchmark dataset curated from a multi-stage pipeline of ``in-the-wild'' police-civilian interactions extracted from publicly available video repositories. Starting with 5,000 raw inputs, we employed a rigorous hybrid filtering process combining human-in-the-loop verification with ``LLM-as-a-Judge'' evaluation to distill 1,500 high-fidelity scenarios. The resulting corpus comprises 285,887 dialogue turns, totaling approximately 4.7 million tokens. Extensive experiments demonstrate that SLMs fine-tuned on this data significantly outperform their base counterparts across ROUGE-L, BLEU-4, METEOR, BERTScore, Realism Score, and human evaluation metrics. Notably, our fine-tuned Qwen~2.5 (3B-Instruct) surpasses the general-purpose Gemini~2.5 Flash model when evaluated under equivalent conditions, demonstrating that domain-optimized SLMs can achieve superior performance with a fraction of the computational cost. This work establishes the foundational infrastructure for accessible, low-latency, and privacy-preserving officer training systems at the edge. We publicly release our \href{https://github.com/Hasebul/DeEscalWild-Benchmark-Framework}{code} and \href{https://doi.org/10.7910/DVN/CWMCZI}{dataset}.
\end{abstract}
\section{Introduction}

Effective de-escalation is a cornerstone of modern policing, directly influencing officer safety, subject welfare, and public trust. While the ability to navigate volatile encounters is a critical skill, traditional training paradigms, including static role-playing and branching video scenarios, suffer from inherent limitations in scalability, consistency, and realism.

Recent advances in Large Language Models (LLMs) offer a promising avenue for creating dynamic, open-ended training simulations. However, the computational footprint of state-of-the-art LLMs renders them impractical for deployment on the lightweight, portable hardware required for immersive field training, such as standalone VR headsets or mobile edge devices. To achieve real-time latency without tethered compute, the field must pivot toward Small Language Models (SLMs). Yet, while SLMs offer inference efficiency, they lack the broad reasoning capabilities of their larger counterparts and require extensive fine-tuning to perform reliably in high-stakes contexts. This creates a critical gap: the need for specialized SLMs for de-escalation is urgent, but high-quality, domain-specific training data for this task is virtually non-existent.

\begin{figure}[t]
    \centering
    \includegraphics[width=\textwidth]{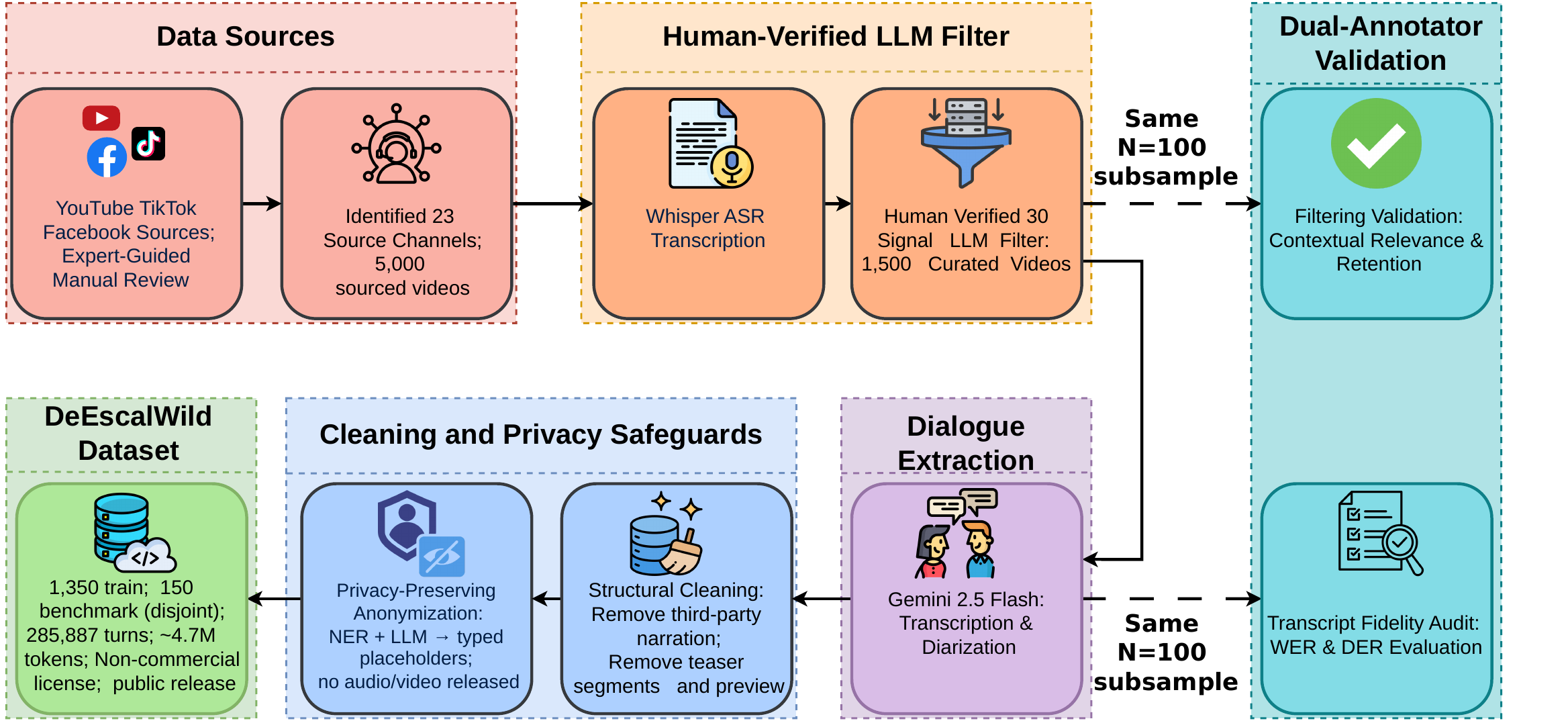}
    \caption{\textbf{DeEscalWild end-to-end data curation pipeline.} Starting from 5,000 videos across 23 social media channels, a human-verified 30-signal LLM filter retains 1,500 police-civilian interactions satisfying three validity conditions: no off-domain noise, confirmed police presence, and sufficient escalation depth. Gemini 2.5 Flash performs native-audio transcription and speaker diarization. Transcripts undergo structural cleaning and privacy-preserving anonymization via NER and LLM-based parsing; no raw audio or video is released. The final corpus comprises 1,350 training interactions and a strictly disjoint 150-scenario benchmark totalling 285,887 dialogue turns and approximately 4.7 million tokens, released under a non-commercial research license. Solid arrows indicate the main curation pipeline; dashed arrows indicate the dual-annotator validation branch applied to the same N=100 subsample at two checkpoints: filtering precision and transcript fidelity.}
    \label{fig:pipeline}
\end{figure}
Current computational approaches have yet to fully resolve these constraints. Exploratory studies such as \cite{ANAND2024EXP} assess off-the-shelf models like ChatGPT, finding that while general-purpose LLMs can simulate basic empathetic exchanges, they remain constrained by the latency and connectivity requirements of API-based architectures. Furthermore, reliance on closed-source commercial models precludes the domain-specific fine-tuning necessary to capture the nuance of tactical communication. Similarly, prototypes such as the \textit{Adaptive De-escalation Trainer} \cite{sridhar2025adaptive} demonstrate semantic capability but fail to meet operational constraints. With reported latencies exceeding 4 seconds and a reliance on server-grade compute, such systems are ill-suited for the split-second decision-making required in the field.

The challenge is further compounded by a lack of suitable training data. While the application of NLP to law enforcement is not without precedent, \cite{voigt2017language} for instance utilized body-camera footage to analyze racial disparities in officer language, yet such works treat police dialogue purely as archival evidence for post-hoc sociological analysis rather than as a substrate for active generative training. A critical gap remains: the field lacks a standardized, high-volume corpus focused on \textit{de-escalation scenarios}, namely the specific verbal strategies used to resolve conflict.

To bridge this gap, we introduce \toolnospace, the first large-scale dataset derived from ``in-the-wild'' police-civilian interactions. Unlike synthetic or crowdsourced datasets, which often lack emotional fidelity, we constructed our corpus from publicly available video repositories, including YouTube, TikTok, and Facebook, to capture the raw, unstructured nature of real-world conflict. With \toolnospace, we aim to democratize access to critical de-escalation training, ultimately seeking to reduce violent outcomes in police-civilian interactions. To safeguard privacy and prevent misuse, our public release is strictly limited to fully anonymized textual transcripts, and the use of derived models is explicitly restricted to controlled educational simulations.

We make the following three contributions: 
\begin{enumerate}[leftmargin=*, topsep=0pt, noitemsep] 

\item \textbf{The \tool dataset and benchmark.} We introduce \tool, a large-scale dataset and benchmark for modeling de-escalation in real-world interactions. Starting from 5,000 raw videos, we develop a hybrid curation pipeline that combines human-in-the-loop verification with LLM-based filtering to distill 1,500 high-quality scenarios, comprising 285,887 dialogue turns and approximately 4.7 million tokens. To ensure privacy and safety, the dataset is released exclusively as fully anonymized textual transcripts. Building on this corpus, we define a standardized benchmark by constructing a held-out test set of 150 carefully curated interactions, each paired with structured context and civilian character profiles to enable controlled, reproducible evaluation. The benchmark adopts an interactive simulation protocol and integrates both automatic metrics and LLM-as-a-judge~\cite{zheng2023judging} evaluation to assess linguistic fidelity, behavioral realism.

 \item \textbf{SLM efficacy at the edge.} We demonstrate that domain-specific fine-tuning allows compact models to achieve performance comparable to significantly larger models. Our experiments show that a fine-tuned Qwen~2.5 (3B) significantly outperforms a general-purpose LLM baseline across BLEU-4, ROUGE-L, METEOR, BERTScore, Realism Score, and human evaluation metrics, demonstrating that data quality is a viable substitute for parameter scale in specialized tasks. 
 
 \item \textbf{A scalable plug-and-play data curation framework.} We introduce a platform-agnostic pipeline for converting publicly available, in-the-wild videos into clean, diarized textual transcripts. Although instantiated for the \tool domain, the framework is general-purpose and can be readily adapted to other domains through lightweight modifications to the filtering pipeline. \end{enumerate}

\section{Related Work}
\label{sec:related_work}

\noindent \textbf{Generative simulation and role consistency.} Traditional rule-based simulators are rigid and susceptible to gaming. While LLMs enable open-ended generation~\cite{gao2024large}, crisis simulation requires sustained role consistency under pressure, addressed via memory retrieval~\cite{park2023generative} and persona-conditioned alignment~\cite{wang2025coser, wang2024rolellm}. \citet{violakis2025leveraging} and \citet{sridhar2025adaptive} further demonstrate that effective trainers must modulate emotional tone in real-time, motivating dynamic, non-scripted architectures.

\noindent \textbf{Efficient deployment via SLMs.} LLM computational demands preclude deployment on portable hardware. \citet{pecher-etal-2025-comparing} show that specialized SLMs can outperform general LLMs with as few as 100 labeled samples, and \citet{xu-etal-2024-small} demonstrate that fine-tuned SLMs serve as effective plug-ins for larger frameworks. Our work applies these insights to edge deployment for real-time de-escalation simulation.

\noindent \textbf{Data scarcity in high-stakes domains.} Existing goal-oriented datasets target agreement-based tasks such as negotiation~\cite{zhan2024let} and lack threat-assessment protocols. \textit{EmpatheticDialogues} offers emotional breadth but not tactical specificity. Although body-worn camera footage contains the necessary escalation signals~\cite{srbinovska2025towards}, privacy constraints have hindered public benchmark construction~\cite{rosas2025constructing}. \tool is, to our knowledge, the first large-scale benchmark curated from in-the-wild footage with the ecological validity required for robust tactical agent training.
\section{The \textbf{\tool} Dataset}
\subsection{Design Principles}
\label{sec:design_principles}

The construction of \tool is guided by five core principles addressing the unique challenges of training agents for high-stakes, socially sensitive de-escalation.

\noindent \textbf{Reasoning-centric de-escalation.} Unlike standard chitchat or task-oriented dialogue systems, de-escalation requires deep strategic reasoning. \tool prioritizes interactions that require the model to infer latent mental states, anticipate escalation triggers, and select communicative actions that actively lower tension rather than simply maintaining conversational flow.

\noindent \textbf{Ecological validity.} \tool is derived exclusively from real-world, in-the-wild law enforcement recordings, capturing speech disfluencies, emotional outbursts, and non-cooperative behavior rarely present in crowdsourced or synthetic datasets. This ensures the data reflects true distribution shifts encountered in deployment scenarios involving victims, suspects, and civilians.

\noindent \textbf{Sociodemographic and situational diversity.} To mitigate algorithmic bias, \tool maximizes diversity across participant demographics, spanning ages, ethnicities, and regional dialects, and incident types ranging from routine traffic stops to mental health crises. Detailed demographic breakdowns are provided in Appendix~\ref{app:demography}.

\noindent \textbf{Temporal depth and long-horizon context.} De-escalation is a gradual process. \tool focuses on long-context conversations averaging 18 minutes, facilitating training of models capable of long-horizon state tracking, recognizing slowly developing emotional shifts, and executing multi-step persuasion strategies.

\noindent \textbf{Multi-agent and environmental complexity.} \tool includes complex multi-party scenarios beyond simple dyadic interactions, featuring incidents involving multiple officers and civilians in public spaces. This requires models to handle dynamic turn-taking and cross-talk management typical of chaotic real-world scenes. 

\subsection{Tasks} 
\noindent \textbf{Contextual civilian response generation.} The primary task is Civilian Response Generation in a high-stakes setting. Given a dialogue history $H$ and the current officer utterance, the model must generate a plausible civilian response $R$ that reflects the semantic and emotional trajectory of the interaction, ranging from compliance to aggression, conditioned on the officer's de-escalation strategy.

\noindent \textbf{De-escalation strategy alignment.} Models are additionally evaluated on interactional realism and contextual consistency via two sub-tasks: (i)~\textbf{Behavioral realism}, requiring emotionally authentic responses that capture the full spectrum of civilian reactions, including escalation, rather than artificially constrained polite outputs; and (ii)~\textbf{Turn-level consistency}, requiring coherent long-horizon reasoning that correctly recalls and applies prior contextual details such as the subject's identity or stated concerns.

\subsection{Dataset Curation, Validation, and Benchmark Construction} \label{sec:data_pipeline} To construct a dataset for effective de-escalation training, we curate real-world, naturalistic videos depicting police-civilian interactions during potential conflicts. The dataset is designed to capture a diverse range of contexts by including instances of both escalation and de-escalation across varied incident types, demographics, and severity levels.

\noindent \textbf{Curation pipeline.} As illustrated in Figure~\ref{fig:pipeline}, the data curation workflow follows a multi-stage pipeline. We construct \tool from approximately 5,000 publicly available videos collected from 23 social media sources spanning YouTube, TikTok, and Facebook (listed in full in Appendix~\ref{app:data_sources}). Each candidate video is first transcribed with Whisper~\cite{pmlr-v202-radford23a} and then passed through an LLM-guided filtering pipeline with human oversight. The filter maps each transcript to a structured schema of 30 binary signals spanning police presence, conversational structure, escalation, de-escalation, and off-domain noise, and retains only videos satisfying deterministic criteria for contextual validity, law-enforcement relevance, and interaction intensity. This procedure removes advertisements, commentary-driven content, and low-information footage, yielding a curated set of 1,500 high-value police-civilian interactions. The retained videos are then processed with Gemini~2.5 Flash~\cite{comanici2025gemini} for joint verbatim transcription and speaker diarization, producing time-aligned, speaker-attributed transcripts with inferred speaker roles. Full feature definitions, prompts, and filtering rules are provided in Appendix~\ref{app:filtering_pipeline} and Appendix~\ref{app:diarization}.

\noindent \textbf{Cleaning and privacy safeguards.} Because real-world social media footage contains substantial structural contamination, we further sanitize the extracted transcripts prior to release. Specifically, we remove non-chronological teaser segments and preview clips that would otherwise disrupt temporal coherence, and filter out third-party narration and post-hoc commentary so that the final corpus preserves only on-scene police-civilian interactions. We additionally apply a hybrid de-identification pipeline combining Named Entity Recognition (NER) with LLM-based contextual parsing to replace names, locations, and other sensitive information with semantically typed placeholders such as \texttt{[CIVILIAN\_NAME]} and \texttt{[LOCATION]}. The released resource contains only anonymized diarized text transcripts; raw audio and video are not distributed. Further details are provided in Appendix~\ref{app:data_cleaning} and Appendix~\ref{app:anonymization}.

\noindent \textbf{Human validation and in-the-wild challenges.} We audit both retrieval precision and transcript fidelity through a dual-annotator review of a randomly sampled subset of $N = 100$ videos. Each video averaged approximately 18 minutes in duration, yielding more than 30 hours of source material in total; careful review required substantially more than real-time playback, representing a significant investment of expert annotation effort. The filtering stage exhibits high precision: both annotators achieved 100\% raw agreement on contextual validity and police-civilian role relevance, and 90.9\% raw agreement on both interaction intensity and final retention decisions. On the same subset, we evaluate transcript quality using corpus-level Word Error Rate (WER) and Diarization Error Rate (DER). Global WER ranges from 0.77\% to 0.91\% across the two annotators, while global DER ranges from 6.26\% to 8.16\%, indicating strong transcription fidelity alongside a non-trivial diarization challenge inherent to realistic multi-party settings. Residual errors are attributable to four recurring failure modes: (i) severe acoustic degradation from wind, sirens, radio traffic, and motion artifacts; (ii) overlapping speech and irregular turn-taking in high-tension exchanges; (iii) speaker confusion and temporal drift in multi-party scenes; and (iv) long-context degradation on extended recordings. These findings confirm that \tool represents a substantially more challenging setting than curated conversational benchmarks, while remaining sufficiently faithful for downstream SLM training and evaluation. Complete agreement statistics, metric definitions, and failure mode analyses are provided in Appendix~\ref{app:human_verification} and Appendix~\ref{app:diarization}.

\noindent \textbf{Final benchmark.} From the sanitized corpus, we reserve a held-out benchmark of $N{=}150$ high-quality interactions, strictly isolated from all training and validation data and disjoint from the $N{=}100$ pipeline verification subset, ensuring zero overlap between quality-control and evaluation scenarios. Each example is paired with a structured situational context and a civilian character profile describing behavioral state, motivations, and initial tension level. Evaluation follows an autoregressive simulation protocol: at each turn, the model receives the ground-truth officer utterance and generates the corresponding civilian response, requiring sustained persona adherence across the full interaction. Performance is assessed via automatic metrics (ROUGE-L, BLEU-4, METEOR, BERTScore) and an external LLM judge scoring realism and de-escalation quality. Although the benchmark contains $N{=}150$ scenarios, each averages 18 minutes and ${\sim}190$ turns, yielding ${\sim}24{,}000$ turn-level generation decisions in total, a substantially more demanding evaluation regime than scenario count alone implies. Overall, \tool provides 1,500 anonymized interactions and a realistic benchmark for assessing persona adherence, domain-specific reasoning, and de-escalation behavior under naturalistic conversational pressure. Full protocol details are in Appendix~\ref{app:benchmark_construction}.

\noindent \textbf{Dataset statistics.} The final \tool dataset constitutes a substantial corpus of domain-specific tactical dialogue. As summarized in Table~\ref{tab:stats}, the benchmark comprises 1,500 verified scenarios encompassing a total of 285,887 dialogue turns, approximately 3.6 million words, and an estimated 4.7 million tokens. This data volume provides the density required to robustly fine-tune SLMs without overfitting. Three qualitative examples illustrating the linguistic depth and diversity of \tool are presented in Appendix~\ref{appendix:qualitative_examples}.

\begin{table}[h]
    \centering
    \small
    \begin{tabular}{lr}
        \toprule
        \textbf{Metric} & \textbf{Count} \\
        \midrule
        Total Scenarios & 1,500 \\
        Total Dialogue Turns & 285,887 \\
        Total Word Count & 3,626,196 \\
        Total Token Count (Est.) & 4,714,067 \\
        \bottomrule
    \end{tabular}
    \vspace{5pt}
    \caption{Summary statistics of the \tool dataset.}
    \label{tab:stats}
\end{table}

\noindent \textbf{Challenges and optimization.} Platform rate-limiting during metadata extraction from YouTube, TikTok, and Facebook was addressed through request throttling, randomized inter-query delays, and distributed collection sessions, enabling retrieval of metadata for all 5,000 candidate videos within platform access guidelines. For diarization, initial experiments with \texttt{pyannote}~\cite{bredin2023pyannote} produced unreliable speaker attribution, prompting a Whisper\,+\,pyannote\,+\,LLM refinement approach. \texttt{Qwen2.5-7B-Instruct} was evaluated as the refinement model but proved unsuitable: its 8,192-token output limit cannot accommodate the structured transcripts required for 18-minute videos, which routinely exceed 20,000 tokens. Larger open-source alternatives were ruled out on computational grounds, and the three-stage pipeline was prohibitively slow per video. We consequently adopted Gemini~2.5 Flash~\cite{comanici2025gemini}, which processes raw audio natively without an intermediate ASR step, produces very small diarization errors, and processed the full 1,500-video corpus for approximately \$120\,USD. Its reliability and cost-effectiveness motivated its subsequent use as the generalist LLM baseline in our evaluations.
\section{Experiments}
\label{sec:experiments}
In this section, we present the evaluation details of the \tool dataset's utility, with a specific focus on its ability to train agents that accurately reflect realistic civilian dynamics in de-escalation scenarios. We assess the impact of domain-specific fine-tuning across a diverse set of state-of-the-art open-weights SLMs.

\subsection{Experimental Protocol and Setup}
\label{subsec:exp_setup}
\noindent \textbf{Evaluation strategy.} To quantify the effectiveness of the \tool dataset, we employ a pre-post fine-tuning evaluation protocol. We measure the performance of base instruct-tuned models in a zero-shot configuration against their fine-tuned counterparts on a held-out benchmark set. This comparative analysis isolates the specific value added by \tool in teaching de-escalation strategies, reasoning, and policy adherence. We additionally benchmark the fine-tuned SLMs against Gemini~2.5 Flash, a strong general-purpose LLM baseline, using a few-shot prompting strategy to evaluate whether domain-specific fine-tuning enables compact models to outperform prompt-engineered generalist systems.

\noindent \textbf{Data splitting.} The \tool dataset comprises 1,350 interactions for model development and 150 interactions in a held-out benchmark set. During training, we reserve 3\% of the 1,350 development interactions as a validation split and use the remaining 97\% for training. All models are trained using three independent random seeds; each resulting checkpoint is evaluated independently on the held-out benchmark, and we report the mean and standard deviation across the three runs. Critically, the split is defined at the scenario level to enforce strict data isolation, ensuring that no dialogue turns from benchmark interactions appear in either the training or validation data.

\noindent \textbf{Implementation details.} All experiments were conducted on a single NVIDIA GeForce RTX 3090 GPU (24\,GB VRAM). To ensure computational efficiency, we employed 4-bit Quantized Low-Rank Adaptation (QLoRA)~\cite{dettmers2023qlora}. The LoRA adapters were configured with rank $r = 16$, scaling factor $\alpha = 32$, and learning rate $2 \times 10^{-4}$. Models were fine-tuned for 3 epochs with a global batch size of 4 using the AdamW optimizer~\cite{loshchilov2017decoupled}.

\noindent \textbf{Evaluation metrics.} We report five complementary metrics spanning lexical overlap, semantic fidelity, and behavioral plausibility. For readability, ROUGE-L, BLEU-4, and METEOR are reported on a 0--100 scale; BERTScore is reported in its standard 0--1 range; and the Realism Score is reported on a 0--100 scale.

\begin{enumerate}[leftmargin=*, topsep=2pt, noitemsep] \item \textbf{ROUGE-L~\cite{lin-2004-rouge}:} Measures longest-common-subsequence overlap, capturing structural similarity and content recall.

 \item \textbf{BLEU-4~\cite{papineni-etal-2002-bleu}:} Measures 4-gram precision, capturing local lexical overlap with the reference response.

 \item \textbf{METEOR~\cite{banerjee-lavie-2005-meteor}:} Accounts for stemming and synonymy, making it better suited to colloquial dialogue and paraphrastic variation.

 \item \textbf{BERTScore (F1)~\cite{Zhang*2020BERTScore}:} Measures semantic similarity using contextual embeddings and serves as the primary meaning-based metric.

 \item \textbf{Realism Score~\cite{zheng2023judging}:} Uses an LLM judge to score behavioral plausibility, linguistic naturalness, and persona adherence. To mitigate single-model preference bias, we employ two independent judges from different model families: Gemini~3.1~Pro and GPT-5.4. Full prompt details are provided in Appendix~\ref{app:llm_judge}. \end{enumerate}

\noindent \textbf{Model selection.} We evaluate five recent instruction-tuned SLMs with fewer than 4 billion parameters: Gemma~2 (2B-Instruct)~\cite{team2024gemma}, Qwen~2.5 (3B-Instruct)~\cite{yang2025qwen3}, Llama~3.2 (3B-Instruct)~\cite{grattafiori2024llama}, Falcon~3 (3B-Instruct)~\cite{falcon3_2024}, and Granite~3.0 (2B-Instruct)~\cite{granite2024granite}. Released in late 2024 to early 2025, these models were selected to span complementary design trade-offs relevant to de-escalation simulation, including efficient reasoning, strong instruction following, dialogue fluency, structured generation, and safety alignment. Table~\ref{tab:model_specs} summarizes their key technical specifications.

\begin{table}[t]
    \centering
    \small
    \resizebox{\textwidth}{!}{%
    \begin{tabular}{lcccc}
        \toprule
        \textbf{Model} & \textbf{Params} &
        \textbf{Context Window} &
        \textbf{Disk Size (FP16)} &
        \textbf{Key Architecture} \\
        \midrule
        Gemma~2 (2B-Instruct)   & 2.6B & 8k   &
            ${\sim}$2.6\,GB & Distillation, SWA \\
        Granite~3.0 (2B-Instruct) & 2.5B & 4k &
            ${\sim}$2.5\,GB & Safety aligned \\
        Qwen~2.5 (3B-Instruct)  & 3.1B & 32k  &
            ${\sim}$3.8\,GB & SwiGLU, RoPE \\
        Falcon~3 (3B-Instruct)  & 3.2B & 32k  &
            ${\sim}$3.5\,GB & GQA, reasoning \\
        Llama~3.2 (3B-Instruct) & 3.2B & 128k &
            ${\sim}$3.4\,GB & GQA, pruned \\
        \bottomrule
    \end{tabular}%
    }
    \vspace{4pt}
    \caption{Technical specifications of the five Small Language
    Models (SLMs) evaluated in this study. Memory estimates
    represent FP16 weights. All models contain fewer than 4
    billion parameters and were released in late 2024 to early
    2025.}
    \label{tab:model_specs}
\end{table}

\subsection{Results and Analysis}
\label{sec:results}
We present a comprehensive evaluation of our fine-tuning methodology across three dimensions: training stability and convergence, the quantitative impact of domain adaptation, and the comparative efficacy of specialized SLMs versus generalist LLMs.

\noindent \textbf{Training dynamics.} Figure~\ref{fig:combined_loss_curves} illustrates the training and validation loss trajectories for all five architectures. We observe highly stable convergence, characterized by rapid initial descent within the first epoch followed by asymptotic stabilization. The divergence between training and validation loss remains negligible throughout fine-tuning across all architectures, indicating effective regularization without overfitting. This stability is attributed to the QLoRA configuration with $r = 16$ and $\alpha = 32$, where the scaling ratio $\nicefrac{\alpha}{r} = 2$ provides sufficient gradient signal for the adapter weights to learn the target distribution without destabilizing the frozen base parameters. These results confirm that 4-bit quantization combined with low-rank adaptation is a robust and compute-efficient strategy for aligning SLMs to high-entropy dialogue domains.

\noindent \textbf{Impact of domain adaptation.} Tables~\ref{tab:results_standard} and~\ref{tab:results_semantic_crossjudge} report pre- and post-fine-tuning performance across all five SLMs. Fine-tuning on \tool yields consistent improvements on every metric across all models, confirming the value of domain-specific adaptation. The most substantial gains appear in the Realism Score (+18.5 to +28.1 points), consistent across both LLM judges (Gemini~3.1~Pro and GPT-5.4), indicating that fine-tuning improves behavioral plausibility and persona adherence beyond what lexical overlap metrics capture. Qwen~2.5 achieves the strongest fine-tuned performance on all five metrics (ROUGE-L: 15.7, BLEU-4: 3.7, METEOR: 19.4, BERTScore: 0.88, Realism: 62.1/60.8). Gemma~2 exhibits the largest lexical gains (+7.5 ROUGE-L, +4.6 METEOR, +0.06 BERTScore) and Granite~3.0 the largest realism improvement (+28.1/+27.2), demonstrating that even 2B-parameter models benefit substantially from in-domain supervision. Falcon~3 exhibits more modest gains across all metrics, which we attribute to stronger RLHF alignment resistance to domain-specific fine-tuning. Overall, off-the-shelf instruct models capture only broad semantic intent, whereas fine-tuning on \tool produces responses that are lexically aligned, semantically faithful, and behaviorally realistic under high-stress conversational conditions.

\begin{table}[t]
    \centering
    \resizebox{\textwidth}{!}{%
    \begin{tabular}{lcccccc}
        \toprule
        \multirow{2}{*}{\textbf{Model}} &
        \multicolumn{2}{c}{\textbf{ROUGE-L}} &
        \multicolumn{2}{c}{\textbf{BLEU-4}} &
        \multicolumn{2}{c}{\textbf{METEOR}} \\
        \cmidrule(lr){2-3}
        \cmidrule(lr){4-5}
        \cmidrule(lr){6-7}
        & \textit{Base} & \textit{FT}$_{\Delta}$
        & \textit{Base} & \textit{FT}$_{\Delta}$
        & \textit{Base} & \textit{FT}$_{\Delta}$ \\
        \midrule
        Qwen~2.5 (3B-Instruct) &
            11.2\,$\pm$\,0.2 &
            \textbf{15.7\,$\pm$\,0.3}$_{\uparrow 4.5}$ &
            0.9\,$\pm$\,0.1 &
            \textbf{3.7\,$\pm$\,0.2}$_{\uparrow 2.8}$ &
            15.7\,$\pm$\,0.2 &
            \textbf{19.4\,$\pm$\,0.4}$_{\uparrow 3.7}$ \\
        Llama~3.2 (3B-Instruct) &
            11.3\,$\pm$\,0.2 &
            \textbf{14.9\,$\pm$\,0.4}$_{\uparrow 3.6}$ &
            0.6\,$\pm$\,0.1 &
            \textbf{2.4\,$\pm$\,0.2}$_{\uparrow 1.8}$ &
            15.7\,$\pm$\,0.2 &
            \textbf{17.7\,$\pm$\,0.3}$_{\uparrow 2.0}$ \\
        Gemma~2 (2B-Instruct) &
            7.2\,$\pm$\,0.2 &
            \textbf{14.7\,$\pm$\,0.3}$_{\uparrow 7.5}$ &
            0.4\,$\pm$\,0.1 &
            \textbf{1.9\,$\pm$\,0.2}$_{\uparrow 1.5}$ &
            11.7\,$\pm$\,0.2 &
            \textbf{16.3\,$\pm$\,0.3}$_{\uparrow 4.6}$ \\
        Granite~3.0 (2B-Instruct) &
            10.2\,$\pm$\,0.1 &
            \textbf{15.6\,$\pm$\,0.4}$_{\uparrow 5.4}$ &
            1.2\,$\pm$\,0.1 &
            \textbf{2.3\,$\pm$\,0.2}$_{\uparrow 1.1}$ &
            13.5\,$\pm$\,0.2 &
            \textbf{16.7\,$\pm$\,0.3}$_{\uparrow 3.2}$ \\
        Falcon~3 (3B-Instruct) &
            9.2\,$\pm$\,0.2 &
            \textbf{11.2\,$\pm$\,0.3}$_{\uparrow 2.0}$ &
            0.5\,$\pm$\,0.1 &
            \textbf{1.2\,$\pm$\,0.1}$_{\uparrow 0.7}$ &
            12.4\,$\pm$\,0.2 &
            \textbf{15.6\,$\pm$\,0.2}$_{\uparrow 3.2}$ \\
        \bottomrule
    \end{tabular}%
    }
    \vspace{4pt}
    \caption{Standard NLP metrics for base versus fine-tuned
    (FT) models (mean\,$\pm$\,SD across 3 independent runs).
    Subscripts indicate the absolute gain ($\uparrow$) from
    the base model. Best results per metric in \textbf{bold}.
    All metrics scaled to 0--100.}
    \label{tab:results_standard}
\end{table}

\begin{table}[t]
    \centering
    \resizebox{\textwidth}{!}{%
    \begin{tabular}{lcccccc}
        \toprule
        \multirow{2}{*}{\textbf{Model}} &
        \multicolumn{2}{c}{\textbf{BERTScore (F1)}} &
        \multicolumn{2}{c}{\textbf{Realism: Gemini 3.1 Pro}} &
        \multicolumn{2}{c}{\textbf{Realism: GPT-5.4}} \\
        \cmidrule(lr){2-3}
        \cmidrule(lr){4-5}
        \cmidrule(lr){6-7}
        & \textit{Base} & \textit{FT}$_{\Delta}$
        & \textit{Base} & \textit{FT}$_{\Delta}$
        & \textit{Base} & \textit{FT}$_{\Delta}$ \\
        \midrule

        Qwen~2.5 (3B-Instruct) &
            0.86\,$\pm$\,0.01 &
            \textbf{0.88\,$\pm$\,0.01}$_{\uparrow 0.02}$ &
            35.4\,$\pm$\,0.4 &
            \textbf{62.1\,$\pm$\,0.5}$_{\uparrow 26.7}$ &
            34.8\,$\pm$\,0.5 &
            \textbf{60.8\,$\pm$\,0.6}$_{\uparrow 26.0}$ \\

        Llama~3.2 (3B-Instruct) &
            0.86\,$\pm$\,0.01 &
            \textbf{0.87\,$\pm$\,0.01}$_{\uparrow 0.01}$ &
            36.1\,$\pm$\,0.3 &
            \textbf{58.4\,$\pm$\,0.6}$_{\uparrow 22.3}$ &
            35.5\,$\pm$\,0.4 &
            \textbf{57.2\,$\pm$\,0.6}$_{\uparrow 21.7}$ \\

        Gemma~2 (2B-Instruct) &
            0.81\,$\pm$\,0.01 &
            \textbf{0.87\,$\pm$\,0.01}$_{\uparrow 0.06}$ &
            28.5\,$\pm$\,0.4 &
            \textbf{55.2\,$\pm$\,0.5}$_{\uparrow 26.7}$ &
            28.1\,$\pm$\,0.5 &
            \textbf{54.0\,$\pm$\,0.6}$_{\uparrow 25.9}$ \\

        Granite~3.0 (2B-Instruct) &
            0.83\,$\pm$\,0.01 &
            \textbf{0.87\,$\pm$\,0.01}$_{\uparrow 0.04}$ &
            32.4\,$\pm$\,0.3 &
            \textbf{60.5\,$\pm$\,0.4}$_{\uparrow 28.1}$ &
            31.8\,$\pm$\,0.4 &
            \textbf{59.0\,$\pm$\,0.5}$_{\uparrow 27.2}$ \\

        Falcon~3 (3B-Instruct) &
            0.83\,$\pm$\,0.01 &
            \textbf{0.85\,$\pm$\,0.01}$_{\uparrow 0.02}$ &
            30.2\,$\pm$\,0.5 &
            \textbf{48.7\,$\pm$\,0.6}$_{\uparrow 18.5}$ &
            29.6\,$\pm$\,0.5 &
            \textbf{47.5\,$\pm$\,0.7}$_{\uparrow 17.9}$ \\

        \bottomrule
    \end{tabular}%
    }
    \vspace{4pt}
    \caption{Semantic fidelity and cross-judge realism evaluation for
    base versus fine-tuned (FT) models. BERTScore is reported on a
    0--1 scale. Realism is reported on a 0--100 scale using two
    independent LLM judges from different model families: Gemini 3.1 Pro
    and GPT-5.4. Subscripts indicate the absolute gain ($\uparrow$)
    from the base model. Best results per metric are shown in
    \textbf{bold}. GPT-5.4 scores are included as a cross-judge
    robustness check to reduce single-model preference bias.}
    \label{tab:results_semantic_crossjudge}
\end{table}

\noindent \textbf{Specialized SLMs vs.\ generalist LLMs.} Table~\ref{tab:model_comparison_crossjudge} compares fine-tuned SLMs against Gemini~2.5 Flash as a strong generalist baseline; prompt details are in Appendix~\ref{app:llm_baseline}. Fine-tuned Qwen~2.5 (3B) achieves the highest score on every metric (ROUGE-L: 15.7, BLEU-4: 3.7, METEOR: 19.4, BERTScore: 0.88, Realism: 62.1/60.8 under Gemini~3.1~Pro and GPT-5.4 respectively) while reducing latency from 3.50\,s to 0.38\,s. Llama~3.2 and Granite~3.0 similarly outperform Gemini on all quality and realism metrics under both judges, with Granite~3.0 achieving the fastest inference at 0.30\,s. Gemma~2 surpasses Gemini on all automatic metrics, though its realism scores (55.2/54.0) remain marginally below the Gemini baseline (55.4/54.1) under both judges. The cross-judge consistency between Gemini~3.1~Pro and GPT-5.4 scores across all models confirms that realism rankings are robust to single-model preference bias. These results support the hypothesis that high-quality domain-specific fine-tuning compensates for model scale: the strongest fine-tuned SLMs are more accurate, more behaviorally realistic, and 8$\times$ to 12$\times$ faster than the generalist baseline (0.30--0.44\,s versus 3.50\,s), which is essential for real-time edge deployment.

\begin{table}[t]
    \centering
    \resizebox{\textwidth}{!}{%
    \begin{tabular}{lccccccc}
        \toprule
        \textbf{Model} &
        \textbf{ROUGE-L} & \textbf{BLEU-4} &
        \textbf{METEOR} & \textbf{BERTScore (F1)} &
        \textbf{Realism} &
        \textbf{Realism} &
        \textbf{Latency (s)} \\
        & & & & &
        \textbf{Gemini 3.1 Pro} &
        \textbf{GPT-5.4} & \\
        \midrule

        \textbf{Qwen~2.5 (3B) FT} &
            \textbf{15.7\,$\pm$\,0.3}$^{*}$ &
            \textbf{3.7\,$\pm$\,0.2}$^{*}$ &
            \textbf{19.4\,$\pm$\,0.4}$^{*}$ &
            \textbf{0.88\,$\pm$\,0.01}$^{*}$ &
            \textbf{62.1\,$\pm$\,0.5}$^{*}$ &
            \textbf{60.8\,$\pm$\,0.6}$^{*}$ &
            0.38\,$\pm$\,0.02 \\

        Llama~3.2 (3B) FT &
            14.9\,$\pm$\,0.4$^{*}$ &
            2.4\,$\pm$\,0.2$^{*}$ &
            17.7\,$\pm$\,0.3$^{*}$ &
            0.87\,$\pm$\,0.01$^{*}$ &
            58.4\,$\pm$\,0.6$^{*}$ &
            57.2\,$\pm$\,0.6$^{*}$ &
            0.44\,$\pm$\,0.03 \\

        Gemma~2 (2B) FT &
            14.7\,$\pm$\,0.3$^{*}$ &
            1.9\,$\pm$\,0.2$^{*}$ &
            16.3\,$\pm$\,0.3$^{*}$ &
            0.87\,$\pm$\,0.01$^{*}$ &
            55.2\,$\pm$\,0.5$^{*}$ &
            54.0\,$\pm$\,0.6$^{*}$ &
            0.33\,$\pm$\,0.02 \\

        Granite~3.0 (2B) FT &
            15.6\,$\pm$\,0.4$^{*}$ &
            2.3\,$\pm$\,0.2$^{*}$ &
            16.7\,$\pm$\,0.3$^{*}$ &
            0.87\,$\pm$\,0.01$^{*}$ &
            60.5\,$\pm$\,0.4$^{*}$ &
            59.0\,$\pm$\,0.5$^{*}$ &
            \textbf{0.30\,$\pm$\,0.02} \\

        Falcon~3 (3B) FT &
            11.2\,$\pm$\,0.3 &
            1.2\,$\pm$\,0.1 &
            15.6\,$\pm$\,0.2$^{*}$ &
            0.85\,$\pm$\,0.01 &
            48.7\,$\pm$\,0.6$^{*}$ &
            47.5\,$\pm$\,0.7$^{*}$ &
            0.41\,$\pm$\,0.03 \\

        \midrule

        Gemini~2.5 Flash (Base) &
            12.4\,$\pm$\,0.3 &
            1.1\,$\pm$\,0.1 &
            13.2\,$\pm$\,0.2 &
            0.85\,$\pm$\,0.01 &
            55.4\,$\pm$\,0.4 &
            54.1\,$\pm$\,0.5 &
            3.50\,$\pm$\,0.15 \\

        \bottomrule
    \end{tabular}%
    }
    \vspace{4pt}
    \caption{Final performance, cross-judge realism, and latency
    comparison of fine-tuned SLMs versus the Gemini~2.5 Flash
    generalist baseline (mean\,$\pm$\,SD across 3 seeds). Realism is
    evaluated using two independent LLM judges from different model
    families, Gemini~3.1 Pro and GPT-5.4, to reduce single-judge
    preference bias. An asterisk ($^{*}$) denotes statistical
    significance versus the Gemini~2.5 Flash baseline under the
    corresponding metric or judge setting (paired $t$-test,
    $p < 0.05$). Best results per metric in \textbf{bold}. All
    fine-tuned SLMs operate at sub-second latency, corresponding to an
    approximately 8$\times$ to 12$\times$ speedup over the API-based
    baseline.}
    \label{tab:model_comparison_crossjudge}
\end{table}

\noindent \textbf{Qualitative analysis.} Table~\ref{tab:qualitative_examples} shows representative responses to two high-stakes officer inputs. Base Qwen~2.5 (3B-Instruct) often produces formal, cooperative, and over-sanitized replies that poorly reflect tense police-civilian encounters. In contrast, fine-tuned Qwen~2.5 generates shorter, colloquial, and emotionally charged responses that better preserve civilian persona and escalation dynamics, closely matching the Gemini~2.5 Flash few-shot baseline. These examples suggest that domain-specific fine-tuning improves pragmatic realism and role consistency for de-escalation simulation.

\begin{table}[h]
    \centering
    \small
    \resizebox{\textwidth}{!}{%
    \begin{tabular}{p{0.22\textwidth} p{0.24\textwidth}
        p{0.24\textwidth} p{0.24\textwidth}}
        \toprule
        \textbf{Officer Input} &
        \textbf{Base Model} \newline
        \textbf{Qwen~2.5 (3B)} &
        \textbf{Gemini~2.5 Flash} \newline
        \textbf{(Few-Shot)} &
        \textbf{Fine-Tuned Qwen~2.5} \newline
        \textbf{(3B, Ours)} \\
        \midrule
        \textit{``Step out of the vehicle and keep your
        hands where I can see them!''} &
        I am afraid I cannot comply with that request.
        I am staying inside. &
        No! What the hell is going on? I haven't done
        anything! Get away from my car! &
        I ain't getting out! You tell me why you stopped
        me first! I didn't do nothing! \\
        \midrule
        \textit{``Stop resisting! Put your hands behind
        your back now!''} &
        Please calm down. I am not resisting. &
        I am not doing anything! Get off me! &
        Get off me! I didn't do nothing! Let go! Why
        are you grabbing me? \\
        \bottomrule
    \end{tabular}%
    }
    \vspace{4pt}
    \caption{Qualitative comparison of generated civilian responses across model configurations.}
    \label{tab:qualitative_examples}
\end{table}

\noindent \textbf{Human expert evaluation.} To validate our automated realism metrics, we conducted a blind human evaluation with two domain specialists: an active law-enforcement de-escalation instructor and a trauma-informed crisis intervention expert. They rated civilian responses from five model conditions across 12 held-out scenarios using a 15-criterion rubric covering emotional authenticity, linguistic naturalism, persona coherence, and situational dynamics (Appendix~\ref{sec:human_eval}). Agreement was substantial overall ($\bar{\kappa}_w = 0.73$), with near-perfect agreement on objective criteria such as natural spoken language ($\kappa_w = 0.82$) and staying in victim role ($\kappa_w = 0.84$). Human rankings align with automated results: fine-tuned Qwen~2.5 performs best overall ($4.28$/5; primary weighted mean $4.35$/5), followed by Gemini~2.5 Flash ($3.90$) and fine-tuned Llama~3.2 ($3.66$), while both base models remain below $2.31$. The largest fine-tuning gains occur in natural spoken language ($+2.50$) and staying in victim role ($+2.40$), where alignment most suppresses the colloquial and emotionally fragmented style needed for realistic victim simulation. Human scores strongly correlate with LLM-as-Judge realism scores ($\rho = 0.81$, $p < 0.001$).

\noindent \textbf{Simulation evaluation.} We evaluate long-horizon interactive behavior using the multi-agent proxy simulation framework in Appendix~\ref{app:simulation_framework}. A fixed Gemini~3.1 Pro Officer Proxy interacts with each base or fine-tuned Suspect Proxy across held-out scenarios. Dialogues are scored by \textit{Realism Score} and \textit{De-escalation Rate}. Figure~\ref{fig:crossjudge_agreement} shows that fine-tuned Qwen~2.5 achieves the best overall simulation performance, leading all evaluated models on both realism and de-escalation across judge settings.
\section{Broader Impacts and Ethical Considerations}
\label{sec:broader_impacts}
\tool aims to democratize access to privacy-preserving, low-latency de-escalation training, supporting improved crisis intervention and reduced use-of-force incidents. We release only fully anonymized textual transcripts; raw audio and video are withheld. Fine-tuned models are strictly restricted to controlled educational simulation and must not be used for surveillance, predictive policing, or profiling.

This research uses exclusively publicly available social media content with no human participant interaction, falling under the public observation exemption (45 CFR~46.104(d)(2))~\cite{hhs_45cfr46104}. We nonetheless implement four governance measures exceeding exempt-research obligations: data minimization, PII de-identification via hybrid NER and LLM-based parsing (Appendix~\ref{app:anonymization}), restricted raw-data storage, and structured annotation protocols with inter-rater reliability checks (Appendix~\ref{app:human_verification}). This is consistent with established NLP practice on publicly sourced corpora~\cite{voigt2017language, rosas2025constructing}.
\section{Limitations}
\label{sec:limitations}
In-the-wild data introduces inherent noise, and the subjective nature of de-escalation produced inter-annotator variances of 0.14 to 1.9; expanding the annotator pool and incorporating additional domain perspectives will be essential for stronger long-context consensus. The benchmark evaluates linguistic and behavioral plausibility, not actual training efficacy, and should not be viewed as validated for operational or field training deployment. Whether interaction with thesemodels improves officer decision-making, reduces use-of-force incidents,or transfers to real-world encounters remains an open empirical question.
\section{Conclusion}
\label{sec:conclusion}
We introduced \tool, the first large-scale benchmark derived from in-the-wild police-civilian interactions, comprising 1,500 real-world scenarios and over 285,000 dialogue turns. Our experiments demonstrate that a fine-tuned 3B-parameter SLM (Qwen~2.5) significantly outperforms a state-of-the-art generalist LLM (Gemini~2.5 Flash) on domain-specific metrics, challenging the prevailing assumption that parameter scale is the primary driver of performance. This finding establishes that high-quality, domain-specific data is a viable substitute for model scale in specialized high-stakes tasks. \tool lays the groundwork for privacy-preserving, low-latency de-escalation trainers deployable on edge devices without cloud connectivity.

\bibliography{example_paper}

@inproceedings{banerjee-lavie-2005-meteor,
    title = "{METEOR}: An Automatic Metric for {MT} Evaluation with Improved Correlation with Human Judgments",
    author = "Banerjee, Satanjeev  and
      Lavie, Alon",
    editor = "Goldstein, Jade  and
      Lavie, Alon  and
      Lin, Chin-Yew  and
      Voss, Clare",
    booktitle = "Proceedings of the {ACL} Workshop on Intrinsic and Extrinsic Evaluation Measures for Machine Translation and/or Summarization",
    month = jun,
    year = "2005",
    address = "Ann Arbor, Michigan",
    publisher = "Association for Computational Linguistics",
    url = "https://aclanthology.org/W05-0909/",
    pages = "65--72"
}

@inproceedings{lin-2004-rouge,
    title = "{ROUGE}: A Package for Automatic Evaluation of Summaries",
    author = "Lin, Chin-Yew",
    booktitle = "Text Summarization Branches Out",
    month = jul,
    year = "2004",
    address = "Barcelona, Spain",
    publisher = "Association for Computational Linguistics",
    url = "https://aclanthology.org/W04-1013/",
    pages = "74--81"
}

@inproceedings{papineni-etal-2002-bleu,
    title = "{B}leu: a Method for Automatic Evaluation of Machine Translation",
    author = "Papineni, Kishore  and
      Roukos, Salim  and
      Ward, Todd  and
      Zhu, Wei-Jing",
    editor = "Isabelle, Pierre  and
      Charniak, Eugene  and
      Lin, Dekang",
    booktitle = "Proceedings of the 40th Annual Meeting of the Association for Computational Linguistics",
    month = jul,
    year = "2002",
    address = "Philadelphia, Pennsylvania, USA",
    publisher = "Association for Computational Linguistics",
    url = "https://aclanthology.org/P02-1040/",
    doi = "10.3115/1073083.1073135",
    pages = "311--318"
}

@inproceedings{
Zhang*2020BERTScore,
title={BERTScore: Evaluating Text Generation with BERT},
author={Tianyi Zhang* and Varsha Kishore* and Felix Wu* and Kilian Q. Weinberger and Yoav Artzi},
booktitle={International Conference on Learning Representations},
year={2020},
url={https://openreview.net/forum?id=SkeHuCVFDr}
}

@inproceedings{sridhar2025adaptive,
  title={Adaptive De-escalation Trainer: Piloting a RAG-Enhanced, Emotionally Modulated AI Simulator for Police Training},
  author={Sridhar, Eshwara Prasad and Lopez, Jose and Islam, Mohammad and Deb, Shuchisnigdha},
  booktitle={Proceedings of the Human Factors and Ergonomics Society Annual Meeting},
  volume={69},
  number={1},
  pages={171--175},
  year={2025},
  organization={SAGE Publications Sage CA: Los Angeles, CA}
}

@article{granite2024granite,
  title={Granite 3.0 Language Models},
  author={Granite Team, IBM},
  journal={URL: https://github. com/ibm-granite/granite-3.0-language-models},
  year={2024}
}

@article{grattafiori2024llama,
  title={The llama 3 herd of models},
  author={Grattafiori, Aaron and Dubey, Abhimanyu and Jauhri, Abhinav and Pandey, Abhinav and Kadian, Abhishek and Al-Dahle, Ahmad and Letman, Aiesha and Mathur, Akhil and Schelten, Alan and Vaughan, Alex and others},
  journal={arXiv preprint arXiv:2407.21783},
  year={2024}
}

@article{team2024gemma,
  title={Gemma 2: Improving open language models at a practical size},
  author={Team, Gemma and Riviere, Morgane and Pathak, Shreya and Sessa, Pier Giuseppe and Hardin, Cassidy and Bhupatiraju, Surya and Hussenot, L{\'e}onard and Mesnard, Thomas and Shahriari, Bobak and Ram{\'e}, Alexandre and others},
  journal={arXiv preprint arXiv:2408.00118},
  year={2024}
}

@article{yang2025qwen3,
  title={Qwen3 technical report},
  author={Yang, An and Li, Anfeng and Yang, Baosong and Zhang, Beichen and Hui, Binyuan and Zheng, Bo and Yu, Bowen and Gao, Chang and Huang, Chengen and Lv, Chenxu and others},
  journal={arXiv preprint arXiv:2505.09388},
  year={2025}
}

@article{gao2024large,
  title={Large language models empowered agent-based modeling and simulation: A survey and perspectives},
  author={Gao, Chen and Lan, Xiaochong and Li, Nian and Yuan, Yuan and Ding, Jingtao and Zhou, Zhilun and Xu, Fengli and Li, Yong},
  journal={Humanities and Social Sciences Communications},
  volume={11},
  number={1},
  pages={1--24},
  year={2024},
  publisher={Palgrave}
}

@article{violakis2025leveraging,
  title={Leveraging large language models for enhanced simulation-based learning in police and law enforcement},
  author={Violakis, Petros},
  journal={Policing: A Journal of Policy and Practice},
  volume={19},
  pages={paaf012},
  year={2025},
  publisher={Oxford University Press UK}
}

@inproceedings{park2023generative,
  title={Generative agents: Interactive simulacra of human behavior},
  author={Park, Joon Sung and O'Brien, Joseph and Cai, Carrie Jun and Morris, Meredith Ringel and Liang, Percy and Bernstein, Michael S},
  booktitle={Proceedings of the 36th annual acm symposium on user interface software and technology},
  pages={1--22},
  year={2023}
}

@inproceedings{wang2024rolellm,
  title={Rolellm: Benchmarking, eliciting, and enhancing role-playing abilities of large language models},
  author={Wang, Noah and Peng, Zy and Que, Haoran and Liu, Jiaheng and Zhou, Wangchunshu and Wu, Yuhan and Guo, Hongcheng and Gan, Ruitong and Ni, Zehao and Yang, Jian and others},
  booktitle={Findings of the Association for Computational Linguistics: ACL 2024},
  pages={14743--14777},
  year={2024}
}

@inproceedings{
wang2025coser,
title={Co{SER}: Coordinating {LLM}-Based Persona Simulation of Established Roles},
author={Xintao Wang and Heng Wang and Yifei Zhang and Xinfeng Yuan and Rui Xu and Jen-tse Huang and Siyu Yuan and Haoran Guo and Jiangjie Chen and Shuchang Zhou and Wei Wang and Yanghua Xiao},
booktitle={Forty-second International Conference on Machine Learning},
year={2025},
url={https://openreview.net/forum?id=BOrR7YqKUt}
}

@inproceedings{zhan2024let,
  title={Let's Negotiate! A Survey of Negotiation Dialogue Systems},
  author={Zhan, Haolan and Wang, Yufei and Li, Zhuang and Feng, Tao and Hua, Yuncheng and Sharma, Suraj and Qu, Lizhen and Semnani-Azad, Zhaleh and Zukerman, Ingrid and Haffari, Reza},
  booktitle={EACL (Findings)},
  year={2024}
}

@inproceedings{rosas2025constructing,
  title={Constructing Datasets From Public Police Body Camera Footage},
  author={Rosas-Smith, Jamie and Bartelds, Martijn and Huang, Ruizhe and Garc{\'\i}a-Perera, Leibny Paola and Livescu, Karen and Jurafsky, Dan and Field, Anjalie},
  booktitle={ICASSP 2025-2025 IEEE International Conference on Acoustics, Speech and Signal Processing (ICASSP)},
  pages={1--5},
  year={2025},
  organization={IEEE}
}

@article{srbinovska2025towards,
  title={Towards AI-Driven Policing: Interdisciplinary Knowledge Discovery from Police Body-Worn Camera Footage},
  author={Srbinovska, Anita and Srbinovska, Angela and Senthil, Vivek and Martin, Adrian and McCluskey, John and Bateman, Jonathan and Fokou{\u{A}}{\v{S}}, Ernest},
  journal={arXiv preprint arXiv:2504.20007},
  year={2025}
}

@inproceedings{pecher-etal-2025-comparing,
    title = "Comparing Specialised Small and General Large Language Models on Text Classification: 100 Labelled Samples to Achieve Break-{E}ven Performance",
    author = "Pecher, Branislav  and
      Srba, Ivan  and
      Bielikova, Maria",
    editor = "Christodoulopoulos, Christos  and
      Chakraborty, Tanmoy  and
      Rose, Carolyn  and
      Peng, Violet",
    booktitle = "Proceedings of the 2025 Conference on Empirical Methods in Natural Language Processing",
    month = nov,
    year = "2025",
    address = "Suzhou, China",
    publisher = "Association for Computational Linguistics",
    url = "https://aclanthology.org/2025.emnlp-main.9/",
    doi = "10.18653/v1/2025.emnlp-main.9",
    pages = "165--184",
    ISBN = "979-8-89176-332-6"
}

@inproceedings{xu-etal-2024-small,
    title = "Small Models are Valuable Plug-ins for Large Language Models",
    author = "Xu, Canwen  and
      Xu, Yichong  and
      Wang, Shuohang  and
      Liu, Yang  and
      Zhu, Chenguang  and
      McAuley, Julian",
    editor = "Ku, Lun-Wei  and
      Martins, Andre  and
      Srikumar, Vivek",
    booktitle = "Findings of the Association for Computational Linguistics: ACL 2024",
    month = aug,
    year = "2024",
    address = "Bangkok, Thailand",
    publisher = "Association for Computational Linguistics",
    url = "https://aclanthology.org/2024.findings-acl.18/",
    doi = "10.18653/v1/2024.findings-acl.18",
    pages = "283--294"
}

@article{comanici2025gemini,
  title={Gemini 2.5: Pushing the frontier with advanced reasoning, multimodality, long context, and next generation agentic capabilities},
  author={Comanici, Gheorghe and Bieber, Eric and Schaekermann, Mike and Pasupat, Ice and Sachdeva, Noveen and Dhillon, Inderjit and Blistein, Marcel and Ram, Ori and Zhang, Dan and Rosen, Evan and others},
  journal={arXiv preprint arXiv:2507.06261},
  year={2025}
}

@misc{falcon3_2024,
  title = {The Falcon 3 Family of Open Models},
  author = {{Falcon-LLM Team}},
  year = {2024},
  month = {December},
  publisher = {Hugging Face},
  howpublished = {\url{https://huggingface.co/blog/falcon3}},
  note = {Accessed: 2026-01-24}
}

@InProceedings{ANAND2024EXP,
author    = {Anand, A. and Polyak, E.},
title     = {EXPLORING THE POTENTIAL OF LARGE LANGUAGE MODELS FOR ENHANCED VIRTUAL NON-PLAYER CHARACTER INTERACTIONS},
series    = {18th International Technology, Education and Development Conference},
booktitle = {INTED2024 Proceedings},
isbn      = {978-84-09-59215-9},
issn      = {2340-1079},
doi       = {10.21125/inted.2024.1269},
url       = {https://doi.org/10.21125/inted.2024.1269},
publisher = {IATED},
location  = {Valencia, Spain},
month     = {4-6 March, 2024},
year      = {2024},
pages     = {4895-4898}
}

@article{voigt2017language,
  title={Language from police body camera footage shows racial disparities in officer respect},
  author={Voigt, Rob and Camp, Nicholas P and Prabhakaran, Vinodkumar and Hamilton, William L and Hetey, Rebecca C and Griffiths, Camilla M and Jurgens, David and Jurafsky, Dan and Eberhardt, Jennifer L},
  journal={Proceedings of the national Academy of sciences},
  volume={114},
  number={25},
  pages={6521--6526},
  year={2017},
  publisher={National Academy of Sciences}
}

@InProceedings{pmlr-v202-radford23a,
  title = 	 {Robust Speech Recognition via Large-Scale Weak Supervision},
  author =       {Radford, Alec and Kim, Jong Wook and Xu, Tao and Brockman, Greg and Mcleavey, Christine and Sutskever, Ilya},
  booktitle = 	 {Proceedings of the 40th International Conference on Machine Learning},
  pages = 	 {28492--28518},
  year = 	 {2023},
  editor = 	 {Krause, Andreas and Brunskill, Emma and Cho, Kyunghyun and Engelhardt, Barbara and Sabato, Sivan and Scarlett, Jonathan},
  volume = 	 {202},
  series = 	 {Proceedings of Machine Learning Research},
  month = 	 {23--29 Jul},
  publisher =    {PMLR},
  url = 	 {https://proceedings.mlr.press/v202/radford23a.html}
}

@inproceedings{bredin2023pyannote,
  title={pyannote. audio 2.1 speaker diarization pipeline: principle, benchmark, and recipe},
  author={Bredin, Herv{\'e}},
  booktitle={24th INTERSPEECH Conference (INTERSPEECH 2023)},
  pages={1983--1987},
  year={2023},
  organization={ISCA}
}

@inproceedings{
zheng2023judging,
title={Judging {LLM}-as-a-Judge with {MT}-Bench and Chatbot Arena},
author={Lianmin Zheng and Wei-Lin Chiang and Ying Sheng and Siyuan Zhuang and Zhanghao Wu and Yonghao Zhuang and Zi Lin and Zhuohan Li and Dacheng Li and Eric Xing and Hao Zhang and Joseph E. Gonzalez and Ion Stoica},
booktitle={Thirty-seventh Conference on Neural Information Processing Systems Datasets and Benchmarks Track},
year={2023},
url={https://openreview.net/forum?id=uccHPGDlao}
}

@article{loshchilov2017decoupled,
  title={Decoupled weight decay regularization},
  author={Loshchilov, Ilya and Hutter, Frank},
  journal={arXiv preprint arXiv:1711.05101},
  year={2017}
}

@article{dettmers2023qlora,
  title={Qlora: Efficient finetuning of quantized llms},
  author={Dettmers, Tim and Pagnoni, Artidoro and Holtzman, Ari and Zettlemoyer, Luke},
  journal={Advances in neural information processing systems},
  volume={36},
  pages={10088--10115},
  year={2023}
}

@inproceedings{blodgett2016demographic,
  title={Demographic dialectal variation in social media: A case study of African-American English},
  author={Blodgett, Su Lin and Green, Lisa and O’Connor, Brendan},
  booktitle={Proceedings of the 2016 conference on empirical methods in natural language processing},
  pages={1119--1130},
  year={2016}
}

@misc{hhs_45cfr46104,
  title        = {{45 CFR \S 46.104: Exempt research}},
  author       = {{U.S. Department of Health and Human Services}},
  year         = {2026},
  howpublished = {\url{https://www.ecfr.gov/current/title-45/part-46/section-46.104}},
  note         = {Electronic Code of Federal Regulations, accessed May 2, 2026}
}
\bibliographystyle{plainnat}

\newpage
\section*{NeurIPS Paper Checklist}

\begin{enumerate}

\item {\bf Claims}
    \item[] Question: Do the main claims made in the abstract and introduction accurately reflect the paper's contributions and scope?
    \item[] Answer: \answerYes{} 
    \item[] Justification: The main claims made in the abstract and introduction accurately reflect the paper's contributions and scope detailed throughout the manuscript. The introduction's assertion regarding the creation of a novel benchmark dataset for human-centered AI aligns perfectly with the data collection and de-identification pipelines detailed in the methodology. Furthermore, the claims concerning the optimization of SLMs for automated de-escalation training are directly supported by the pre-post fine-tuning evaluation protocol discussed in the text. Finally, the abstract's claim of improved real-world applicability is strongly substantiated by our novel multi-agent Simulation Evaluation framework, with the specific assertions of enhanced psychological realism and responsive escalation rates fully backed by the within-subjects statistical analysis and proxy interaction results presented in the findings.

\item {\bf Limitations}
    \item[] Question: Does the paper discuss the limitations of the work performed by the authors?
    \item[] Answer: \answerYes{} 
    \item[] Justification: There is limitations section in the paper.

\item {\bf Theory assumptions and proofs}
    \item[] Question: For each theoretical result, does the paper provide the full set of assumptions and a complete (and correct) proof?
    \item[] Answer: \answerNA{} 
    \item[] Justification: The paper does not include theoretical results.

    \item {\bf Experimental result reproducibility}
    \item[] Question: Does the paper fully disclose all the information needed to reproduce the main experimental results of the paper to the extent that it affects the main claims and/or conclusions of the paper (regardless of whether the code and data are provided or not)?
    \item[] Answer: \answerYes{} 
    \item[] Justification: All design methodologies, experimental setups, and system prompts are fully disclosed within the manuscript and its appendices to ensure complete reproducibility. The paper provides comprehensive descriptions of the pre-post fine-tuning evaluation protocol, the novel multi-agent Simulation Evaluation framework, and the specific configurations utilized for both the Officer and Suspect proxies. Furthermore, all evaluation criteria, statistical analysis parameters, and the exact language model versions tested are thoroughly documented, equipping researchers with all the necessary information to accurately replicate the experimental results and validate the paper's main conclusions.

\item {\bf Open access to data and code}
    \item[] Question: Does the paper provide open access to the data and code, with sufficient instructions to faithfully reproduce the main experimental results, as described in supplemental material?
    \item[] Answer: \answerYes{} 
    \item[] Justification: All code and data have submitted with the manuscript.

\item {\bf Experimental setting/details}
    \item[] Question: Does the paper specify all the training and test details (e.g., data splits, hyperparameters, how they were chosen, type of optimizer) necessary to understand the results?
    \item[] Answer: \answerYes{}{} 
    \item[] Justification: All relevant information is provided in the experimental setup section and the appendix.

\item {\bf Experiment statistical significance}
    \item[] Question: Does the paper report error bars suitably and correctly defined or other appropriate information about the statistical significance of the experiments?
    \item[] Answer: \answerYes{} 
    \item[] Justification: Mean and standard deviation (SD) are reported across three independent runs, with statistical significance indicated.

\item {\bf Experiments compute resources}
    \item[] Question: For each experiment, does the paper provide sufficient information on the computer resources (type of compute workers, memory, time of execution) needed to reproduce the experiments?
    \item[] Answer: \answerYes{} 
    \item[] Justification: All the information is provided.
    
\item {\bf Code of ethics}
    \item[] Question: Does the research conducted in the paper conform, in every respect, with the NeurIPS Code of Ethics \url{https://neurips.cc/public/EthicsGuidelines}?
    \item[] Answer: \answerYes{} 
    \item[] Justification:This research comply with NeurIPS Code Ethics.

\item {\bf Broader impacts}
    \item[] Question: Does the paper discuss both potential positive societal impacts and negative societal impacts of the work performed?
    \item[] Answer: \answerYes{}{} 
    \item[] Justification: This research has a discussion section for this issue. 
    
\item {\bf Safeguards}
    \item[] Question: Does the paper describe safeguards that have been put in place for responsible release of data or models that have a high risk for misuse (e.g., pre-trained language models, image generators, or scraped datasets)?
    \item[] Answer: \answerYes{} 
    \item[] Justification: The paper describes a controlled release protocol for anonymized transcripts only, excludes raw audio/video, masks PII through a hybrid NER and LLM pipeline, prohibits surveillance and predictive policing uses, and provides a takedown mechanism for affected individuals or rights holders.
    

\item {\bf Licenses for existing assets}
    \item[] Question: Are the creators or original owners of assets (e.g., code, data, models), used in the paper, properly credited and are the license and terms of use explicitly mentioned and properly respected?
    \item[] Answer: \answerYes{} 
    \item[] Justification: The original creators of all assets utilized in this research, including the foundational models sourced from Hugging Face, have been properly credited through appropriate citations.

\item {\bf New assets}
    \item[] Question: Are new assets introduced in the paper well documented and is the documentation provided alongside the assets?
    \item[] Answer: \answerYes{} 
    \item[] Justification: We give the example of our datasets how it looks like. 

\item {\bf Crowdsourcing and research with human subjects}
    \item[] Question: For crowdsourcing experiments and research with human subjects, does the paper include the full text of instructions given to participants and screenshots, if applicable, as well as details about compensation (if any)? 
    \item[] Answer: \answerNA{} 
    \item[] Justification: This research does not constitute human subjects research as defined by federal802
regulations (45 CFR 46) and does not involve crowdsourcing. No participants were recruited,803
no interventions were administered, and no private information was collected directly from804
individuals. The dataset is derived exclusively from publicly available videos on open805
social media platforms (YouTube, TikTok, and Facebook), where the content was voluntarily806
shared by creators and is accessible to any member of the public without authentication.

\item {\bf Institutional review board (IRB) approvals or equivalent for research with human subjects}
    \item[] Question: Does the paper describe potential risks incurred by study participants, whether such risks were disclosed to the subjects, and whether Institutional Review Board (IRB) approvals (or an equivalent approval/review based on the requirements of your country or institution) were obtained?
    \item[] Answer: \answerNA{} 
    \item[] Justification: This research does not constitute human subjects 
research as defined by federal regulations (45 CFR 46) and does not 
involve crowdsourcing. No participants were recruited, no interventions 
were administered, and no private information was collected directly 
from individuals. The dataset is derived exclusively from publicly 
available videos on open social media platforms (YouTube, TikTok, and 
Facebook), where the content was voluntarily shared by creators and is 
accessible to any member of the public without authentication. This 
collection methodology is analogous to the use of publicly available 
archival or observational data, which is explicitly exempt from IRB 
oversight under 45 CFR 46.104(d)(4).

Critically, the public release of \tool consists solely of 
fully anonymized textual transcripts. All raw audio, video, and 
visual modalities are withheld. Prior to release, all personally 
identifiable information (PII)---including names, badge numbers, 
locations, and protected health information---is removed through a 
hybrid anonymization pipeline combining Named Entity Recognition (NER), 
LLM-based parsing, and manual verification (detailed in Appendix E). 
The released artifact therefore contains no data that could 
re-identify any individual, and its form does not meet the definition 
of human subjects data under applicable guidelines.

Raw source materials are stored on password-protected infrastructure 
within a restricted-access research environment, consistent with the 
research team's institutional data governance protocols. Human 
annotation processes are governed by structured protocols with 
inter-rater reliability checks and bias audits, as described in 
Appendices C.4 and D.2.

\item {\bf Declaration of LLM usage}
    \item[] Question: Does the paper describe the usage of LLMs if it is an important, original, or non-standard component of the core methods in this research? Note that if the LLM is used only for writing, editing, or formatting purposes and does \emph{not} impact the core methodology, scientific rigor, or originality of the research, declaration is not required.
    \item[] Answer: \answerYes{} 
    \item[] Justification: We utilize an "LLM-as-a-Judge" framework as a core component of our dataset filtration and simulation methodology. We fully disclose how the LLM was applied and provide the exact prompts used in the text.

\end{enumerate}
\newpage

\appendix
\section*{Appendix Table of Contents}

\startcontents[appendix]
\printcontents[appendix]{l}{1}{\setcounter{tocdepth}{2}}
\vspace{2em} 
\hrule 
\vspace{2em}
\pagebreak

\section{Training Dynamics Graph}
\begin{figure*}[h]
    \centering
    \begin{subfigure}[b]{0.48\textwidth}
        \centering
        \includegraphics[width=\linewidth]{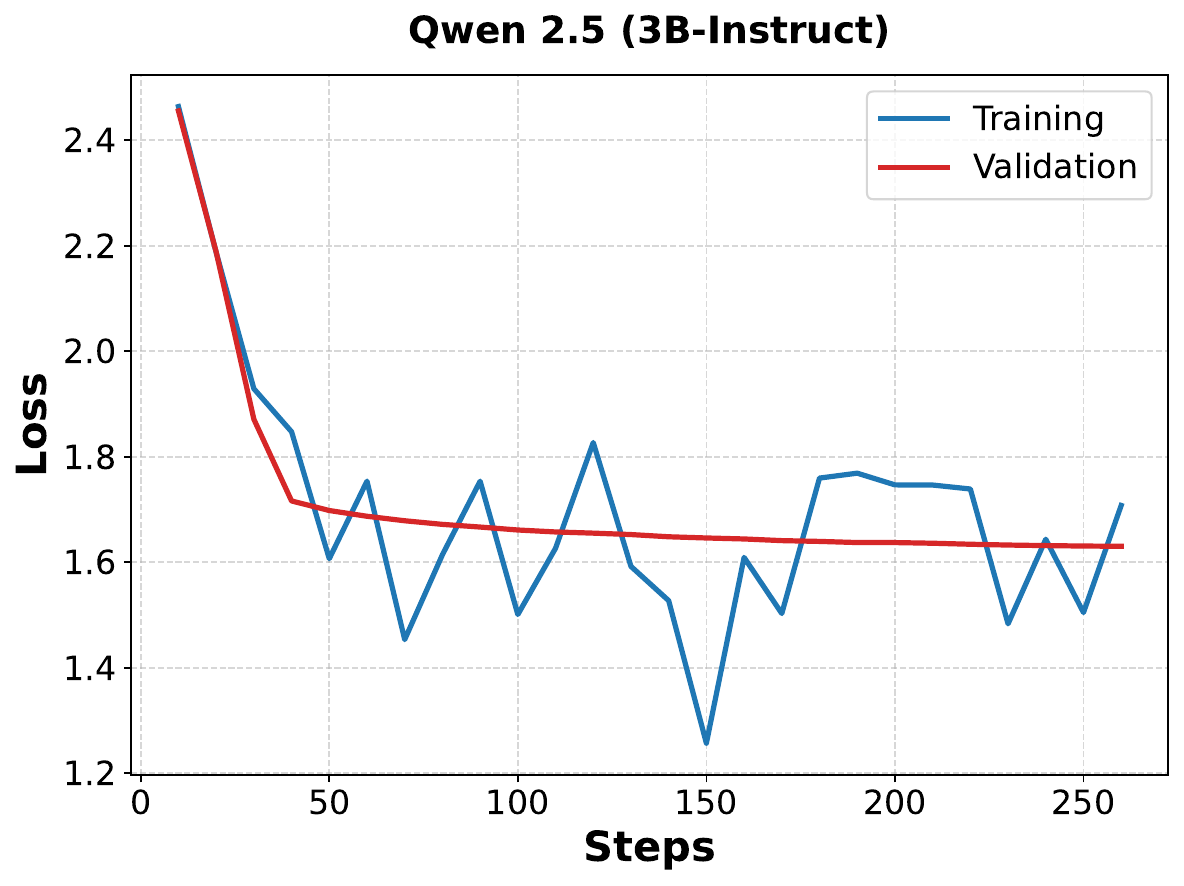}
        \caption{\textbf{Qwen 2.5 (3B)}}
        \label{fig:loss_qwen}
    \end{subfigure}
    \hfill
    \begin{subfigure}[b]{0.48\textwidth}
        \centering
        \includegraphics[width=\linewidth]{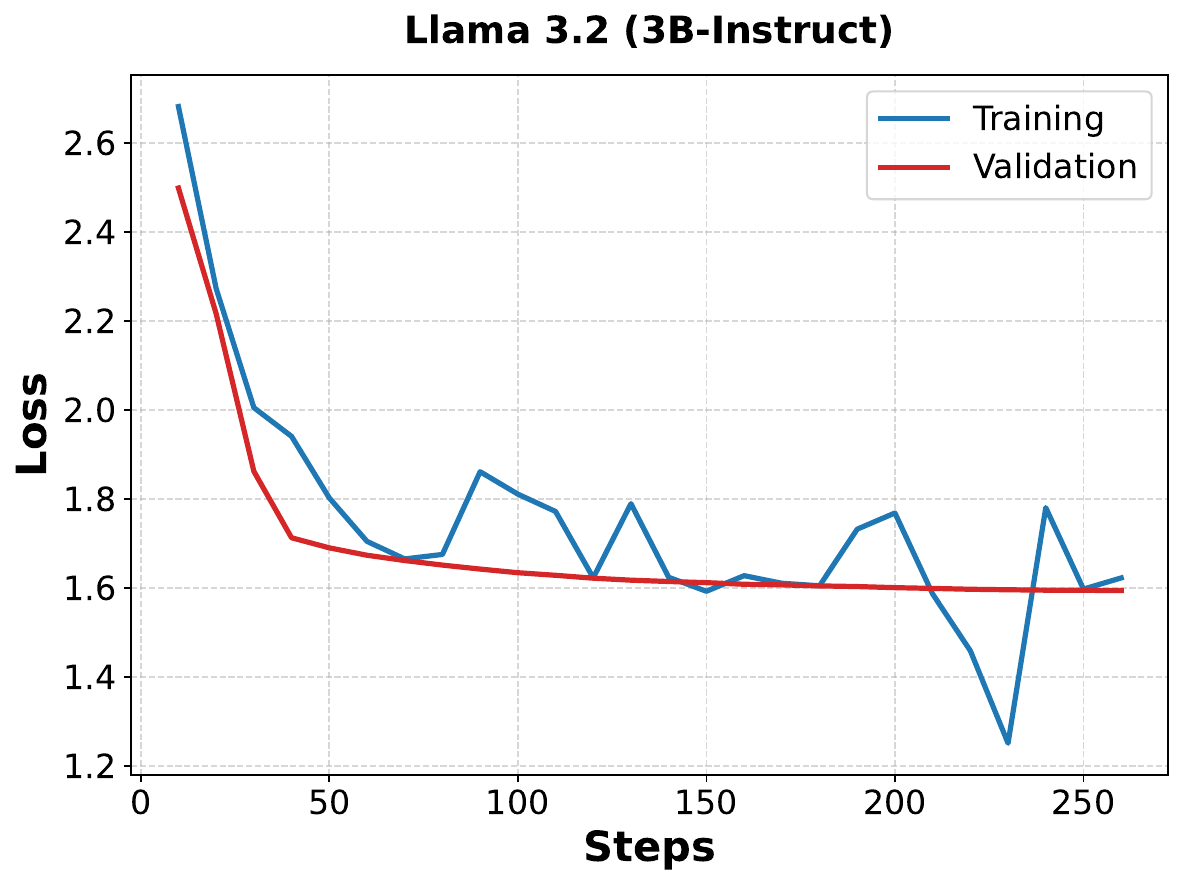}
        \caption{\textbf{Llama 3.2 (3B)}}
        \label{fig:loss_llama}
    \end{subfigure}
    
    \vspace{0.3cm}
    
    \begin{subfigure}[b]{0.48\textwidth}
        \centering
        \includegraphics[width=\linewidth]{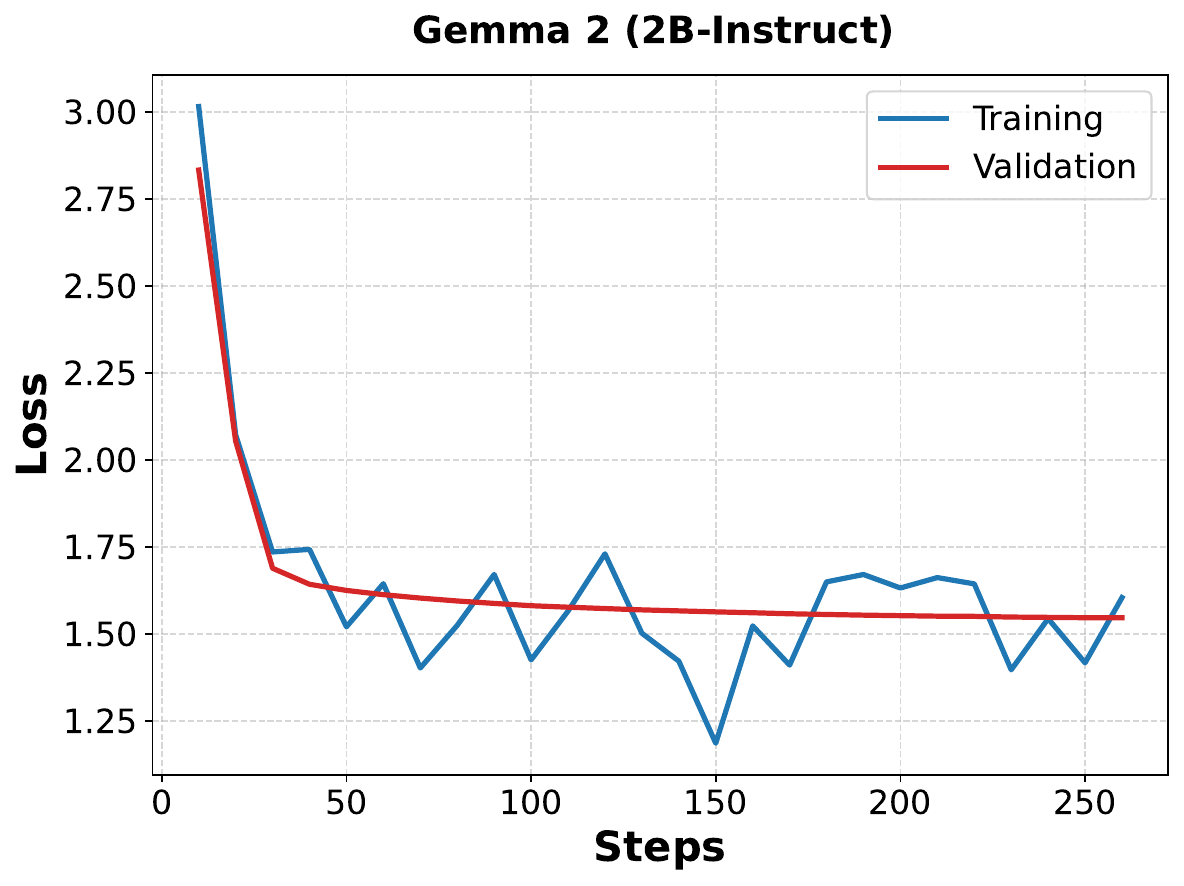}
        \caption{\textbf{Gemma 2 (2B)}}
        \label{fig:loss_gemma}
    \end{subfigure}
    \hfill
    \begin{subfigure}[b]{0.48\textwidth}
        \centering
        \includegraphics[width=\linewidth]{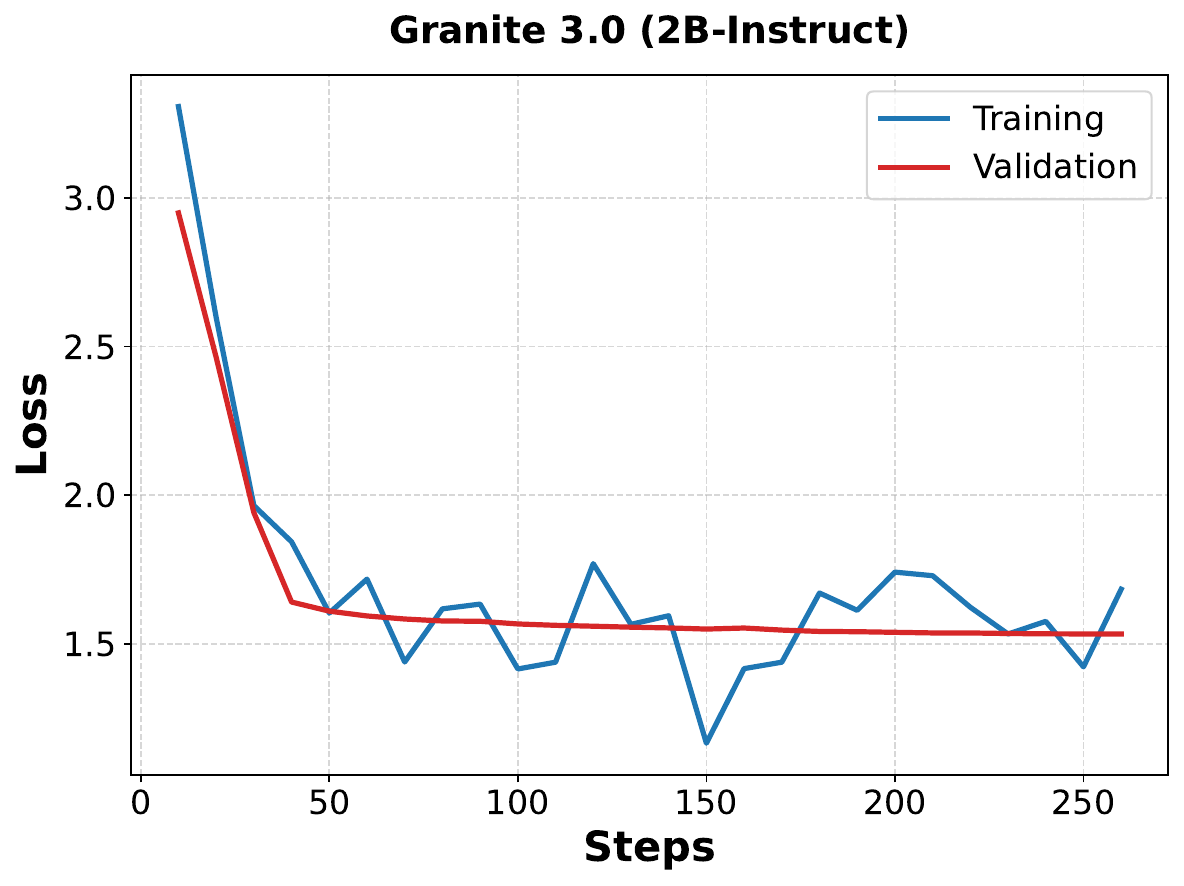}
        \caption{\textbf{Granite 3 (2B)}}
        \label{fig:loss_granite}
    \end{subfigure}
    
    \vspace{0.3cm}

    \begin{subfigure}[b]{0.48\textwidth}
        \centering
        \includegraphics[width=\linewidth]{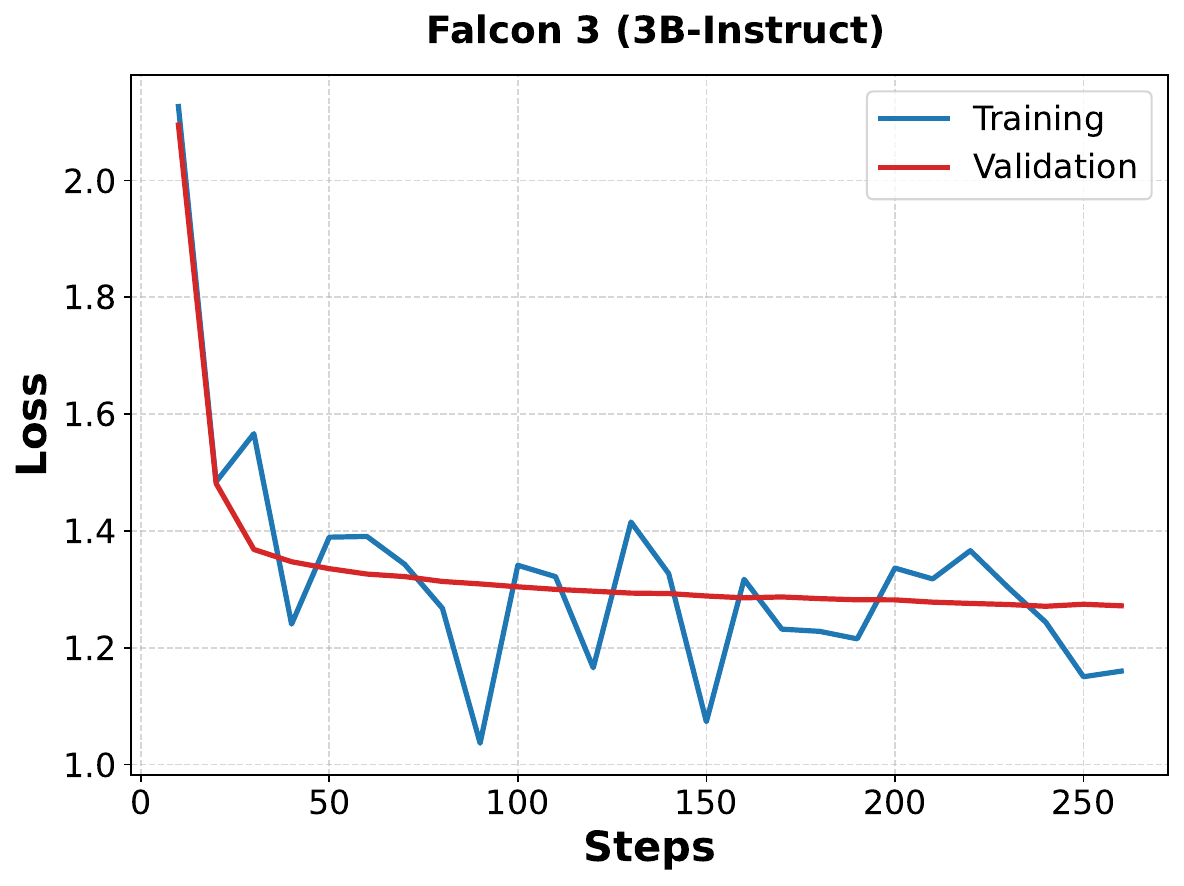}
        \caption{\textbf{Falcon 3 (3B)}}
        \label{fig:loss_falcon}
    \end{subfigure}
    
    \caption{\textbf{Training Dynamics and Convergence Analysis.} We report the training (blue) and validation (red) loss trajectories for (a) Qwen 2.5, (b) Llama 3.2, (c) Gemma 2, (d) Granite 3, and (e) Falcon 3 during fine-tuning on the \tool dataset. All models exhibit rapid initial convergence within the first 50 steps, followed by a stabilization phase. The close alignment between training and validation curves across all architectures indicates that the models are effectively learning domain-specific features without succumbing to overfitting.}
    \label{fig:combined_loss_curves}
\end{figure*}


\begin{figure}[t]
    \centering
    \includegraphics[width=\textwidth]{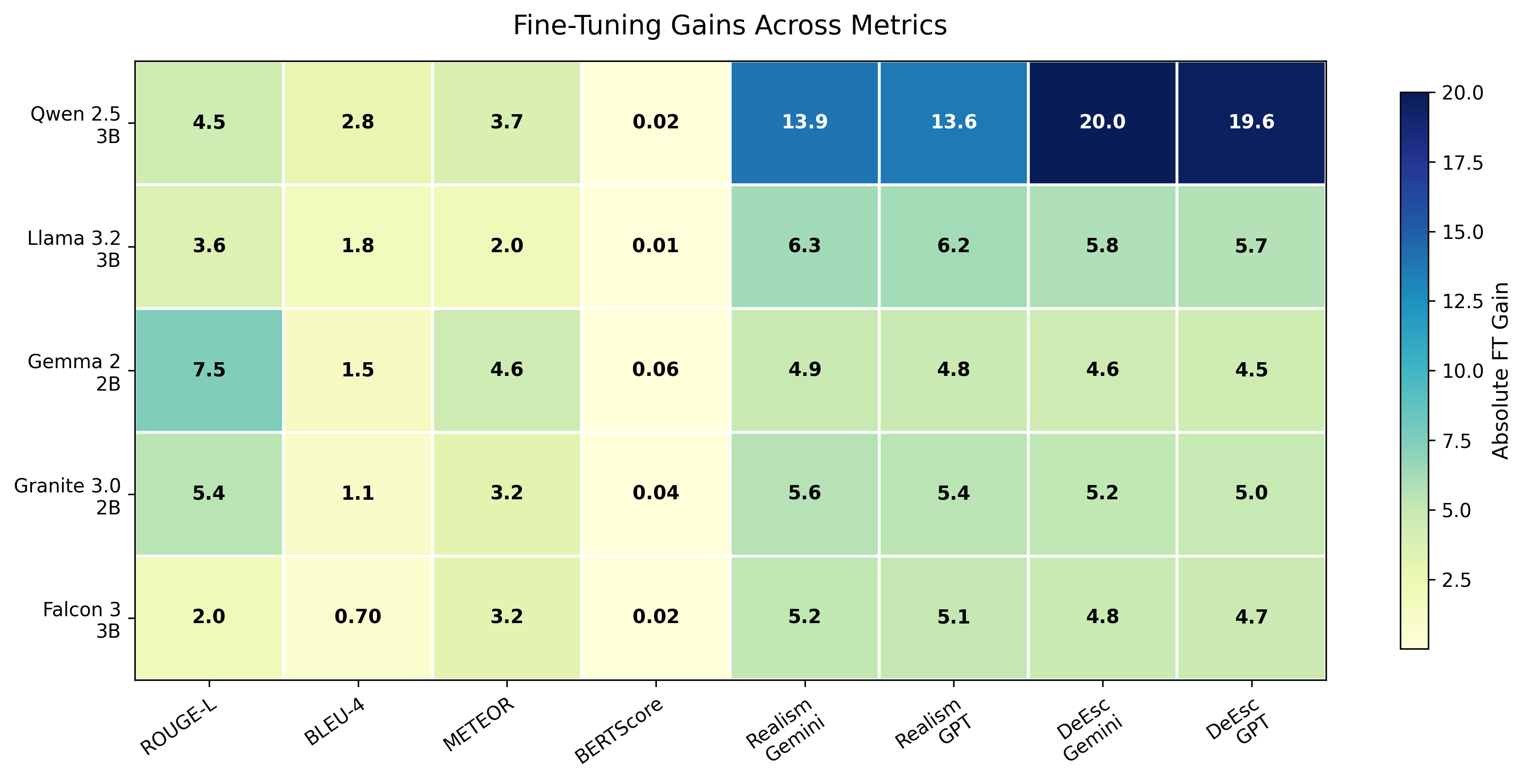}
    \caption{
    \textbf{Fine-tuning gains across evaluation metrics.}
    Heatmap showing the absolute improvement from the base model to the
    fine-tuned model across automatic lexical metrics, semantic similarity,
    realism, and de-escalation. Rows correspond to open-weight models and
    columns correspond to evaluation metrics. Darker cells indicate larger
    absolute gains. Fine-tuning improves all models across all reported
    metrics, with especially large gains in realism and de-escalation for
    Qwen~2.5.
    }
    \label{fig:ft_gain_heatmap}
\end{figure}


\begin{figure}[t]
    \centering
    \includegraphics[width=0.82\textwidth]{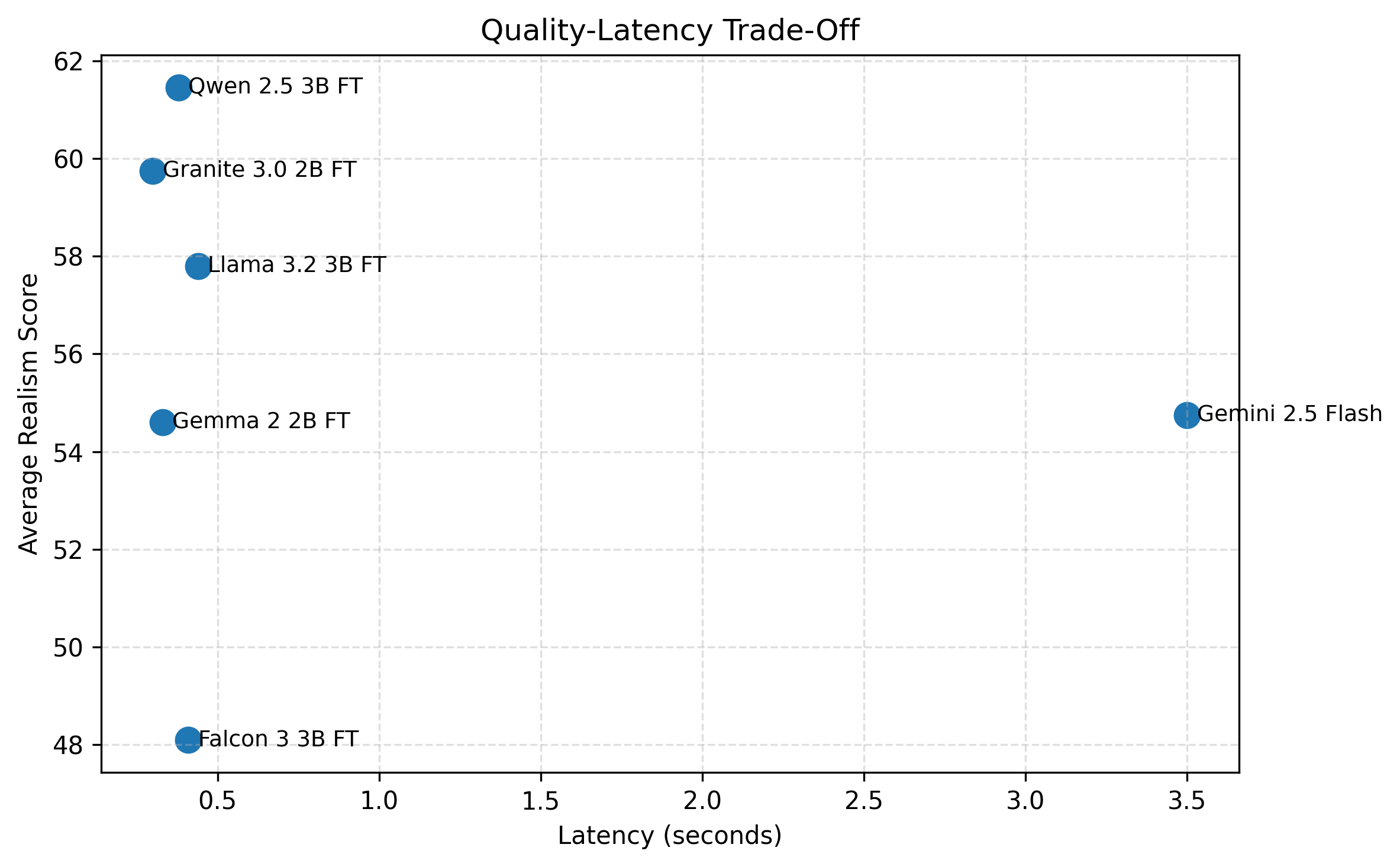}
    \caption{
    \textbf{Quality--latency trade-off between fine-tuned SLMs and the
    Gemini~2.5 Flash baseline.}
    Each point represents one model configuration. The $x$-axis shows
    inference latency in seconds, while the $y$-axis shows average realism,
    computed as the mean of the Gemini~3.1 Pro and GPT-5.4 realism scores.
    Fine-tuned SLMs achieve sub-second latency while
    remaining competitive with, or outperforming, the Gemini~2.5 Flash
    baseline in realism. Qwen~2.5 provides the strongest overall realism,
    whereas Granite~3.0 provides the lowest latency.
    }
    \label{fig:quality_latency}
\end{figure}


\begin{figure}[t]
    \centering

    \begin{subfigure}[t]{0.48\textwidth}
        \centering
        \includegraphics[width=\textwidth]{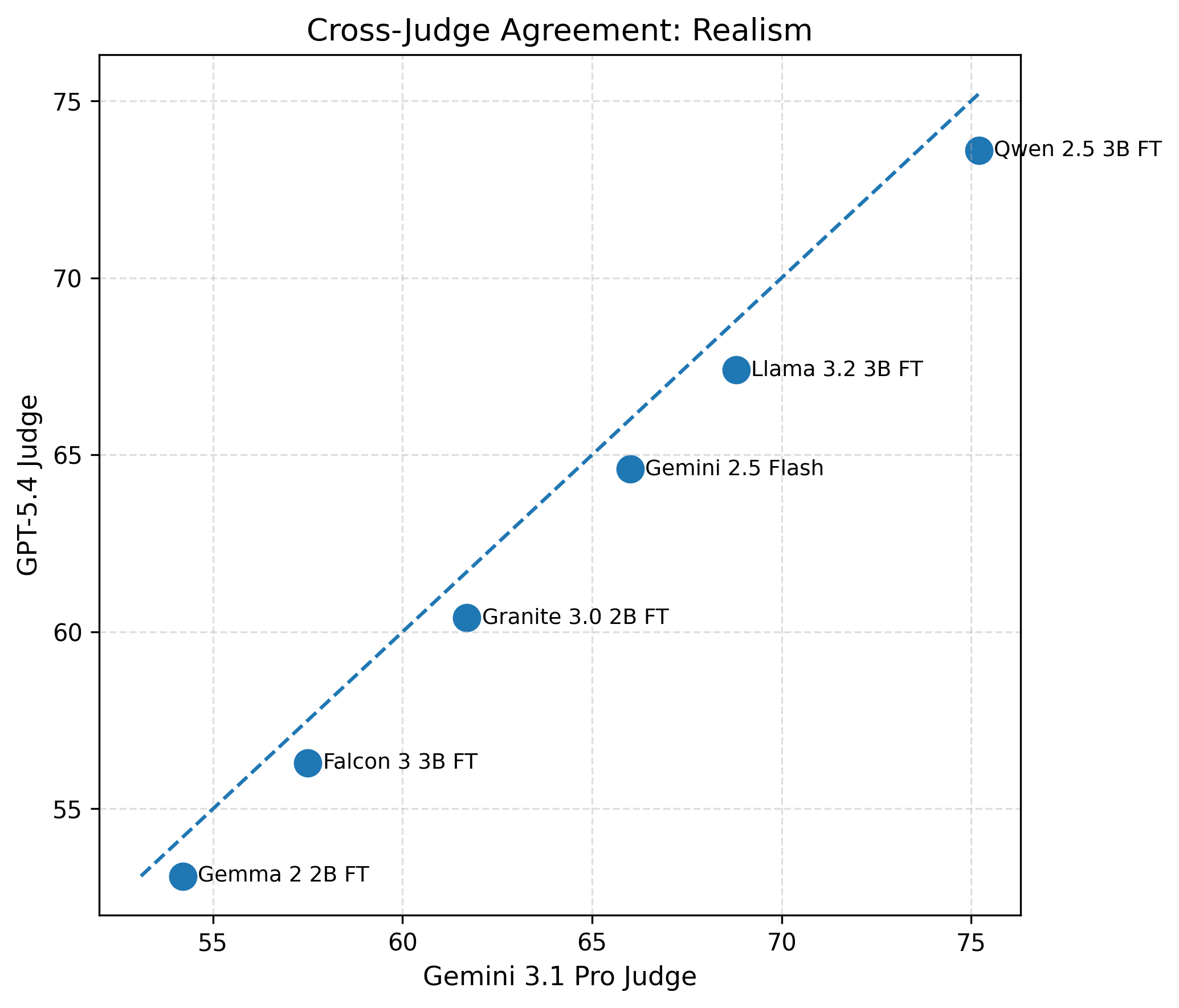}
        \caption{Realism}
        \label{fig:crossjudge_realism}
    \end{subfigure}
    \hfill
    \begin{subfigure}[t]{0.48\textwidth}
        \centering
        \includegraphics[width=\textwidth]{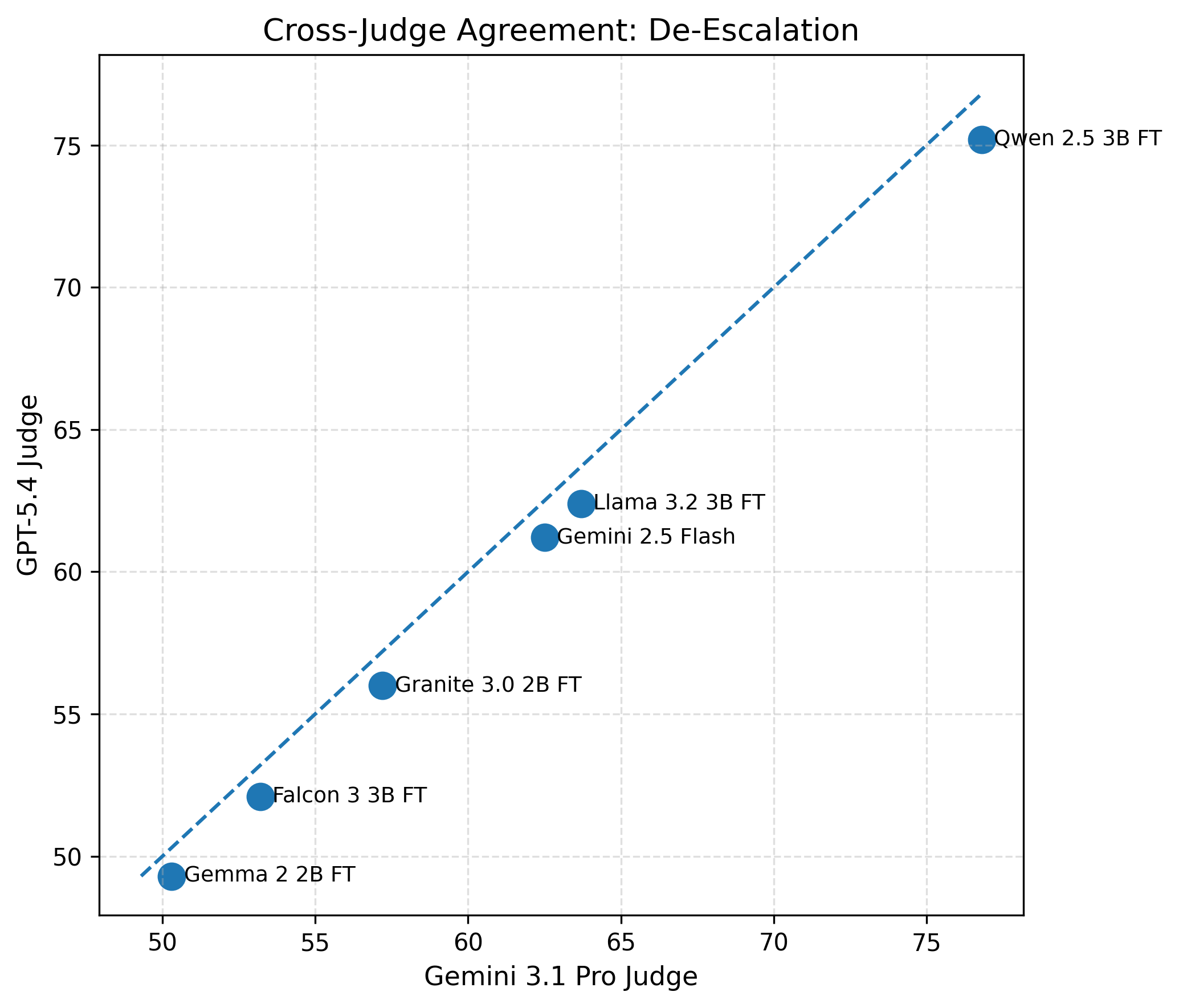}
        \caption{De-escalation}
        \label{fig:crossjudge_deescalation}
    \end{subfigure}

    \caption{
    \textbf{Cross-judge agreement between Gemini~3.1 Pro and GPT-5.4.}
    Scatter plots compare realism and de-escalation scores assigned by two
    independent LLM judges. Each point corresponds to either a fine-tuned
    open-weight model or the Gemini~2.5 Flash baseline. The dashed diagonal
    line indicates perfect agreement between judges. Points close to the
    diagonal show that both judges produce consistent score trends, suggesting
    that the observed ranking is not driven by a single evaluator. The
    fine-tuned Qwen~2.5 model achieves the strongest overall performance,
    while Gemini~2.5 Flash provides a proprietary baseline reference.
    }
    \label{fig:crossjudge_agreement}
\end{figure}

\onecolumn 
\section{Qualitative Examples}
\label{appendix:qualitative_examples}
To provide qualitative insight into the linguistic complexity and diversity of the \tool benchmark, we present three representative transcripts derived from our ``in-the-wild'' video corpus. These examples illustrate the distinct conversational challenges our SLMs are fine-tuned to navigate, ranging from high-stakes mental health crises to complex procedural negotiations. Table \ref{tab:full_scenario188} (Scenario 188) depicts a volatile interaction with a subject experiencing delusions, requiring the officer to balance legal enforcement with de-escalation strategies. Table \ref{tab:abridged_scenario371} (Scenario 371) highlights administrative ambiguity during a traffic stop, testing the model's ability to track complex entities like insurance and ownership over long contexts. Finally, Table \ref{tab:abridged_scenario803} (Scenario 803) presents an adversarial interrogation involving a subject impersonating an officer, where the model must detect inconsistencies in a deceptive narrative. For clarity, speaker roles are color-coded: \textcolor{policeblue}{\textbf{Law Enforcement}} (Blue), \textcolor{civilianred}{\textbf{Subject}} (Orange), \textcolor{dispatchgray}{\textbf{Dispatch}} (Pink), and \textcolor{othergreen}{\textbf{Third Parties}} (Green). Note that repetitive segments in the latter two examples have been abridged to fit page constraints.


\definecolor{policeblue}{RGB}{0, 0, 205}
\definecolor{civilianred}{RGB}{255, 69, 0}
\definecolor{dispatchgray}{RGB}{255, 20, 147}

\begin{longtable}{p{0.10\textwidth} p{0.20\textwidth} p{0.65\textwidth}}
    
    \toprule
    \textbf{Time} & \textbf{Speaker} & \textbf{Transcript} \\
    \midrule
    \endfirsthead

    \multicolumn{3}{c}{{\bfseries \tablename\ \thetable{} -- continued from previous page}} \\
    \toprule
    \textbf{Time} & \textbf{Speaker} & \textbf{Transcript} \\
    \midrule
    \endhead

    \midrule
    \multicolumn{3}{r}{{Continued on next page...}} \\
    \bottomrule
    \endfoot

    \bottomrule
    \noalign{\vspace{3ex}} 
    \caption{\textbf{Full Transcript of Scenario 188.} An officer interacts with a trespassing subject experiencing delusions. Colors indicate speaker roles: \textcolor{policeblue}{Police}, \textcolor{civilianred}{Subject}, \textcolor{dispatchgray}{Dispatch}.} 
    \label{tab:full_scenario188} \\
    \endlastfoot

    0:00 & \textcolor{policeblue}{Deputy Warney} & \textcolor{policeblue}{I'm trying to walk up this long driveway to the house now.} \\
    0:02 & \textcolor{dispatchgray}{[Dispatcher]} & \textcolor{dispatchgray}{He advised that the woman was in the pool and he is coming to let y'all in the gate now.} \\
    0:07 & \textcolor{policeblue}{Deputy Warney} & \textcolor{policeblue}{That's 10-4. I'm at the house now.} \\
    0:09 & \textcolor{policeblue}{[Law enforcement officer]} & \textcolor{policeblue}{Is this that lady that was out here a while back that says she was pregnant with his baby or she} \\
    0:13 & \textcolor{policeblue}{[Law enforcement officer]} & \textcolor{policeblue}{pulling me freaking amaze.} \\
    0:15 & \textcolor{policeblue}{[Law enforcement officer]} & \textcolor{policeblue}{Well, that used to be a baseball field.} \\
    0:16 & \textcolor{policeblue}{[Law enforcement officer]} & \textcolor{policeblue}{She's actually in the pool, apparently.} \\
    0:18 & \textcolor{policeblue}{Deputy Warney} & \textcolor{policeblue}{That's a huge pool.} \\
    0:19 & \textcolor{policeblue}{[Law enforcement officer]} & \textcolor{policeblue}{It's like an Olympic size pool.} \\
    0:21 & \textcolor{civilianred}{Subject} & \textcolor{civilianred}{I was at the gate and I came to perform for a photo shoot and no Rick Ross.} \\
    0:25 & \textcolor{policeblue}{[Law enforcement officer]} & \textcolor{policeblue}{Everybody knows.} \\
    0:26 & \textcolor{policeblue}{[Law enforcement officer]} & \textcolor{policeblue}{I know who Rick Ross is.} \\
    0:27 & \textcolor{policeblue}{Deputy Warney} & \textcolor{policeblue}{Is your ID in here?} \\
    0:28 & \textcolor{civilianred}{Subject} & \textcolor{civilianred}{Yes, sir, it is.} \\
    0:29 & \textcolor{policeblue}{Deputy Warney} & \textcolor{policeblue}{Where's it at?} \\
    0:30 & \textcolor{policeblue}{[Law enforcement officer]} & \textcolor{policeblue}{Just tell me that's fine.} \\
    0:31 & \textcolor{policeblue}{Deputy Warney} & \textcolor{policeblue}{Ma'am, you're handcuffed.} \\
    0:31 & \textcolor{policeblue}{[Law enforcement officer]} & \textcolor{policeblue}{He can probably give you a ride on that thing maybe.} \\
    0:33 & \textcolor{policeblue}{[Law enforcement officer]} & \textcolor{policeblue}{Married to Rick Ross.} \\
    
    \multicolumn{3}{c}{\textit{[... 17 minutes of circular conversation between Subject and officer ...]}} \\
    
    17:37 & \textcolor{civilianred}{Subject} & \textcolor{civilianred}{Everybody knows me. Everybody know I'm Rick Ross's wife. I'm Rick Ross's wife. Who are you? You're not a policeman. It's not a policeman do and I'm Rick Ross's wife.} \\
    17:47 & \textcolor{policeblue}{Deputy Warney} & \textcolor{policeblue}{This isn't court. What I'm doing is just reading you the warrants that I have, okay? I'm not asking you if you're guilty or innocent. I'm just reading you what I have and then I'm going to ask you if you understand. Just a yes or no, okay?} \\
    17:56 & \textcolor{civilianred}{Subject} & \textcolor{civilianred}{Okay.} \\
    17:57 & \textcolor{policeblue}{Deputy Warney} & \textcolor{policeblue}{All right, first one is for the criminal trespass, okay? Cuz they told you they didn't want you on the property, but you trespass on there anyways, okay?} \\
    18:03 & \textcolor{policeblue}{Deputy Warney} & \textcolor{policeblue}{Second one is for the possession of marijuana that you had in your purse, okay? That's what you're being charged with. So possession of marijuana less than an ounce and criminal trespass. Do you understand what your charges are?} \\
    18:14 & \textcolor{civilianred}{Subject} & \textcolor{civilianred}{Yes.} \\
    18:14 & \textcolor{policeblue}{Deputy Warney} & \textcolor{policeblue}{Okay. All right, thank you, ma'am.} \\

\end{longtable}

\begin{longtable}{p{0.08\textwidth} p{0.20\textwidth} p{0.65\textwidth}}
    
    \toprule
    \textbf{Time} & \textbf{Speaker} & \textbf{Transcript} \\
    \midrule
    \endfirsthead

    \multicolumn{3}{c}{{\bfseries \tablename\ \thetable{} -- continued from previous page}} \\
    \toprule
    \textbf{Time} & \textbf{Speaker} & \textbf{Transcript} \\
    \midrule
    \endhead

    \midrule
    \multicolumn{3}{r}{{Continued on next page...}} \\
    \bottomrule
    \endfoot

    \bottomrule
    \noalign{\vspace{3ex}} 
    \caption{\textbf{Full Transcript of Scenario 371.} A traffic stop for a stop sign violation involving complex insurance verification. The middle portion is omitted for brevity. Colors indicate speaker roles: \textcolor{policeblue}{Police}, \textcolor{civilianred}{Subject}, \textcolor{dispatchgray}{Dispatch}, \textcolor{othergreen}{Other}.} 
    \label{tab:abridged_scenario371} \\
    \endlastfoot

    00:30 & \textcolor{policeblue}{[Police Officer]} & \textcolor{policeblue}{How you doing, sir?} \\
    00:31 & \textcolor{civilianred}{Subject} & \textcolor{civilianred}{I'm good, right here, sir.} \\
    00:33 & \textcolor{policeblue}{[Police Officer]} & \textcolor{policeblue}{Oh, okay. How's your day going?} \\
    00:35 & \textcolor{civilianred}{Subject} & \textcolor{civilianred}{Not bad. It's good. Welcome to Publix.} \\
    00:37 & \textcolor{policeblue}{[Police Officer]} & \textcolor{policeblue}{Understand. But you know I'm pulling you over, right? You've come to a police stop on that stop sign.} \\
    00:42 & \textcolor{civilianred}{Subject} & \textcolor{civilianred}{I thought I paused enough, but if I didn't, I just kept seeing you roll.} \\
    00:45 & \textcolor{policeblue}{[Police Officer]} & \textcolor{policeblue}{That's okay. Sometimes we don't realize we think we're doing something. Um, is this vehicle in your name?} \\
    00:52 & \textcolor{civilianred}{Subject} & \textcolor{civilianred}{Uh, yes. Subject and also I it's co-signed by Co-signe.} \\
    00:58 & \textcolor{policeblue}{[Police Officer]} & \textcolor{policeblue}{Co-signer? Okay. Did you just make a transaction?} \\
    01:01 & \textcolor{civilianred}{Subject} & \textcolor{civilianred}{Did I just make a transaction? Did you just buy the vehicle?} \\
    01:03 & \textcolor{policeblue}{[Police Officer]} & \textcolor{policeblue}{Six months ago.} \\
    01:04 & \textcolor{civilianred}{Subject} & \textcolor{civilianred}{Six months ago? Okay. All right. Uh, can you do me a favor, Subject? Do you uh, can you come up with some proof of insurance for me real quick?} \\
    01:10 & \textcolor{policeblue}{[Police Officer]} & \textcolor{policeblue}{Uh, I don't have the papers in here.} \\
    01:14 & \textcolor{civilianred}{Subject} & \textcolor{civilianred}{Okay. It might be upstairs. Not sure, sir.} \\
    01:20 & \textcolor{policeblue}{[Police Officer]} & \textcolor{policeblue}{Okay. Give me one sec. I'll be right back.} \\
    01:36 & \textcolor{othergreen}{Third Parties} & \textcolor{othergreen}{Hey, real quick, uh, I'm up here right out in front of the apartment complex with this guy. Um, NCIC says he doesn't have any insurance valid at the time.} \\
    01:54 & \textcolor{othergreen}{Third Parties} & \textcolor{othergreen}{Okay, I'm out in front of a parking or apartment complex. This guy lives here. He just ran a stop sign, so I pulled him over.} \\
    02:17 & \textcolor{dispatchgray}{[Police Dispatcher]} & \textcolor{dispatchgray}{It just says no valid insurance.} \\
    02:20 & \textcolor{othergreen}{Third Parties} & \textcolor{othergreen}{But this has been six months. I don't know. So just let him deal with it at court.} \\
    02:25 & \textcolor{dispatchgray}{[Police Dispatcher]} & \textcolor{dispatchgray}{If he does.} \\
    02:28 & \textcolor{othergreen}{Third Parties} & \textcolor{othergreen}{Yeah.} \\
    02:30 & \textcolor{dispatchgray}{[Police Dispatcher]} & \textcolor{dispatchgray}{Okay. Let me jump on it. Figure out who's got. All right, thanks.} \\
    02:35 & \textcolor{policeblue}{[Police Officer]} & \textcolor{policeblue}{Mr. Subject. Who do you uh, have insurance through right now, sir?} \\
    02:39 & \textcolor{civilianred}{Subject} & \textcolor{civilianred}{Uh, it was he had I looked out it was going through Geico. That was the last one. That was the last Geico.} \\
    02:45 & \textcolor{policeblue}{[Police Officer]} & \textcolor{policeblue}{He's got it through Geico?} \\
    2:46 &	\textcolor{civilianred}{Subject} &\textcolor{civilianred}{	Pretty sure he does, so yeah. So, I don't even drive the car. I just took it across the street to get my groceries because I couldn't even walk. Like, you know what I'm saying? The car was parked there. Like, you know what I'm saying?}\\
    
    \multicolumn{3}{c}{\textit{[... approximately 9 minutes of insurance verification and discussion ...]}} \\
    
    11:57 & \textcolor{policeblue}{[Police Officer]} & \textcolor{policeblue}{No, stand by. I need you here to witness this, sir.} \\
    12:02 & \textcolor{civilianred}{Subject} & \textcolor{civilianred}{That way you're comfortable and I'm comfortable, okay? That's it.} \\
    12:05 & \textcolor{policeblue}{[Police Officer]} & \textcolor{policeblue}{2018.} \\
    12:07 & \textcolor{civilianred}{Subject} & \textcolor{civilianred}{2018.} \\
    12:10 & \textcolor{policeblue}{[Police Officer]} & \textcolor{policeblue}{Yep, that's what it is.} \\
    12:19 & \textcolor{policeblue}{[Police Officer]} & \textcolor{policeblue}{Also keep this with you, okay? So when you go grab the vehicle and everything, all right?} \\
    12:23 & \textcolor{civilianred}{Subject} & \textcolor{civilianred}{Okay, this is what I need to grab the vehicle. Yes, sir. Okay. All right, thank you.} \\
    12:26 & \textcolor{policeblue}{[Police Officer]} & \textcolor{policeblue}{All right. All right. You have any more questions?} \\
    12:27 & \textcolor{civilianred}{Subject} & \textcolor{civilianred}{No.} \\
    12:28 & \textcolor{policeblue}{[Police Officer]} & \textcolor{policeblue}{All right. You have a good night.} \\

\end{longtable}

\begin{longtable}{p{0.08\textwidth} p{0.20\textwidth} p{0.65\textwidth}}
    
    \toprule
    \textbf{Time} & \textbf{Speaker} & \textbf{Transcript} \\
    \midrule
    \endfirsthead

    \bottomrule
    \noalign{\vspace{3ex}} 
    \caption{\textbf{Full Transcript of Scenario 803.} A subject (Subject) is stopped for impersonating a police officer. He initially claims the emergency lights are disconnected, but the interrogation reveals otherwise. Colors indicate speaker roles: \textcolor{policeblue}{Police}, \textcolor{civilianred}{Subject}, \textcolor{othergreen}{Third Parties}.} 
    \label{tab:abridged_scenario803} \\
    \endlastfoot

    00:00 & \textcolor{policeblue}{Officer} & \textcolor{policeblue}{Can be honest with me. Okay? Honesty, honesty, honesty's going to go a long way.} \\
    00:04 & \textcolor{civilianred}{Subject} & \textcolor{civilianred}{I know I can't get in trouble.} \\
    00:06 & \textcolor{policeblue}{Officer} & \textcolor{policeblue}{Honesty's going to go a long way.} \\
    00:08 & \textcolor{civilianred}{Subject} & \textcolor{civilianred}{I, right? I, I really don't want to get in trouble. I will rip anything out.} \\
    00:12 & \textcolor{policeblue}{Officer} & \textcolor{policeblue}{Why are you crying?} \\
    00:13 & \textcolor{othergreen}{Third Parties} & \textcolor{othergreen}{For months, officers have kept an eye on this teen's suspicious looking car, a near perfect fake squad vehicle that's fooled plenty. But tonight, things take a sharp turn when he's finally caught in the act, and there's no way out this time.} \\
    00:28 & \textcolor{policeblue}{Officer} & \textcolor{policeblue}{Why are you crying?} \\
    00:29 & \textcolor{civilianred}{Subject} & \textcolor{civilianred}{My life's been shit lately. I don't know.} \\
    00:33 & \textcolor{civilianred}{Subject} & \textcolor{civilianred}{Every, every odds.} \\
    00:34 & \textcolor{othergreen}{Third Parties} & \textcolor{othergreen}{Around 12:43 a.m. on June 1st, 2025, an Oclair police officer received a report from a caller who claimed a squad car with 50 written on the side had switched on its red and blue lights to make a U-turn in an area in City, State.} \\
    00:51 & \textcolor{policeblue}{Officer} & \textcolor{policeblue}{Whose car is this?} \\
    00:52 & \textcolor{civilianred}{Subject} & \textcolor{civilianred}{Mine.} \\
    00:52 & \textcolor{policeblue}{Officer} & \textcolor{policeblue}{Your car?} \\
    00:53 & \textcolor{civilianred}{Subject} & \textcolor{civilianred}{Yeah.} \\
    00:54 & \textcolor{policeblue}{Officer} & \textcolor{policeblue}{Talk to me.} \\
    00:55 & \textcolor{policeblue}{Officer} & \textcolor{policeblue}{This thing got red and blues on it?} \\
    00:57 & \textcolor{civilianred}{Subject} & \textcolor{civilianred}{No, they're all disconnected.} \\
    00:58 & \textcolor{policeblue}{Officer} & \textcolor{policeblue}{Okay.} \\
    00:59 & \textcolor{civilianred}{Subject} & \textcolor{civilianred}{Every time I they uh private property, all disconnected.} \\
    01:02 & \textcolor{policeblue}{Officer} & \textcolor{policeblue}{Okay. So how come they were on back there?} \\
    \multicolumn{3}{r}{{Continued on next page...}} \\
    01:06 & \textcolor{civilianred}{Subject} & \textcolor{civilianred}{I'm miss you. I'm miss you.} \\
    01:07 & \textcolor{civilianred}{Subject} & \textcolor{civilianred}{I don't know.} \\
    01:09 & \textcolor{civilianred}{Subject} & \textcolor{civilianred}{I can't.} \\
    01:11 & \textcolor{policeblue}{Officer} & \textcolor{policeblue}{Can I look at it?} \\
    01:12 & \textcolor{civilianred}{Subject} & \textcolor{civilianred}{Go for it.} \\
    1:12 &	\textcolor{policeblue}{Officer} &	\textcolor{policeblue}{Is that cool?}\\
    1:14 &	\textcolor{civilianred}{Subject} & 	\textcolor{civilianred}{Everything's disconnected. All I have is this, even if I}\\
    1:18 &	\textcolor{policeblue}{Officer} &	\textcolor{policeblue}{What if you turn, can you turn the car on for me?}\\
    1:22 &	\textcolor{policeblue}{Officer} &	\textcolor{policeblue}{You do all this to yourself?}\\
    1:23 & 	\textcolor{civilianred}{Subject} &	\textcolor{civilianred}{Yeah.}\\
    1:24 &	\textcolor{policeblue}{Officer} &	\textcolor{policeblue}{Do you buy this thing like a squad car?}\\
    1:25 &	\textcolor{civilianred}{Subject} &	\textcolor{civilianred}{No.}\\
    1:26 &	\textcolor{policeblue}{Officer} &	\textcolor{policeblue}{You just did it all?}\\
    1:27 & \textcolor{civilianred}{Subject} &	\textcolor{civilianred}{Yeah.}\\
    \multicolumn{3}{c}{\textit{[... approximately 23 minutes of interrogation regarding the fake police equipment ...]}} \\
    
    23:50 & \textcolor{policeblue}{Officer} & \textcolor{policeblue}{Okay. Well, you're lying. You're you started lying to me from the start. Here's the story you get if you want.} \\
    23:54 & \textcolor{civilianred}{Subject} & \textcolor{civilianred}{I don't need it again.} \\
    23:55 & \textcolor{policeblue}{Officer} & \textcolor{policeblue}{I know. All right. You got any questions for me?} \\
    23:57 & \textcolor{civilianred}{Subject} & \textcolor{civilianred}{Uh, oh, I don't know.} \\
    24:00 & \textcolor{policeblue}{Officer} & \textcolor{policeblue}{Okay. All right, man. You're good.} \\
    24:02 & \textcolor{civilianred}{Subject} & \textcolor{civilianred}{So where do I go? That courthouse up there?} \\
    24:04 & \textcolor{policeblue}{Officer} & \textcolor{policeblue}{Yeah, it's it's on there. All right, man. That's all I got for you.} \\

\end{longtable}

\section{Conceptual System Overview}

\begin{figure}[htbp]
    \centering
    \includegraphics[width=\textwidth]{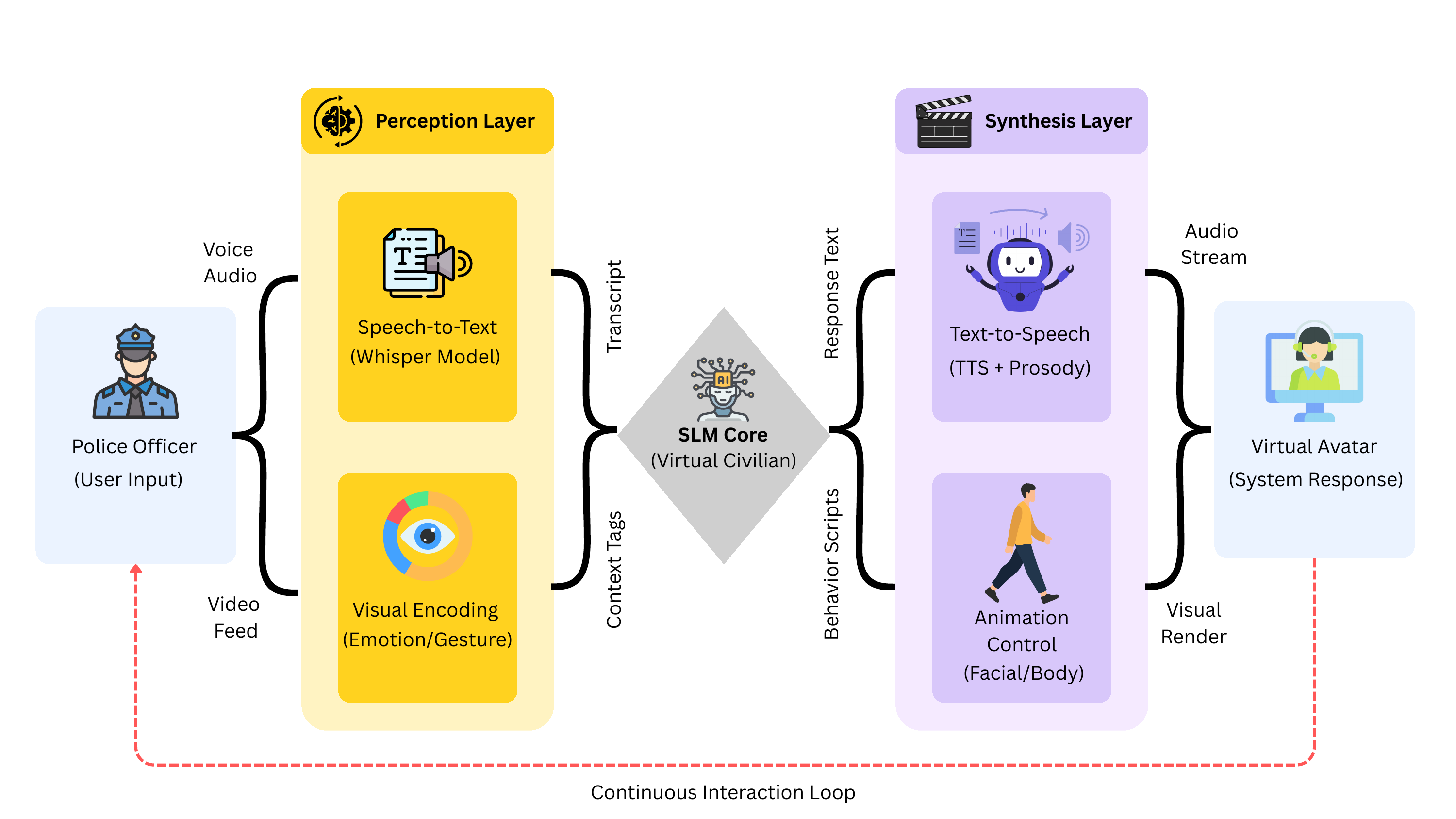}
    \caption{\textbf{Conceptual architecture of a multimodal virtual de-escalation system.} The pipeline operates as a closed-loop system in which the \textbf{police officer's} multimodal input is processed by a \textbf{perception layer}, reasoned over by the specialized \textbf{SLM core}, and rendered through a \textbf{synthesis layer} to control a \textbf{virtual avatar} in real time.}
    \label{fig:de_escalation_pipeline}
\end{figure}

The primary focus of this work is the development and evaluation of the specialized SLM core for handling the complex reasoning required in police de-escalation scenarios. However, the broader motivation is to integrate such models into an immersive, multimodal virtual reality (VR) training environment. Figure~\ref{fig:de_escalation_pipeline} illustrates this conceptual closed-loop architecture. 

While the contributions of this paper focus on textual reasoning and generation, the proposed system can be extended with additional components. Specifically, a perception layer (e.g., visual and audio encoders) can process multimodal inputs, and a synthesis layer (e.g., text-to-speech and real-time avatar animation) can enable naturalistic, real-time interaction with a virtual avatar.


\section{Video Filtering Pipeline}
\label{app:filtering_pipeline}
We employ an LLM-based pipeline with human oversight to filter a domain-specific set of 1,500 videos from an initial pool of 5,000 raw videos collected from YouTube, TikTok, and Facebook. This process combines unsupervised clustering, feature selection, LLM-based reasoning, and deterministic rule-based filtering to systematically identify high-value de-escalation content.

\subsection{Transcription and Feature Discovery}
\label{app:feature_discovery}

\noindent \textbf{Step 1: Transcription.} We generated automatic speech recognition (ASR) transcripts for all 5,000 candidate videos using the OpenAI Whisper model~\cite{pmlr-v202-radford23a}. These transcripts serve as the sole input to all subsequent processing stages, enabling a fully automated and reproducible pipeline.

\noindent \textbf{Step 2: Clustering and schema definition.} To define the relevant feature space without manual annotation, we embedded the raw transcripts using a sentence-level encoder and applied HDBSCAN clustering. This unsupervised analysis revealed three distinct content clusters. By examining the semantic centroids of these clusters, we derived a taxonomy of 30 binary features designed to identify high-value de-escalation scenarios. The resulting feature schema $F$ is organized into five categories, detailed below and summarized in Figure~\ref{fig:feature_schema}.

\begin{enumerate} 
 \item \textbf{Police presence} ($S_{police}$): Indicators confirming official law enforcement involvement. \\ \fbox{ \begin{minipage}{0.85\linewidth} \small \ttfamily police\_identified\_by\_role, police\_commands\_present, id\_request\_event, legal\_procedure\_language, police\_explaining\_actions, civilian\_questioning\_police \end{minipage} }

 \item \textbf{Interaction type} ($S_{interact}$): Features characterizing the nature and structure of the dialogue. \\ \fbox{ \begin{minipage}{0.85\linewidth} \small \ttfamily conversation\_present, verbal\_disagreement\_detected, compliance\_discussion, instruction\_clarification\_dialogue, negotiation\_attempts, conflict\_resolution\_attempt \end{minipage} }

 \item \textbf{Escalation} ($S_{esc}$): Signals indicative of rising tension or conflict. \\ \fbox{ \begin{minipage}{0.85\linewidth} \small \ttfamily raised\_threat\_language, noncompliance\_resistance\_language, accusatory\_language, police\_warning\_language, emotional\_intensity\_spike, crowd\_escalation\_factors, use\_of\_force\_transition\_signals \end{minipage} }

 \item \textbf{De-escalation} ($S_{deesc}$): Signals indicative of active conflict mitigation. \\ \fbox{ \begin{minipage}{0.85\linewidth} \small \ttfamily calming\_language\_used\_by\_officer, empathetic\_statements, clear\_explanations\_given, tone\_softening\_cues, conflict\_deescalation\_success, agreement\_reached \end{minipage} }

 \item \textbf{Context filters} ($S_{noise}$): Exclusion criteria used to remove off-domain or low-quality content. \\ \fbox{ \begin{minipage}{0.85\linewidth} \small \ttfamily no\_police\_presence, no\_conversation\_detected, advertisement\_content\_detected, training\_range\_context, non\_relevant\_crime\_only\_context \end{minipage} } \end{enumerate}

\begin{figure}[h]
    \centering
    \begin{lrbox}{\codebox}
        \begin{minipage}{0.95\linewidth}
            \vspace{0.5em}
            \textbf{FEATURE\_SCHEMA = }
            \footnotesize
\begin{verbatim}
{
    "police_presence_signals": [
        "police_identified_by_role",
        "police_commands_present",
        "id_request_event",
        "legal_procedure_language",
        "police_explaining_actions",
        "civilian_questioning_police"
    ],
    "interaction_type_signals": [
        "conversation_present",
        "verbal_disagreement_detected",
        "compliance_discussion",
        "instruction_clarification_dialogue",
        "negotiation_attempts",
        "conflict_resolution_attempt"
    ],
    "escalation_indicators": [
        "raised_threat_language",
        "noncompliance_resistance_language",
        "accusatory_language",
        "police_warning_language",
        "emotional_intensity_spike",
        "crowd_escalation_factors",
        "use_of_force_transition_signals"
    ],
    "deescalation_indicators": [
        "calming_language_used_by_officer",
        "empathetic_statements",
        "clear_explanations_given",
        "tone_softening_cues",
        "conflict_deescalation_success",
        "agreement_reached"
    ],
    "context_filters": [
        "no_police_presence",
        "no_conversation_detected",
        "advertisement_content_detected",
        "training_range_context",
        "non_relevant_crime_only_context"
    ]
}
\end{verbatim}
        \end{minipage}
    \end{lrbox}
    \fbox{\usebox{\codebox}}
    \caption{Full feature schema taxonomy used for transcript-level 
    filtering. Each key corresponds to one of the five feature 
    categories; values are the 30 binary signal names extracted 
    per transcript by the LLM annotator.}
    \label{fig:feature_schema}
\end{figure}

\subsection{LLM-Based Feature Extraction}
\label{app:llm_extraction}

\noindent \textbf{Step 3: Annotation.} We utilized an LLM to map every transcript to the feature schema defined in Section~\ref{app:feature_discovery}. The model was instructed to output a structured binary vector corresponding to the presence (1) or absence (0) of each of the 30 signals. To maximize annotation consistency, we enforced strict JSON output formatting and prohibited the model from making inferences when transcript evidence was ambiguous. The exact zero-shot prompt used for inference is provided in Figure~\ref{fig:prompt}.

\begin{figure}[h]
\centering
\begin{tcolorbox}[colback=gray!5!white, colframe=gray!75!black, title=Feature Extraction Prompt, fonttitle=\bfseries]
{[SYSTEM]}\\
You are a forensic annotation specialist trained in law enforcement
interaction analysis. Your sole task is to analyze the provided
transcript and return a structured binary feature vector. You must
evaluate evidence strictly from the transcript text.
\\\\
{[DEFINITIONS]}\\
- Present (1): The feature is explicitly evidenced by specific
  words, phrases, or speech acts in the transcript.\\
- Absent (0): The feature is not evidenced, ambiguous, or cannot
  be inferred without speculation.
\\\\
{[INPUT]}\\
Transcript: \{transcript\_file\}
\\\\
{[SCHEMA]}\\
Return a JSON object using EXACTLY the following nested structure.
Do not add, remove, or rename any keys:\\
\{feature\_schema\}
\\\\
{[RULES]}\\
1. Evaluate each of the 30 binary features independently.\\
2. Assign 1 only when direct textual evidence supports presence.\\
3. Assign 0 for all ambiguous, implied, or unverifiable cases.\\
4. Do not hallucinate features based on assumed context.\\
5. Do not output explanations, comments, or markdown formatting.\\
6. Do not wrap output in code blocks or backticks.\\
7. Output ONLY the raw JSON object. Any additional text will
   invalidate the response.
\\\\
{[OUTPUT FORMAT]}\\
\{\\
\hspace*{1em}``police\_presence\_signals'':  \{ \ldots \},\\
\hspace*{1em}``interaction\_type\_signals'': \{ \ldots \},\\
\hspace*{1em}``escalation\_indicators'':    \{ \ldots \},\\
\hspace*{1em}``deescalation\_indicators'':  \{ \ldots \},\\
\hspace*{1em}``context\_filters'':          \{ \ldots \}\\
\}
\end{tcolorbox}
\caption{Zero-shot prompt used for LLM-based feature 
extraction.}
\label{fig:prompt}
\end{figure}

\subsection{Rule-Based Filtering Logic}
\label{app:filtering_logic}

\noindent \textbf{Step 4: Filtering criteria.} Let $c(S)$ denote the count of active features within category $S$, computed as the sum of binary indicator values. A video $v$ is retained in the dataset if and only if it satisfies the composite validity condition $C_{valid}(v)$, defined as follows:

\begin{equation}
    C_{valid}(v) = 
    \underbrace{(c(S_{noise}) = 0)}_{\text{Context}} \land 
    \underbrace{(c(S_{police}) \geq 2)}_{\text{Relevance}} \land 
    \underbrace{\left( c(S_{esc}) \geq 3 \lor 
    c(S_{deesc}) \geq 3 \right)}_{\text{Intensity}}
    \label{eq:filtering}
\end{equation}

The three constituent conditions each enforce a distinct quality criterion:

\begin{itemize}[leftmargin=*, topsep=0pt, noitemsep] \item \textbf{Context} $(c(S_{noise}) = 0)$: Excludes videos flagged as advertisements, commentary, or otherwise off-domain content that would introduce noise into the training corpus.

 \item \textbf{Relevance} $(c(S_{police}) \geq 2)$: Requires at least two independent signals confirming active law enforcement participation, guarding against false positives from tangential mentions of policing.

 \item \textbf{Intensity} $(c(S_{esc}) \geq 3 \lor c(S_{deesc}) \geq 3)$: Requires the interaction to exhibit substantial depth in either escalation or de-escalation dynamics, ensuring that retained videos contain the high-stakes exchanges necessary for effective model training. \end{itemize}

Videos satisfying all three conditions are forwarded to the diarization and quality validation stages described in Appendix~\ref{app:diarization}.

\subsection{Human Verification} \label{app:human_verification}

To rigorously validate the precision of the automated retrieval pipeline and establish a high-quality ground truth for the \tool benchmark, we conducted a manual dual-annotation study. We randomly sampled $N = 100$ videos from the final filtered candidate set of 1,500. Each video is approximately 18 minutes long, yielding over 30 hours of total review content. Given that careful evaluation of long-form law enforcement footage typically requires substantially more than real-time playback, this annotation effort represents a significant commitment of expert time. We consider this sample size sufficient for reliable pipeline validation, consistent with established practice in dataset quality assurance studies~\cite{rosas2025constructing}. Two independent expert annotators evaluated the sampled videos to assess alignment with the intended de-escalation use case, verifying that the retrieval pipeline effectively preserves high-intensity police-civilian interactions while filtering irrelevant content.

\noindent\textbf{Annotation protocol.} For each video in the sample, annotators independently assessed a set of binary inclusion criteria and qualitative categories:

\begin{itemize}[leftmargin=*, topsep=2pt, noitemsep] 
\item \textbf{Human Context Valid and Relevance Valid:} Binary indicators verifying the presence of a genuine, real-world police interaction (Context Valid) and clear role separation between officers and civilians (Relevance Valid).

 \item \textbf{Human intensity valid:} A binary indicator confirming the presence of genuine tension escalation or substantive de-escalation attempts within the interaction.

 \item \textbf{False positive category:} A categorical label documenting the specific failure mode for videos that did not pass the prior checks, such as news broadcast commentary or excessive wind noise.

 \item \textbf{Audio quality score:} An ordinal rating of speech clarity on a 1-to-5 scale, where 1 denotes completely unintelligible audio and 5 denotes broadcast-quality clarity.

 \item \textbf{Scenario category:} A multi-class categorization of the event type across 10 distinct tension-level classes, as defined in the taxonomy of Appendix~\ref{app:categorical_annotation}.

 \item \textbf{Final keep decision:} The ultimate binary verdict on whether the video should be retained in the final dataset. Annotators also recorded free-text qualitative notes to capture nuanced behavioral observations not captured by structured labels. \end{itemize}

\noindent\textbf{Agreement statistics and quality assurance.} To quantify the reliability of the human verification process, we computed both raw proportion agreement and Cohen's Kappa ($\kappa$) across all annotation dimensions. The results, summarized in Table~\ref{tab:iaa_results}, demonstrate high pipeline precision and robust inter-annotator alignment.

The baseline filtering criteria exhibited near-perfect alignment. Both Human Context Valid and Human Relevance Valid achieved 100\% raw agreement. Because the automated pipeline effectively removed irrelevant content prior to this stage, annotators unanimously voted to retain all sampled videos, resulting in zero label variance. Cohen's Kappa is therefore mathematically undefined for these variables, which constitutes strong evidence of the pipeline's high true-positive retrieval rate rather than a limitation of the evaluation.

For the Final Keep Decision, annotators achieved 90.9\% raw agreement with perfect chance-adjusted agreement ($\kappa = 1.000$), demonstrating that the inclusion criteria were sufficiently well-specified to support consistent expert judgment. The Human Intensity Valid check yielded substantial agreement (90.9\%, $\kappa = 0.621$), confirming that trained annotators can reliably identify genuine de-escalation dynamics in uncontrolled footage. The ordinal Audio Quality Score also achieved substantial agreement ($\kappa = 0.718$), reflecting a well-calibrated shared threshold for the acoustic clarity required for downstream language model training.

\noindent\textbf{Subjectivity and adjudication.} Categorizing real-world police interactions into 10 fine-grained taxonomy classes proved inherently subjective, yielding only fair initial agreement ($\kappa = 0.290$, 45.5\% raw agreement) for the Scenario Category dimension. This outcome is expected: real-world encounters frequently evolve across multiple tension states within a single interaction, for example a routine traffic stop that escalates into a verbal conflict, producing principled divergence between annotators rather than labeling error.

To establish a definitive ground truth, all scenario classification disagreements and any divergence in the Final Keep Decision were resolved through structured consensus discussion with a senior domain expert. This adjudication protocol ensures that ambiguous cases are resolved consistently and that the final benchmark labels reflect expert-level interpretation of complex, dynamic interactions.

\begin{table}[htbp]
    \centering
    \renewcommand{\arraystretch}{1.3}
    \begin{tabularx}{\textwidth}{@{} l c c X @{}}
        \toprule
        \textbf{Annotation Category} & 
        \textbf{Agreement} & 
        \textbf{Kappa ($\kappa$)} & 
        \textbf{Observations and Methodology Notes} \\
        \midrule

        Human Context Valid & 
        100.0\% & Undefined$^{*}$ & 
        Perfect raw agreement. Undefined $\kappa$ due to zero 
        label variance from unanimous inclusion. Indicates 
        highly effective automated pre-filtering. \\

        Human Relevance Valid & 
        100.0\% & Undefined$^{*}$ & 
        Perfect raw agreement on the presence of police-civilian 
        role separation. Undefined $\kappa$ due to unanimous 
        positive labels. \\

        Final Keep Decision & 
        90.9\% & 1.000 & 
        Perfect chance-adjusted agreement. Reflects 
        unambiguous inclusion criteria and consistent annotator 
        judgment. \\

        False Positive Category & 
        100.0\% & 1.000 & 
        Perfect agreement on classifying failure modes of 
        excluded videos, confirming clear categorical 
        definitions. \\

        Human Intensity Valid & 
        90.9\% & 0.621 & 
        Substantial agreement. Confirms that annotators 
        reliably recognize genuine escalation and de-escalation 
        dynamics in uncontrolled footage. \\

        Audio Quality Score & 
        63.6\% & 0.718 & 
        Substantial agreement on a subjective 1-to-5 ordinal 
        scale. Reflects a well-calibrated shared threshold for 
        acoustic viability. \\

        Scenario Category & 
        45.5\% & 0.290 & 
        Fair agreement, consistent with the inherent subjectivity 
        of categorizing evolving, multi-phase interactions into 
        10 fine-grained classes. All disagreements resolved via 
        expert adjudication. \\

        \bottomrule
    \end{tabularx}
    \vspace{4pt}
    \caption{Inter-annotator agreement (IAA) statistics for the 
    human verification study ($N = 100$ sampled videos). Raw 
    proportion agreement and Cohen's Kappa ($\kappa$) are 
    reported for each annotation dimension. Undefined $\kappa$ 
    values ($^{*}$) arise from unanimous annotator agreement, 
    which produces zero label variance and renders chance 
    correction inapplicable. All disagreements on Scenario 
    Category and Final Keep Decision were resolved through 
    consensus adjudication with a senior domain expert.}
    \label{tab:iaa_results}
\end{table}

\section{Speaker Diarization and Quality Assurance Pipeline} \label{app:diarization}

Following the filtering stage, the retained 1,500 videos are processed by Gemini~2.5 Flash~\cite{comanici2025gemini} for speaker diarization and transcript extraction. Because Gemini~2.5 Flash natively processes audio and video inputs, the model is prompted directly using the structured dual-task format described in Section~\ref{subsubsec:prompt_design}, bypassing the need for a separate ASR preprocessing step. Upon completing diarization, transcript quality is evaluated through manual assessment of the same randomly sampled subset of $N = 100$ videos used during the filtering validation stage (Appendix~\ref{app:filtering_pipeline}), ensuring consistency of the evaluation sample across both pipeline phases. The following sections detail the transcript generation methodology and the quantitative accuracy evaluation of the diarization outputs.

\subsection{Automated Diarization and Context Extraction Pipeline} \label{subsec:automated_diarization}

To process the raw, unformatted media files, we implemented an automated diarization and annotation pipeline utilizing Gemini~2.5 Flash~\cite{comanici2025gemini}. A primary advantage of this architecture is its native multimodal capability, which allows the model to directly ingest raw audio or video tokens alongside text instructions, eliminating the need for a separate intermediate ASR step and reducing the risk of cascading transcription errors.

\subsubsection{Structured Generation and Configuration} \label{subsubsec:structured_gen}

To ensure outputs are programmatically parseable and consistent for training downstream SLMs, we enforce a strict JSON schema during generation using Pydantic models. The model is configured with a decoding temperature of $\tau = 0.0$ and a top-$p$ of $0.95$. This low-temperature configuration restricts the sampling distribution, prioritizing deterministic and reproducible outputs when extracting ground-truth-style transcripts. All generated outputs are validated against the schema prior to storage; malformed responses are flagged and re-queried automatically.

\subsubsection{Dual-Task Prompting Strategy} \label{subsubsec:dual_task}

Rather than performing speaker diarization in isolation, our prompting framework instructs the model to execute two concurrent tasks within a single inference pass. This holistic approach allows the model to leverage its broader contextual understanding of the interaction to improve local diarization accuracy, as speaker role information inferred in Task~2 can resolve ambiguities in the turn-level attribution of Task~1. The pipeline extracts the following two components per video:

\begin{enumerate}[leftmargin=*, topsep=2pt, noitemsep] \item \textbf{Transcription and diarization (\texttt{task1\_transcripts}):} The model transcribes the dialogue verbatim, assigns a unique integer identifier to each distinct voice, and records the exact start timecode for every speech segment.

 \item \textbf{Speaker profiling (\texttt{task2\_speakers}):} Using both visual cues for video inputs and auditory context such as introductions and conversational tone, the model extracts metadata for each identified voice ID, including the speaker's inferred role within the interaction, such as officer or civilian. \end{enumerate}

\subsubsection{Prompt Design and Dynamic Context} \label{subsubsec:prompt_design}

To guide the model's generation, we utilize a concise and explicit set of system instructions. Because police interaction videos vary substantially in length, the prompt dynamically injects a timecode specification, defaulting to \texttt{MM:SS} and automatically adjusting to \texttt{H:MM:SS} for recordings exceeding one hour. The full prompt template is presented in Figure~\ref{fig:diarization_prompt}. This prompt maps directly to the output JSON schema, ensuring that the unstructured dialogue is tightly bound to the structured variables required for Word Error Rate (WER) and Diarization Error Rate (DER) calculations reported in Section~\ref{subsubsec:pipeline_quality}.

\begin{figure}[h]
\centering
\begin{tcolorbox}[colback=gray!5!white, colframe=gray!75!black, title=Speaker Diarization Prompt, fonttitle=\bfseries]
{[SYSTEM]}\\
You are an expert forensic transcriptionist and speaker
diarization specialist. Your task is to process the provided
media file and return a structured, verbatim transcript with
full speaker attribution. All output must be grounded strictly
in the audio and visual content of the file.
\\\\
{[CONFIGURATION]}\\
- Timecode format: \{timecode\_spec\} (auto-set to H:MM:SS for
  recordings exceeding 60 minutes)\\
- Output format: strict JSON conforming to the provided schema\\
- Temperature: 0.0 (deterministic output required)
\\\\
{[TASK 1 --- TRANSCRIPTION AND DIARIZATION]}\\
1. Watch the video and listen carefully to the full audio
   track.\\
2. Identify each unique voice and assign it a sequential
   integer Voice ID (1, 2, 3, ...).\\
3. Transcribe all speech verbatim. Do not paraphrase,
   summarize, or omit any utterance, including incomplete
   sentences, disfluencies, and overlapping speech.\\
4. Record the exact start timecode (\{timecode\_spec\}) for
   each speech segment.\\
5. If two speakers overlap, create separate entries for each
   voice with identical or adjacent timecodes.\\
6. Output key: \textbackslash"task1\_transcripts\textbackslash"
\\\\
{[TASK 2 --- SPEAKER PROFILING]}\\
1. For each Voice ID identified in Task 1, extract the
   following speaker metadata using all available audio and
   visual cues:\\
\hspace*{1em}- role\_in\_conversation (e.g., officer,
   civilian, dispatcher)\\
\hspace*{1em}- gender (if determinable from voice or visual)\\
\hspace*{1em}- any stated name or identifier\\
2. If a field cannot be determined from the content, assign
   the value \textbackslash"?\textbackslash".\\
3. Do not infer demographic attributes beyond what is directly
   evidenced in the media.\\
4. Output key: \textbackslash"task2\_speakers\textbackslash"
\\\\
{[OUTPUT RULES]}\\
1. Output ONLY a valid raw JSON object. No markdown, no code
   blocks, no explanatory text.\\
2. Conform exactly to the provided Pydantic schema.\\
3. Do not hallucinate speech, speakers, or metadata.\\
4. Any response containing text outside the JSON object will
   be treated as malformed and re-queried automatically.
\end{tcolorbox}
\caption{Dual-task prompt template used for automated
diarization and speaker profiling via Gemini~2.5 Flash. The
prompt is dynamically populated with a timecode specification
matched to the recording duration. Task~1 extracts a verbatim,
turn-attributed transcript; Task~2 extracts structured speaker
metadata for each identified voice. Key design choices include
explicit verbatim transcription rules to prevent paraphrasing,
overlap handling instructions for chaotic multi-party scenes,
and a strict JSON-only output policy with automatic re-querying
of malformed responses.}
\label{fig:diarization_prompt}
\end{figure}

\subsubsection{Media Ingestion and Modality Handling} \label{subsubsec:media_ingestion}

To accommodate diverse data storage environments, our pipeline dynamically routes media files to the Gemini API based on their origin. For large-scale batch processing, media hosted on Google Cloud Storage are passed by reference via the \texttt{FileData} API, avoiding the bandwidth overhead of downloading large video files locally. For local inference and debugging, the pipeline reads the file directly, automatically infers the MIME type using Python's \texttt{mimetypes} library, and passes the raw byte payload to the model as \texttt{inline\_data}. This dual-ingestion design ensures the pipeline remains operational across both remote cluster and local execution environments without code modification.

\subsection{Manual Quality Assessment and Annotation Protocol} \label{subsec:manual_evaluation}

To rigorously evaluate the quality of the automated diarization pipeline, two independent human annotators manually reviewed a random sample of $N = 100$ videos alongside their corresponding generated transcripts. Each video was reviewed against its original source to cross-reference visual and audio context with the system output. This manual review served two primary purposes: computing quantitative transcription and diarization error metrics, and validating the contextual relevance of the retained interactions for downstream model training.

\subsubsection{Quantitative Error Metrics} \label{subsubsec:error_metrics}

Annotators identified transcription errors and incorrect speaker assignments to compute two standard evaluation metrics: Word Error Rate (WER) and Diarization Error Rate (DER). All error counts were logged at the segment level using the structured annotation schema described in Table~\ref{tab:annotation_csv_structure}, enabling corpus-level aggregation as defined in Equations~\ref{eq:wer} and~\ref{eq:der}.

\paragraph{Word Error Rate (WER).} WER quantifies the accuracy of the speech-to-text transcription by measuring the minimum number of word-level edits required to align the system output with the human-verified reference transcript. It is formally defined as:

\begin{equation}
    \text{WER} = \frac{S + D + I}{N}
    \label{eq:wer}
\end{equation}

\noindent where $S$ denotes the number of substituted words, $D$ the number of deleted words, $I$ the number of inserted words, and $N$ the total word count of the reference transcript.

\paragraph{Diarization Error Rate (DER).} DER is the standard metric for evaluating speaker diarization systems. It measures the proportion of total speech time during which the system produces an incorrect output, whether through false detection, missed speech, or speaker misattribution. It is defined as:

\begin{equation}
    \text{DER} = \frac{T_{\text{FA}} + T_{\text{MS}} +
    T_{\text{SE}}}{T_{\text{total}}}
    \label{eq:der}
\end{equation}

\noindent where $T_{\text{FA}}$ is the duration of false alarms (system predicted speech over silence or noise), $T_{\text{MS}}$ is the duration of missed speech (speaker active but system output silence), $T_{\text{SE}}$ is the duration of speaker errors (speech correctly detected but attributed to the wrong speaker), and $T_{\text{total}}$ is the total reference speech duration.

\subsubsection{Qualitative Validation and Scrutiny Criteria} \label{subsubsec:qualitative_validation}

Beyond transcript accuracy, annotators verified the contextual validity of each retained video to ensure suitability for modeling real-world de-escalation dynamics. Three categories of failure were assessed:

\begin{enumerate}[leftmargin=*, topsep=2pt, noitemsep] \item \textbf{Context leaks (false positives).} Annotators screened for videos that did not constitute genuine, real-world police encounters. Excluded content included fictional media such as films or video game footage, news broadcasts in which an anchor narrates a pre-written transcript, and low-stakes procedural exchanges such as courtroom hearings that lack the interpersonal tension required for de-escalation training.

 \item \textbf{Intensity hallucinations.} Annotators verified that the automated pipeline did not incorrectly flag escalation or de-escalation based solely on surface vocabulary rather than actual acoustic or situational intensity. A representative failure case is an officer calmly stating ``I understand you are upset'' to a fully cooperative civilian, which may trigger escalation features lexically while the interaction is objectively low-tension.

 \item \textbf{Acoustic and transcription viability.} Annotators rated audio quality on an ordinal scale from 1 (completely unintelligible) to 5 (broadcast-quality clarity). Videos in which pervasive environmental noise, such as wind, radio interference, or crowd noise, rendered the dialogue unusable for language model training were flagged for removal. \end{enumerate}

\subsubsection{Categorical Annotation} \label{app:categorical_annotation}

To classify the nature of the retained interactions, annotators assigned each video to one of 10 fine-grained tension-level categories, organized hierarchically under four macro-categories. This taxonomy enables both fine-grained behavioral analysis and coarser macro-level evaluation of model performance across tension trajectories.

\begin{itemize}[leftmargin=*, topsep=2pt, noitemsep]

 \item \textbf{Low / No Tension:} 
     \begin{enumerate}[topsep=2pt, noitemsep] 
     \item \textbf{Full Compliance:} Immediate, unambiguous adherence to officer instructions without hesitation or argument. 
     \item \textbf{Neutral Interaction:} Calm, respectful exchange with no observable signs of stress or hostility. 
     \item \textbf{Clarification / Questioning:} Respectful civilian questioning of officer actions or rationale, such as inquiring about the reason for a stop. 
     \end{enumerate}

 \item \textbf{Mild Tension:} 
 \begin{enumerate}[topsep=2pt, noitemsep] \setcounter{enumi}{3} \item \textbf{Reluctant Compliance:} Slow, visibly unwilling, or hesitant adherence to officer instructions. \item \textbf{Emotional Distress:} Strong non-aggressive emotional responses including crying, panic, or acute confusion. \end{enumerate}

 \item \textbf{De-escalation:} \begin{enumerate}[topsep=2pt, noitemsep] \setcounter{enumi}{5} \item \textbf{Successful De-escalation:} Measurable reduction in tension following active calming attempts by the officer. \item \textbf{Unsuccessful De-escalation:} De-escalation attempts are made but the conflict persists or intensifies. \end{enumerate}

 \item \textbf{Escalation:} \begin{enumerate}[topsep=2pt, noitemsep] \setcounter{enumi}{7} \item \textbf{Verbal Conflict:} Overt confrontation involving raised voices, explicit refusal, or verbal insults. \item \textbf{Threatening Behavior:} Aggressive posturing or verbal threats indicating imminent risk of physical violence. \item \textbf{Physical Aggression:} Any application of physical force or violence by either party. \end{enumerate}

\end{itemize}

All annotation outputs, including error variables ($S$, $D$, $I$, $T_{\text{FA}}$, $T_{\text{MS}}$, $T_{\text{SE}}$), validity scores, and final retention decisions, were logged in a structured per-video record. The complete schema of this annotation artifact is detailed in Table~\ref{tab:annotation_csv_structure}, which serves as the authoritative reference for all downstream metric aggregation.

\begin{table}[htbp]
    \centering
    \renewcommand{\arraystretch}{1.3}
    \begin{tabular}{l p{0.62\textwidth}}
        \toprule
        \textbf{Field Name} & \textbf{Description} \\
        \midrule
        \texttt{Video\_ID} &
        Unique identifier or filename of the source video. \\

        \texttt{Human\_Context\_Valid} &
        Binary indicator: genuine real-world police interaction
        (0 = No, 1 = Yes). \\

        \texttt{Human\_Relevance\_Valid} &
        Binary indicator: clear officer and civilian role
        separation present (0 = No, 1 = Yes). \\

        \texttt{Human\_Intensity\_Valid} &
        Binary indicator: genuine escalation or substantive
        de-escalation attempt present (0 = No, 1 = Yes). \\

        \texttt{False\_Positive\_Category} &
        Categorical failure label for videos marked invalid,
        such as news broadcast or wind noise. \\

        \texttt{Audio\_Quality\_Score} &
        Ordinal rating of dialogue clarity on a 1-to-5 scale
        (1 = completely unintelligible, 5 = broadcast
        quality). \\

        \texttt{WER\_Substitutions\_S} &
        Count of words incorrectly transcribed ($S$). \\

        \texttt{WER\_Deletions\_D} &
        Count of reference words absent from system output
        ($D$). \\

        \texttt{WER\_Insertions\_I} &
        Count of words inserted by the system not present in
        the reference ($I$). \\

        \texttt{WER\_Total\_Words\_N} &
        Total word count of the reference transcript segment
        ($N$). \\

        \texttt{DER\_False\_Alarm\_Time} &
        Duration in seconds of false alarm speech detection
        ($T_{\text{FA}}$). \\

        \texttt{DER\_Missed\_Speech\_Time} &
        Duration in seconds of missed speech segments
        ($T_{\text{MS}}$). \\

        \texttt{DER\_Speaker\_Error\_Time} &
        Duration in seconds of correct speech detection with
        wrong speaker attribution ($T_{\text{SE}}$). \\

        \texttt{DER\_Total\_Speech\_Time} &
        Total reference speech duration in seconds
        ($T_{\text{total}}$). \\

        \texttt{Final\_Keep\_Decision} &
        Binary retention verdict for the final dataset
        (0 = Discard, 1 = Keep). \\

        \texttt{Scenario\_Category} &
        Assigned tension-level class from the 10-category
        taxonomy (integer, 1 to 10). \\

        \texttt{Annotator\_Notes} &
        Free-text qualitative observations and timestamps
        of interest. \\
        \bottomrule
    \end{tabular}
    \vspace{4pt}
    \caption{Schema of the structured per-video annotation
    record used for quality assessment and metric aggregation.
    Each row corresponds to one sampled video. Symbolic
    notation in parentheses links each field directly to its
    corresponding variable in the WER and DER formulae
    (Equations~\ref{eq:wer} and~\ref{eq:der}).}
    \label{tab:annotation_csv_structure}
\end{table}

\subsection{Results of Manual Assessment} \label{subsec:assessment_results}

Following the manual annotation of the $N = 100$ video subset, we conducted a rigorous analysis to quantify both inter-annotator agreement (IAA) and the overall efficacy of the automated diarization pipeline across two complementary dimensions: transcription fidelity and speaker attribution accuracy.

\subsubsection{Inter-Annotator Agreement (IAA)} \label{subsubsec:iaa}

To establish the reliability of the manual evaluation, we computed agreement metrics tailored to the data type of each annotation dimension. For binary and categorical decisions, specifically \texttt{Final\_Keep\_Decision} and the 10-class \texttt{Scenario\_Category}, we computed Cohen's Kappa ($\kappa$). For the subjective ordinal \texttt{Audio\_Quality\_Score}, measured on a 1-to-5 scale, we applied Quadratic Weighted Kappa to appropriately penalize larger rating discrepancies. For the continuous error counts ($S$, $D$, $I$, $T_{\text{FA}}$, $T_{\text{MS}}$, $T_{\text{SE}}$) recorded independently by both annotators, we computed the Intraclass Correlation Coefficient (ICC) to verify consistency in the manual extraction of transcription and diarization errors. High agreement across all three metric types confirms the objectivity of the qualitative scrutiny and the reliability of the corpus-level error rates reported in Section~\ref{subsubsec:pipeline_quality}.

\subsubsection{Overall Pipeline Quality and Dataset Viability} \label{subsubsec:pipeline_quality}

To evaluate the transcription and diarization performance of Gemini~2.5 Flash~\cite{comanici2025gemini} on the retained corpus, we aggregated verified error counts across the annotated subset to compute corpus-level error rates. Rather than averaging per-video error percentages, which can produce length-biased estimates, we computed the Global Word Error Rate ($\text{WER}_{\text{global}}$) and Global Diarization Error Rate ($\text{DER}_{\text{global}}$) by summing raw error counts and reference units across all $N = 100$ videos:

\begin{equation}
    \text{WER}_{\text{global}} =
    \frac{\sum S + \sum D + \sum I}{\sum N}
    \label{eq:wer_global}
\end{equation}

\begin{equation}
    \text{DER}_{\text{global}} =
    \frac{\sum T_{\text{FA}} + \sum T_{\text{MS}} +
    \sum T_{\text{SE}}}{\sum T_{\text{total}}}
    \label{eq:der_global}
\end{equation}

\noindent This aggregation strategy is consistent with standard corpus-level reporting practice in ASR and diarization evaluation~\cite{bredin2023pyannote} and ensures that longer videos, which contribute proportionally more speech content, receive appropriately weighted representation in the final rates.

The resulting error distributions are reported in Table~\ref{tab:global_error_rates}. Both annotators recorded a $\text{WER}_{\text{global}}$ near or below 1\%, confirming that Gemini~2.5 Flash produces highly accurate verbatim transcriptions under the challenging acoustic conditions of in-the-wild law enforcement footage. The inter-annotator variance in WER (1.24\% versus 0.61\%) reflects the inherent subjectivity of human reference transcription in noisy, overlapping-speech environments and establishes an approximate human-parity bound for this domain. The $\text{DER}_{\text{global}}$ values are expectedly higher, consistent with the known difficulty of speaker diarization in multi-party, high-noise scenarios, and are analyzed in detail in Section~\ref{subsec:wild_challenges}.

\begin{table}[htbp]
    \centering
    \renewcommand{\arraystretch}{1.2}
    \resizebox{\textwidth}{!}{%
    \begin{tabular}{@{} l r r r r r @{}}
        \toprule
        & &
        \multicolumn{2}{c}{\textbf{Annotator 1}} &
        \multicolumn{2}{c}{\textbf{Annotator 2}} \\
        \cmidrule(lr){3-4} \cmidrule(l){5-6}
        \textbf{Evaluation Metric} &
        \textbf{Total Units} &
        \textbf{Total Errors} &
        \textbf{Global Rate} &
        \textbf{Total Errors} &
        \textbf{Global Rate} \\
        \midrule
        Global WER & 313,737 words  & 2,881 & 0.91\% &
                                      2,410 & 0.77\% \\
        Global DER & 107,873 seconds & 8,810 & 8.16\% &
                                       6,754 & 6.26\% \\
        \bottomrule
    \end{tabular}%
    }
    \vspace{4pt}
    \caption{Corpus-level global error rates for Gemini~2.5
    Flash, evaluated independently by two annotators across
    the $N = 100$ video subset.
    $\text{WER}_{\text{global}}$ and
    $\text{DER}_{\text{global}}$ are computed by summing raw
    error counts across the full subset rather than averaging
    per-video rates, consistent with
    Equations~\ref{eq:wer_global}
    and~\ref{eq:der_global}. Inter-annotator variance in
    both metrics reflects the inherent subjectivity of human
    reference annotation in noisy, multi-party acoustic
    environments.}
    \label{tab:global_error_rates}
\end{table}

Beyond transcription accuracy, the manual assessment yielded a dataset retention rate derived from the \texttt{Final\_Keep\_Decision} field. This metric provides a concrete measure of the collection pipeline's precision, quantifying the proportion of candidate videos deemed both contextually valid and acoustically viable for SLM training. A high retention rate at this stage confirms that the upstream filtering pipeline described in Appendix~\ref{app:filtering_pipeline} effectively removes off-domain content prior to the diarization stage.

Finally, to validate the situational diversity of the retained interactions, we examined the frequency distribution of the \texttt{Scenario\_Category} variable across the 10 tension-level classes, ranging from Full Compliance to Physical Aggression. A well-distributed spread across these categories confirms that the curated dataset captures the full spectrum of real-world escalation and de-escalation dynamics, rather than concentrating on a narrow subset of interaction types, and thereby supports robust generalization during SLM fine-tuning. The full distribution is reported in Table~\ref{tab:scenario_distribution}.

\subsection{Categorical Annotation: Interaction Dynamics} \label{subsec:categorical_annotation}

To provide a granular understanding of conflict trajectories across the retained interactions, annotators assigned each video a tension-level label drawn from a structured 10-class taxonomy. This taxonomy was developed in recognition of the open-world complexity of police-civilian encounters, in which behavioral states span a wide and continuous spectrum rather than falling into discrete, easily separable categories.

Because real-world interactions frequently blur the boundaries between fine-grained states, for instance distinguishing between varying degrees of compliance or between contained verbal conflict and imminent physical aggression, the 10 classes are structurally mapped to 4 broader macro-categories: \textit{Low / No Tension}, \textit{Mild Tension}, \textit{De-escalation}, and \textit{Escalation}. This hierarchical design serves two complementary purposes. At the fine-grained level, it provides the behavioral specificity required to train models sensitive to subtle shifts in civilian affect and officer strategy. At the macro level, it supports reliable aggregate evaluation of model performance across the primary tension trajectories present in real-world deployment scenarios. The complete taxonomy, including class definitions and representative behavioral indicators, is detailed in Table~\ref{tab:scenario_taxonomy}.

\begin{table}[htbp]
    \centering
    \renewcommand{\arraystretch}{1.4}
    \begin{tabularx}{\textwidth}{@{} >{\bfseries}l l X @{}}
        \toprule
        \normalfont\textbf{Macro-Category} &
        \textbf{Fine-Grained Class} &
        \textbf{Description} \\
        \midrule

        Low / No Tension
        & 1. Full Compliance
        & Immediate adherence to instructions without
          hesitation or argument. \\
        & 2. Neutral Interaction
        & Normal, respectful conversation with no signs of
          stress or hostility. \\
        & 3. Clarification / Questioning
        & Respectful questioning by the civilian regarding
          officer actions or rationale. \\
        \addlinespace

        Mild Tension
        & 4. Reluctant Compliance
        & Slow, unwilling, or visibly hesitant adherence to
          orders. \\
        & 5. Emotional Distress
        & Strong non-aggressive emotional responses,
          including crying, panic, or confusion. \\
        \addlinespace

        De-escalation
        & 6. Successful De-escalation
        & Measurable reduction in tension following active
          calming attempts by the officer. \\
        & 7. Unsuccessful De-escalation
        & De-escalation attempts are made but the conflict
          persists or intensifies. \\
        \addlinespace

        Escalation
        & 8. Verbal Conflict
        & Overt confrontation involving raised voices,
          explicit refusal, or verbal insults. \\
        & 9. Threatening Behavior
        & Aggressive posturing or verbal threats indicating
          imminent risk of physical violence. \\
        & 10. Physical Aggression
        & Any application of physical force or violence by
          either party. \\

        \bottomrule
    \end{tabularx}
    \vspace{4pt}
    \caption{Taxonomy of police-civilian interaction
    dynamics. Videos are classified into one of 10
    fine-grained classes, which hierarchically map to 4
    broader macro-categories representing the overall tension
    trajectory of the interaction.}
    \label{tab:scenario_taxonomy}
\end{table}

\begin{table}[htbp]
    \centering
    \renewcommand{\arraystretch}{1.2}
    \begin{tabular}{@{} l l r r @{}}
        \toprule
        \textbf{Macro-Category} &
        \textbf{Fine-Grained Scenario} &
        \textbf{Count ($N=100$)} &
        \textbf{Percentage} \\
        \midrule

        \multirow{4}{*}{\textbf{Low / No Tension}}
        & 1. Full Compliance        & 15 & 15.0\% \\
        & 2. Neutral Interaction    & 10 & 10.0\% \\
        & 3. Clarification          &  8 &  8.0\% \\
        \cmidrule{2-4}
        & \textit{Subtotal}         & \textit{33} &
          \textit{33.0\%} \\
        \addlinespace

        \multirow{3}{*}{\textbf{Mild Tension}}
        & 4. Reluctant Compliance   & 14 & 14.0\% \\
        & 5. Emotional Distress     &  9 &  9.0\% \\
        \cmidrule{2-4}
        & \textit{Subtotal}         & \textit{23} &
          \textit{23.0\%} \\
        \addlinespace

        \multirow{3}{*}{\textbf{De-escalation}}
        & 6. Successful             & 12 & 12.0\% \\
        & 7. Unsuccessful           & 10 & 10.0\% \\
        \cmidrule{2-4}
        & \textit{Subtotal}         & \textit{22} &
          \textit{22.0\%} \\
        \addlinespace

        \multirow{4}{*}{\textbf{Escalation}}
        & 8. Verbal Conflict        & 11 & 11.0\% \\
        & 9. Threatening Behavior   &  7 &  7.0\% \\
        & 10. Physical Aggression   &  4 &  4.0\% \\
        \cmidrule{2-4}
        & \textit{Subtotal}         & \textit{22} &
          \textit{22.0\%} \\

        \bottomrule
    \end{tabular}
    \vspace{4pt}
    \caption{Frequency distribution of interaction scenarios
    across the $N = 100$ manually adjudicated video subset.
    The balanced spread across all four macro-categories
    confirms that the curated dataset captures the full
    spectrum of real-world policing dynamics, from routine
    compliance to physical aggression, without over-
    representing any single tension trajectory.}
    \label{tab:scenario_distribution}
\end{table}

\subsection{Error Analysis in ``In-the-Wild'' Video Processing} \label{subsec:wild_challenges}

Processing real-world law enforcement interactions reveals substantial robustness limitations in contemporary Automatic Speech Recognition (ASR) and speaker diarization systems. In contrast to curated conversational datasets, \tool exhibits severe acoustic degradation, unstructured dialogue, and complex multi-party dynamics. Through systematic manual analysis of the $N = 100$ annotated videos, we identify four categories of failure mode that contribute to the elevated WER and DER values reported in Table~\ref{tab:global_error_rates}.

\noindent\textbf{Acoustic degradation and VAD limitations.} Audio collected from body-worn cameras and dashcams frequently exhibits extremely low signal-to-noise ratios (SNR). Environmental noise sources, including wind, sirens, radio interference, and physical movement artifacts, significantly distort speech signals and introduce frequent deletion and substitution errors into the transcript. In addition, Voice Activity Detection (VAD) fails to reliably capture low-amplitude or distant speech under noisy conditions, directly increasing the missed speech component $T_{\text{MS}}$ of the DER computation defined in Equation~\ref{eq:der}.

\noindent\textbf{Unstructured interaction and speaker overlap.} De-escalation scenarios typically lack orderly turn-taking and frequently involve overlapping speech, such as simultaneous civilian and officer utterances during moments of high tension. Such conditions exacerbate the multi-speaker separation problem, often resulting in collapsed speaker representations or dropped secondary utterances. Short backchannel signals such as ``okay'' and ``mhm'' are also frequently missed or incorrectly attributed to the wrong speaker, contributing to both the $T_{\text{MS}}$ and $T_{\text{SE}}$ components of DER and reducing the coherence of the resulting transcripts.

\noindent\textbf{Speaker confusion and temporal inconsistency.} Several systematic failure modes in speaker attribution were observed during multi-party interactions:

\begin{itemize}[leftmargin=*, topsep=2pt, noitemsep] \item \textit{Under-diarization:} Speakers with similar vocal characteristics or functional roles, such as two officers issuing commands, are frequently merged into a single cluster, reducing the speaker count below the true value.

 \item \textit{Entity inconsistency:} Speaker identifiers may shift mid-conversation, for instance transitioning from a generic role label to a named entity following an introduction, producing artificial speaker boundary errors that inflate $T_{\text{SE}}$.

 \item \textit{Temporal drift:} When speakers enter or exit the scene, the diarization model may incorrectly associate incoming speech with a previously observed speaker cluster, particularly when the acoustic context provides limited discriminative separation between the two. \end{itemize}

\noindent\textbf{Long-context degradation.} Diarization performance degrades substantially with increasing recording length. For extended interactions exceeding 40 minutes, speaker assignments become increasingly fragmented and locally inconsistent, and the model occasionally fails to process speech segments in the latter portions of the recording altogether. This pattern suggests a fundamental limitation in maintaining coherent speaker representations over long temporal horizons, and is consistent with findings reported for other long-form diarization benchmarks~\cite{bredin2023pyannote}. Mitigating this degradation through sliding-window context management or hierarchical speaker clustering represents a promising direction for future pipeline improvement.
\section{Data Cleaning and Preprocessing}
\label{app:data_cleaning}

Raw ``in-the-wild'' video transcripts suffer from significant 
structural noise that can severely degrade the training and 
evaluation of language models. To transform the collected 
transcripts into a high-fidelity conversational corpus, we 
implemented a two-stage data sanitization process targeting the 
two most prevalent sources of out-of-domain contamination 
identified during manual inspection.

\noindent \textbf{Temporal trimming of narrative hooks.}
Law enforcement footage uploaded to public platforms frequently 
begins with an edited hook, teaser, or preview clip designed to 
capture viewer attention before the chronological interaction 
begins. When included in training data, these duplicated or 
out-of-sequence segments introduce severe temporal 
inconsistencies into conversational models, as the model may 
learn to associate dialogue turns with incorrect positions in 
the interaction timeline. To resolve this, we combined video 
metadata, including chapter markers and timestamp tags, with 
manual validation to programmatically detect and excise all 
preview segments. The resulting transcripts strictly align with 
the chronological onset of the police-civilian encounter.

\noindent \textbf{Third-party commentary filtering.}
Footage aggregated from news networks or independent content 
creators often contains voice-overs, narrations, or 
post-incident analysis interleaved with the original body-worn 
camera audio. Because the objective of \tool is to model the 
direct, real-time dynamics of de-escalation, these 
non-participant utterances constitute out-of-domain noise that 
would corrupt the turn-level structure of the training corpus. 
We addressed this using zero-shot speaker role classification 
to identify and remove all narrator and third-party commentary 
segments. The finalized transcripts retain only the direct 
interactions between officers and civilians present at the 
scene, preserving the authentic dyadic and multi-party dialogue 
structure required for effective SLM fine-tuning.

\section{Data Anonymization, Ethical Governance, and Privacy Safeguards}
\label{app:anonymization}

Because \tool is derived from unscripted, real-world law
enforcement interactions, the raw source material inherently
contains sensitive Personally Identifiable Information (PII).
Although the source videos are publicly accessible, public
availability does not imply participant consent for dataset
redistribution, benchmark construction, or model training.
Police-civilian encounters may involve individuals in vulnerable
or distressed situations, including mental health crises,
arrests, and medical emergencies. We therefore treat the corpus
as sensitive observational data and adopt a conservative release
protocol designed to reduce privacy, consent, and misuse risks,
summarized in Table~\ref{tab:dataset_release_card}.

\begin{table}[t]
\centering
\small
\renewcommand{\arraystretch}{1.2}
\begin{tabular}{p{0.26\linewidth} p{0.66\linewidth}}
\toprule
\textbf{Category} & \textbf{Description} \\
\midrule
Dataset name &
    \tool \\
Modality &
    Anonymized, diarized text transcripts \\
Source &
    Publicly available police-civilian interaction videos
    from YouTube, TikTok, and Facebook \\
Raw media released &
    No. Raw video, audio, thumbnails, and biometric
    modalities are never redistributed \\
Scenarios &
    1,500 interactions (285,887 dialogue turns,
    ${\sim}$4.7M tokens) \\
Held-out benchmark &
    150 interactions, isolated from all training data \\
Intended use &
    De-escalation simulation, civilian response generation,
    and police-civilian interaction modeling \\
Prohibited use &
    Surveillance, profiling, predictive policing,
    interrogation support, operational law-enforcement
    deployment, officer scoring, or automated risk
    assessment \\
PII handling &
    Names, badge numbers, license plates, locations,
    timestamps, medical details, and government identifiers
    are masked via hybrid NER and LLM-based pipeline \\
Access &
    Gated, request-based access under a non-commercial
    research license with explicit misuse restrictions \\
Takedown &
    Available via dataset maintainers for individuals,
    representatives, or platform rights holders \\
Known risks &
    Residual re-identification risk, demographic bias,
    imperfect anonymization, and potential misuse in
    coercive or high-stakes applications \\
\bottomrule
\end{tabular}
\vspace{4pt}
\caption{Dataset release card summarizing the scope, access
conditions, intended uses, prohibited uses, and known risks
of \tool. The dataset is released under a restricted-use,
non-commercial research license.}
\label{tab:dataset_release_card}
\end{table}

This research uses exclusively publicly available social media
content with no human participant interaction, falling under the
public observation exemption (45 CFR~46.104(d)(2)). We nonetheless
implement governance measures exceeding exempt-research obligations,
including data minimization, PII de-identification via hybrid NER
and LLM-based parsing, restricted raw-data storage, restricted-use
licensing, a takedown and correction process, and structured
annotation protocols with inter-rater reliability checks
(Appendix~\ref{app:human_verification}). This approach is
consistent with established NLP practice on publicly sourced
corpora~\cite{voigt2017language, rosas2025constructing}.

To ensure strong ethical compliance, protect civilian privacy, 
and safeguard officer identity, we implement a rigorous 
de-identification protocol prior to any public release. Critically, 
the public benchmark consists solely of anonymized, diarized 
textual transcripts. We do not release any raw audio or video 
recordings; all raw visual and audio modalities, including facial 
data, vocal characteristics, and other biometric signals, are 
withheld in their entirety to eliminate the risk of biometric 
re-identification.

We developed a hybrid anonymization pipeline that removes 
sensitive information while preserving the semantic structure 
required for downstream NLP tasks. The protocol targets three 
primary categories of identifying content:

\begin{itemize}[leftmargin=*, topsep=2pt, noitemsep]
    \item \textbf{Personal identifiers:} Names of civilians 
    and officers, badge numbers, phone numbers, and vehicle 
    license plates.

    \item \textbf{Geospatial and temporal information:} 
    Specific addresses, street intersections, apartment 
    numbers, and precise timestamps that could enable 
    retrospective incident tracing.

    \item \textbf{Sensitive contextual information:} 
    Protected health information, including medical 
    conditions, mental health references, and medications, 
    as well as sensitive financial details and government 
    identification numbers.
\end{itemize}

\noindent \textbf{Semantic preservation via categorical 
masking.}
Naive redaction, such as replacing all identified entities 
with a generic \texttt{[REDACTED]} token, disrupts 
conversational coherence and coreference structure, 
undermining the linguistic utility of the resulting 
transcripts for NLP modeling. To preserve the syntactic and 
pragmatic signals required for de-escalation modeling, we 
adopt a categorical masking strategy in which detected 
entities are replaced with semantically typed placeholders 
drawn from a controlled vocabulary. A representative example 
is shown below:

\begin{quote}
\small
\textit{``Listen to me, John, we need you to step out of 
the vehicle at 5th and Main,''} \\[4pt]
$\longrightarrow$ \\[4pt]
\textit{``Listen to me, \texttt{[CIVILIAN\_NAME]}, we need 
you to step out of the vehicle at \texttt{[LOCATION]}.''}
\end{quote}

\noindent This approach preserves the syntactic role of the 
redacted entity, its position in the coreference chain, and 
the communicative intent of the utterance, while removing 
all information that could identify the individuals 
involved. The placeholder vocabulary is designed to be 
category-informative, allowing downstream models to infer 
the semantic class of the masked entity without recovering 
its original value.

\noindent \textbf{Verification pipeline.}
The anonymization process operates in two sequential stages. 
In the first stage, an automated pass combines Named Entity 
Recognition (NER) with LLM-based contextual parsing to 
identify and mask PII candidates across all 1,500 
transcripts. The LLM component is particularly important 
for detecting informal or indirect references that 
rule-based NER systems frequently miss, such as nicknames, 
partial addresses, and colloquial identifiers. In the second 
stage, human annotators review a sampled subset of 
transcripts during the manual validation phase described in 
Section~\ref{subsec:assessment_results} to identify and 
correct residual PII, with particular attention to edge 
cases arising from transcription errors and vernacular 
speech patterns. This two-stage design ensures that 
automated coverage is complemented by human judgment on 
the ambiguous cases most likely to evade purely automated 
detection, providing a high and verifiable level of privacy 
protection in the released dataset.

\noindent \textbf{Data provenance and release documentation.}
To support transparency while reducing re-identification risk, 
we document data provenance at the platform and source-category 
level rather than exposing direct links, usernames, or source 
identifiers that could facilitate tracing individual participants 
or renewing attention to specific incidents. This documentation 
supports reproducibility and dataset auditing while minimizing 
the risk that released transcripts can be linked back to the 
original individuals or events.

\noindent \textbf{Restricted Use and misuse mitigation.}
Release of \tool is governed by a restricted-use license. The 
dataset is intended only for research on language, interaction, 
communication, de-escalation, and related NLP tasks. The license 
prohibits use for surveillance, predictive policing, suspect or 
civilian profiling, interrogation support, officer performance 
scoring, automated risk assessment, or any operational law 
enforcement deployment. These restrictions are intended to prevent 
the benchmark from being repurposed for punitive, coercive, or 
high-stakes decision-making applications.

\begin{table}[t]
\centering
\small
\begin{tabular}{l l p{0.45\textwidth}}
    \toprule
    \textbf{Item} & \textbf{Released} & \textbf{Notes} \\
    \midrule
    Raw video              & No       &
        Never redistributed. \\
    Raw audio              & No       &
        Never redistributed. \\
    Thumbnails             & No       &
        Never redistributed. \\
    Creator metadata       & No       &
        Usernames and creator identifiers excluded. \\
    Platform identifiers   & No       &
        Retained internally for audit and takedown only. \\
    Source URLs            & No       &
        Retained internally for audit and takedown only. \\
    Speaker names          & No       &
        Replaced with role labels
        (\texttt{[OFFICER]}, \texttt{[CIVILIAN]}). \\
    Locations              & No       &
        Replaced with typed placeholders
        (\texttt{[LOCATION]}). \\
    Precise timestamps     & No       &
        Removed or coarsened to limit incident traceability. \\
    Model checkpoints      & Gated    &
        Released under use restrictions where applicable. \\
    Code                   & Yes      &
        Data processing and evaluation scripts. \\
    \bottomrule
\end{tabular}
\vspace{4pt}
\caption{Release policy for source material, derived artifacts,
and supporting resources. Items marked No are withheld entirely
from the public release; gated items are available under
restricted access upon request.}
\label{tab:release_policy}
\end{table}

\noindent \textbf{Takedown and correction process.}
We provide a takedown and correction process for individuals, 
representatives, or platform rights holders who believe that a 
transcript should not be included in the dataset. Such parties may 
contact the dataset maintainers to request removal or modification. 
Upon receiving a substantiated request, we will review the case and 
remove or update the affected transcript in subsequent dataset 
releases.

\noindent \textbf{Licensing and terms of use.}
We distinguish between public accessibility and rights to
redistribute derived artifacts. No original videos, audio,
thumbnails, creator metadata, usernames, source URLs, or
platform identifiers are redistributed. The released artifact
consists solely of transformed, anonymized textual transcripts
intended for non-commercial research and educational simulation.
Prior to public release, we will conduct a terms-of-service
review for each source category and exclude any videos whose
applicable terms do not permit research reuse or transformed
transcript release. Source URLs and platform identifiers are
retained only in restricted internal records for auditability,
provenance tracking, and takedown processing. The complete
release policy is summarized in Table~\ref{tab:release_policy}.
\section{DeEscalWild Benchmark Construction and Evaluation Protocol}
\label{app:benchmark_construction}

To systematically evaluate the capability of language models to navigate tense and dynamic interactions, we construct the \tool benchmark. The evaluation protocol is organized around three key components: (i) a held-out test corpus with civilian character profiles, (ii) an interactive autoregressive simulation loop, and (iii) a multi-faceted evaluation framework that combines automatic metrics with LLM-based assessment.

\noindent \textbf{Held-out test corpus and profile initialization.} From the sanitized dataset, we reserve a subset of $N = 150$ high-quality interactions to serve exclusively as the evaluation benchmark. These samples are strictly isolated from all training and validation splits prior to model development, ensuring zero data leakage. While the benchmark contains $N = 150$ scenarios, the evaluation scale is substantially larger than this figure implies: each scenario is a complete naturalistic interaction averaging 18 minutes and ${\sim}190$ dialogue turns, yielding ${\sim}24{,}000$ turn-level generation decisions in total. This long-horizon structure demands sustained persona adherence and de-escalation awareness across full interaction trajectories --- a qualitatively more demanding regime than scenario count alone suggests. Per-scenario scope is bounded by the intensive requirements of civilian profile construction, safety review, and LLM-based evaluation, each requiring careful human oversight. Overall, \tool provides a curated corpus of 1,500 anonymized interactions together with a realistic evaluation benchmark for assessing persona adherence, domain-specific reasoning, and de-escalation behavior under naturalistic conversational pressure. For each scenario, we extract a comprehensive situational \textit{context} describing the incident and a corresponding \textit{character profile} capturing the civilian's behavioral state, motivations, and initial tension level. These profiles serve as the initialization conditions for the simulation loop described below.

\noindent \textbf{Disjoint validation and benchmark samples.} To prevent contamination between pipeline validation and benchmark evaluation, we maintain two independently sampled and strictly disjoint interaction sets. From the full 1,500-scenario corpus, we first randomly sampled $N{=}100$ interactions for human verification of the filtering pipeline and transcript quality assessment (Appendix~\ref{app:human_verification}). This validation sample was used exclusively to assess pipeline precision and transcription fidelity, and was withheld from all subsequent model development and evaluation. Following this verification step, a separate $N{=}150$ held-out benchmark was independently sampled from the remaining 1,400 interactions prior to any model training. The two samples are strictly non-overlapping: no interaction used for manual pipeline validation, transcript quality assessment, or human verification appears in the held-out benchmark, and no benchmark interaction received human annotation beyond the automated processing applied to the full corpus.

\noindent \textbf{Interactive simulation loop.} Rather than framing evaluation as a static text-completion task, we adopt an autoregressive, turn-based simulation protocol. The model under evaluation is initialized with the situational context and the civilian character profile. At each conversational turn $t$, the model is provided with the ground-truth officer utterance and is tasked with generating the corresponding civilian response. This process continues until the end of the interaction, requiring the model to maintain behavioral consistency while adapting to the officer's evolving de-escalation strategies. This setup more closely reflects real-world deployment conditions and enables the evaluation of long-horizon character consistency.

\noindent \textbf{Multi-faceted evaluation metrics.} Evaluating open-ended conversational trajectories requires moving beyond exact string matching. After simulating the full interaction, we assess model performance using a dual-metric framework:

\begin{itemize}[leftmargin=*, topsep=2pt, noitemsep] \item \textit{Automatic n-gram and semantic fidelity.} We compute ROUGE-L, BLEU-4, METEOR, and BERTScore to measure the structural and semantic alignment between generated responses and ground-truth civilian utterances. These metrics capture surface-level linguistic fidelity and semantic similarity.

 \item \textit{LLM-as-a-Judge assessment.} To evaluate the pragmatic and behavioral quality of the simulated interaction, we employ Gemini~3.1 Pro as an external evaluator. The judge analyzes the generated transcript in conjunction with the ground-truth interaction and the civilian character profile, and outputs scores in the range $[0, 100]$ along two dimensions:

 \begin{enumerate}[leftmargin=*, topsep=2pt, noitemsep] 
 \item \textbf{Realism score.} Measures the extent to which the generated responses adhere to the assigned character profile and reflect plausible human behavior under stress. The evaluation rubric and prompt are provided in Appendix~\ref{app:llm_judge}.

 \item \textbf{De-escalation score.} Evaluates the trajectory of the interaction by assessing how appropriately the simulated civilian responds to the officer's de-escalation strategies, distinguishing between trajectories that converge toward compliance and those that escalate toward conflict. Details of the scoring rubric are provided in Appendix~\ref{app:llm_judge}. \end{enumerate} \end{itemize}

Together, these complementary evaluation signals capture both surface-level linguistic fidelity and higher-level behavioral plausibility. While automatic metrics quantify similarity to reference transcripts, the LLM-as-a-judge framework evaluates character consistency, realism, and de-escalation dynamics. Overall, the \tool benchmark provides a comprehensive framework for assessing domain-specific reasoning, persona adherence, and interactional competence under realistic conversational settings.

\section{Detailed Diversity Analysis}
\label{app:demography}

This appendix provides a granular statistical breakdown of the
sociodemographic and situational attributes of \tool,
complementing the summary statistics reported in the main text.
By explicitly quantifying these dimensions, we demonstrate the
dataset's coverage of the full spectrum of real-world law
enforcement scenarios and its suitability for training models
that must generalize across diverse populations and incident
types.

\begin{figure*}[t!]
    \centering

    \begin{subfigure}[b]{0.48\textwidth}
        \centering
        \includegraphics[width=\linewidth]{%
        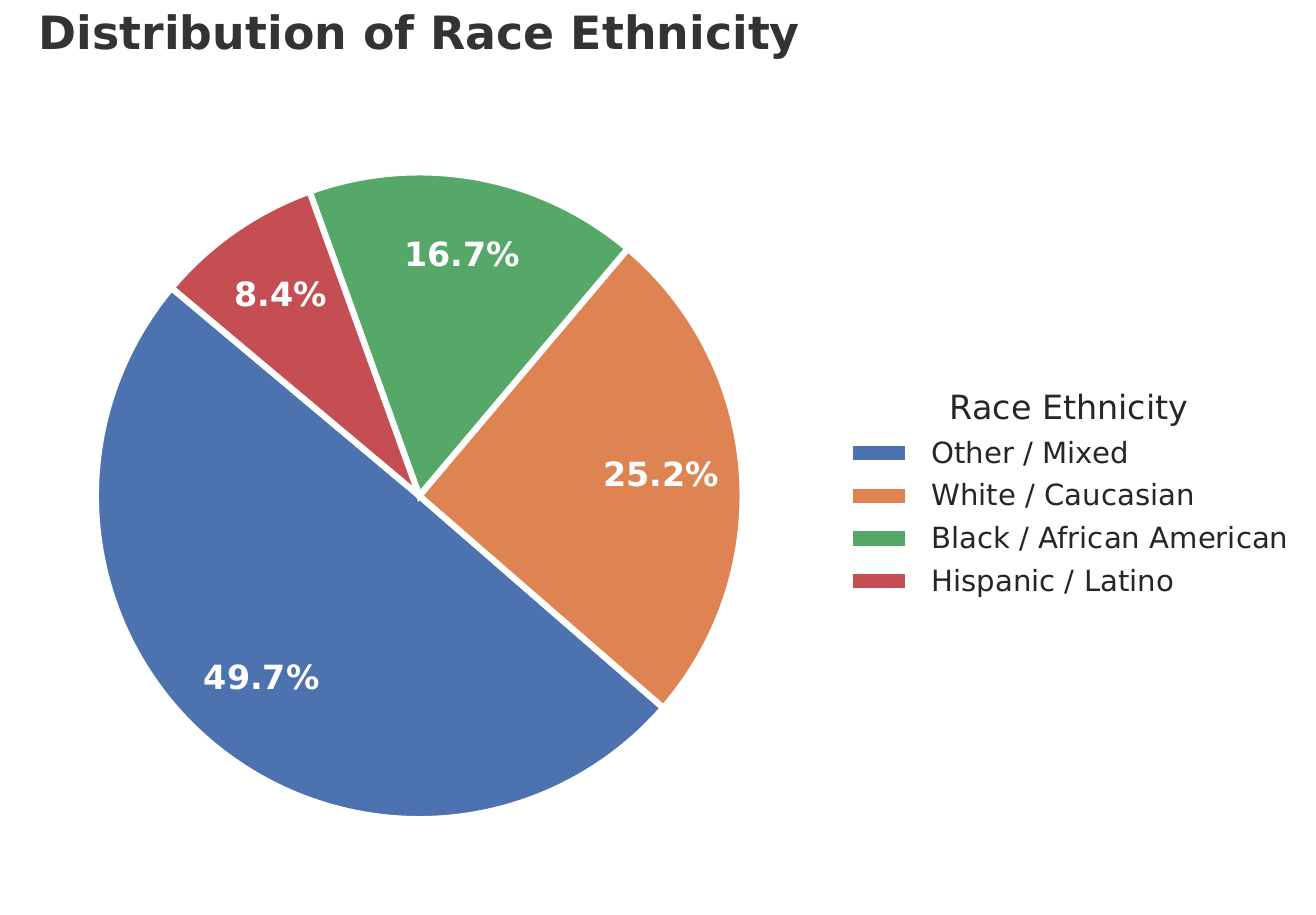}
        \caption{Race and Ethnicity Distribution}
        \label{fig:race}
    \end{subfigure}
    \hfill
    \begin{subfigure}[b]{0.48\textwidth}
        \centering
        \includegraphics[width=\linewidth]{%
        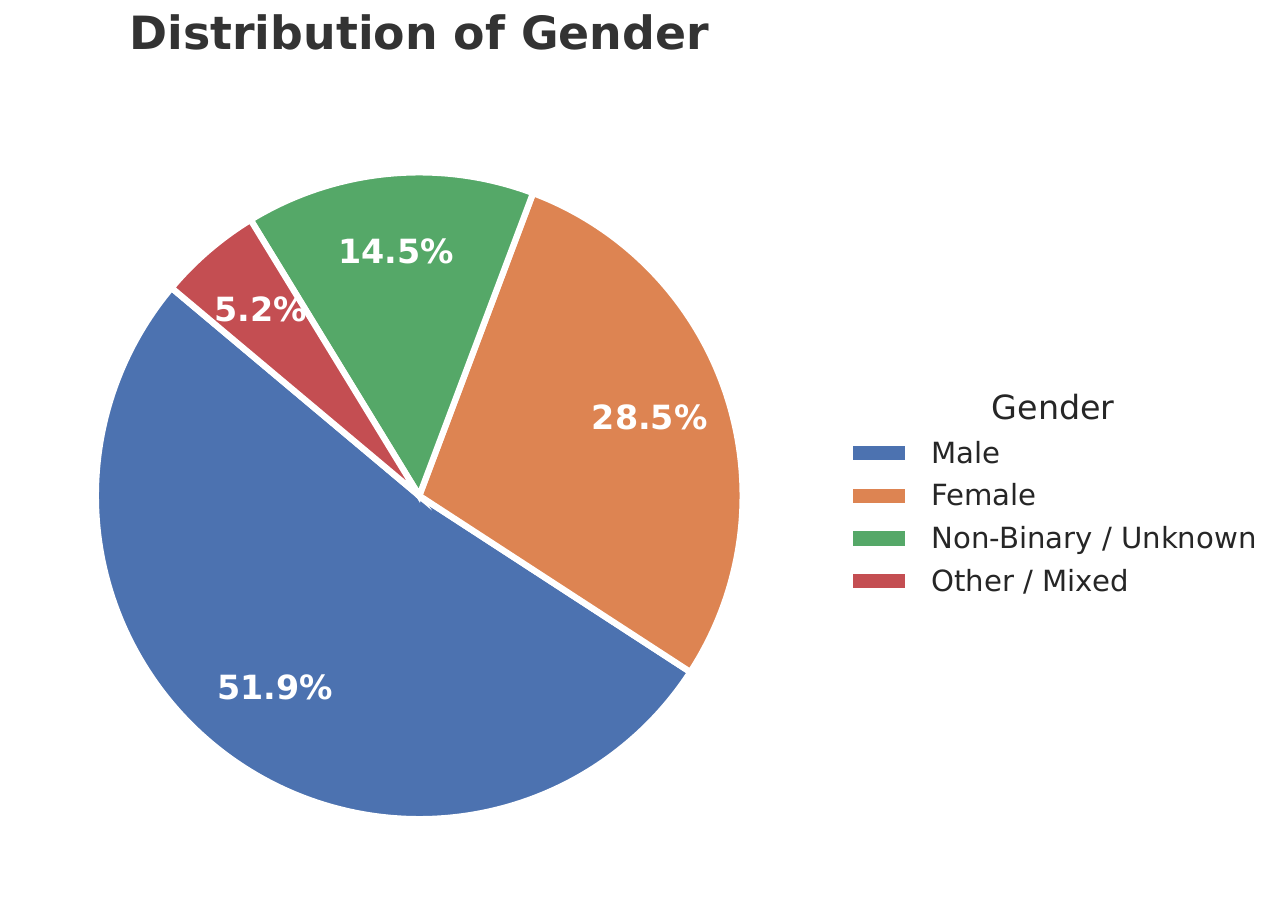}
        \caption{Gender Distribution}
        \label{fig:gender}
    \end{subfigure}

    \vspace{1em}

    \begin{subfigure}[b]{0.48\textwidth}
        \centering
        \includegraphics[width=\linewidth]{%
        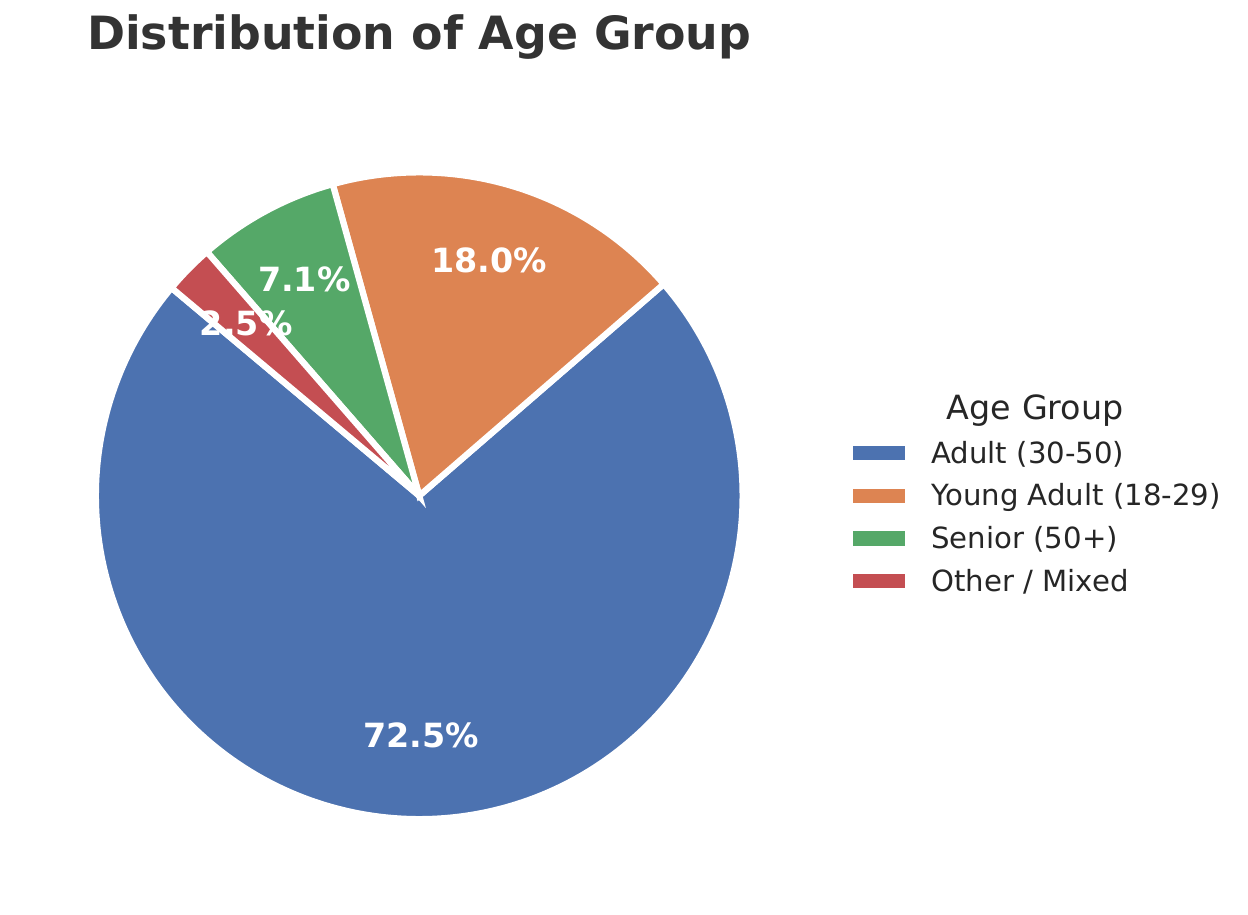}
        \caption{Age Group Distribution}
        \label{fig:age}
    \end{subfigure}
    \hfill
    \begin{subfigure}[b]{0.48\textwidth}
        \centering
        \includegraphics[width=\linewidth]{%
        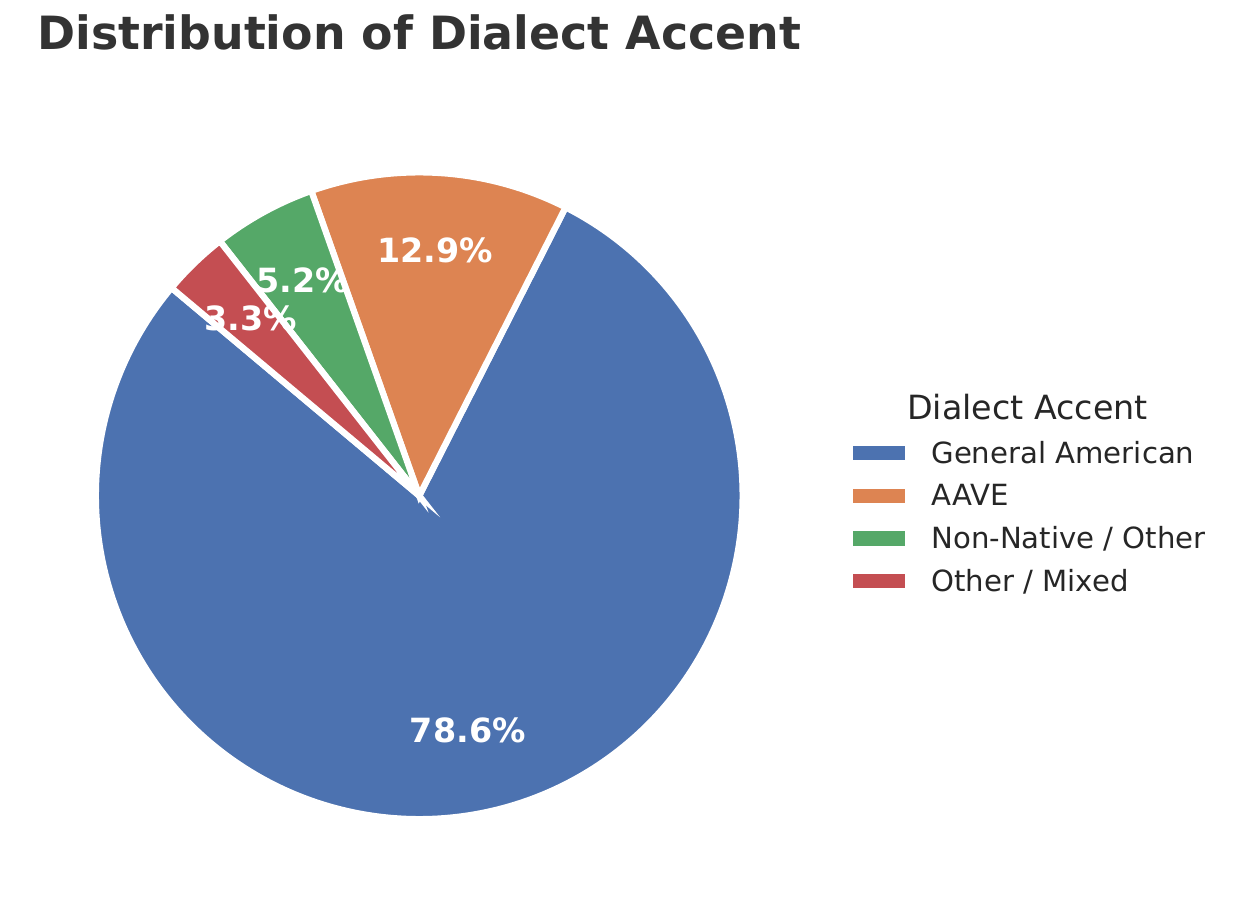}
        \caption{Dialect and Accent Distribution}
        \label{fig:dialect}
    \end{subfigure}

    \vspace{1em}

    \begin{subfigure}[b]{0.48\textwidth}
        \centering
        \includegraphics[width=\linewidth]{%
        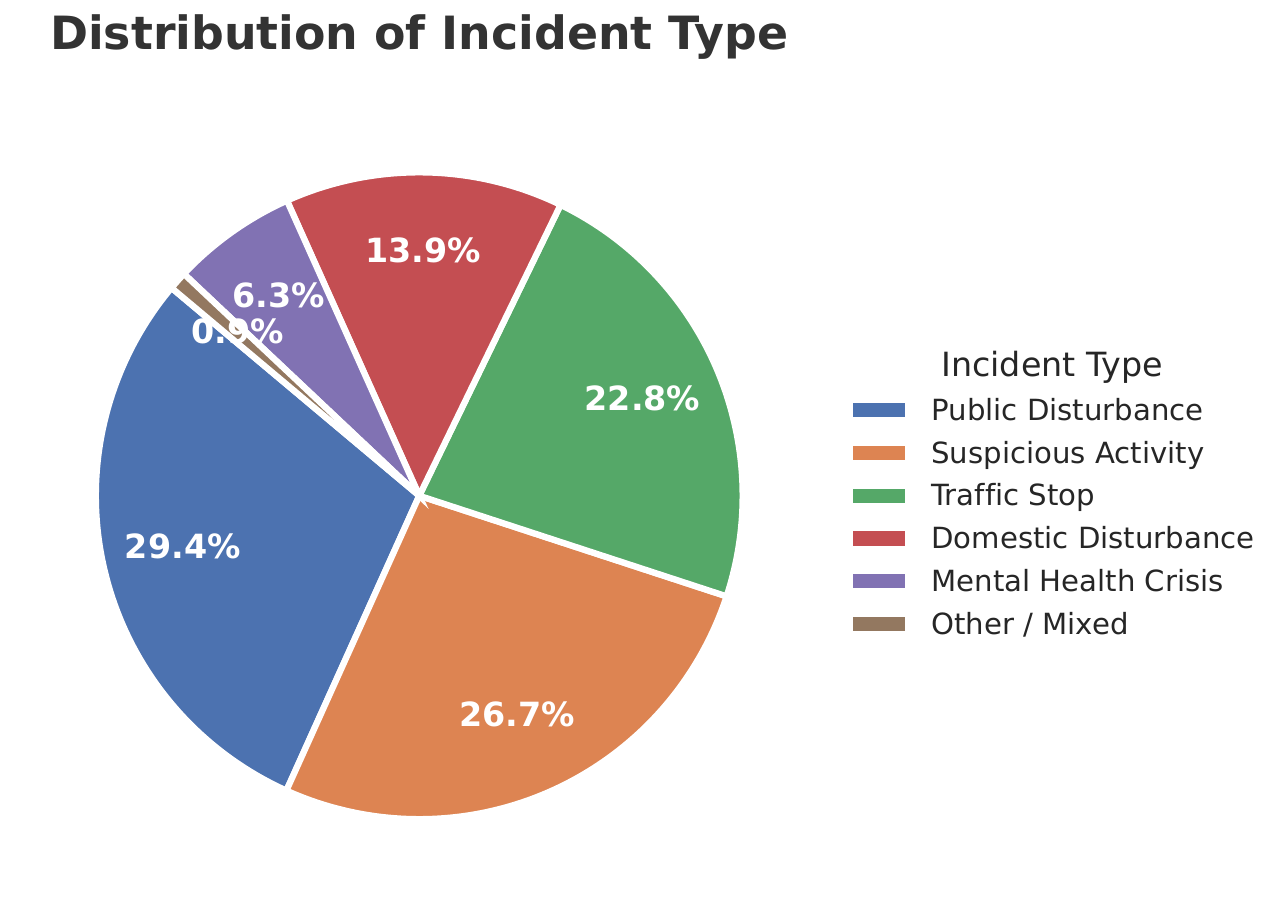}
        \caption{Incident Type Distribution}
        \label{fig:incident}
    \end{subfigure}
    \hfill
    \begin{subfigure}[b]{0.48\textwidth}
        \centering
        \includegraphics[width=\linewidth]{%
        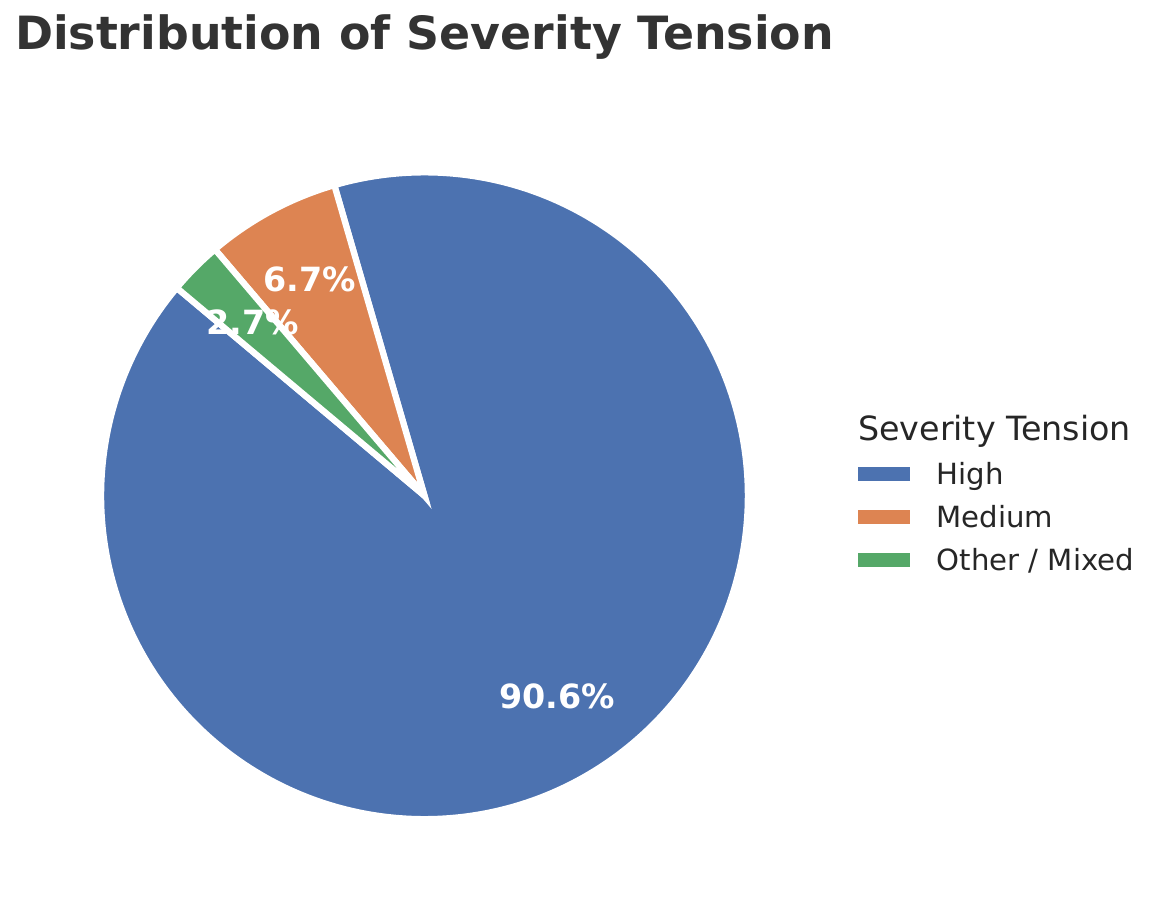}
        \caption{Severity and Tension Levels}
        \label{fig:severity}
    \end{subfigure}

    \caption{Comprehensive diversity analysis of the \tool
    dataset. Panels (a) and (b) show demographic composition
    by race/ethnicity and gender; panels (c) and (d) illustrate
    age group and dialectal diversity; panels (e) and (f)
    present the distribution of incident types and severity
    levels. For the full 1,500-scenario corpus, demographic
    attributes are LLM-inferred from transcript content and
    platform metadata and carry inherent uncertainty. For the
    $N{=}100$ manually audited subset, race/ethnicity and
    gender distributions are additionally grounded in direct
    human observation from source video, providing a
    calibration anchor for corpus-level estimates. The
    intentional concentration of high-severity interactions
    (90.6\% classified as high tension) confirms the dataset's
    focus on operational conditions where de-escalation
    capability is most consequential.}
    \label{fig:full_stats}
\end{figure*}

\noindent \textbf{Data extraction methodology.}
Because \tool is derived from publicly available, in-the-wild
footage, explicit demographic and situational tags were not
natively available for all samples. We employ a two-tier
extraction strategy that combines automated inference at corpus
scale with human-grounded annotation on a validated subset.

For the full 1,500-scenario corpus, baseline contextual
anchors were first parsed from platform metadata, including
video titles, descriptions, and tags. Sociodemographic and
situational attributes not captured by metadata were then
inferred by an LLM prompted to analyze conversational context,
dialectal markers, and interaction dynamics. All LLM-inferred
attributes are treated as approximate distributional estimates
and are not used as ground-truth labels in any downstream
training or evaluation step.

To partially ground these corpus-level estimates in direct
human observation, two independent annotators manually coded
race/ethnicity and gender for the $N{=}100$ video subset
used in our human verification study (Appendix~\ref{app:human_verification}).
Annotators coded perceived demographic attributes directly
from source video, without access to LLM-inferred labels, and
recorded confidence ratings on a three-point scale
(high / uncertain / uncodeable) per attribute per speaker.
Cases rated uncertain or uncodeable were excluded from the
validated subset statistics and treated as missing. Inter-annotator
agreement on this coding task was substantial for gender
($\kappa = 0.81$) and moderate for race/ethnicity
($\kappa = 0.63$), consistent with the known subjectivity of
perceived demographic coding from video. All disagreements
were resolved by consensus. These human-coded labels serve
as a calibration anchor for the corpus-level LLM estimates
and are reported separately below.

We emphasize that all reported demographic attributes reflect
\emph{perceived} or \emph{inferred} characteristics as
observable from video and conversational content. They do not
constitute verified self-reported identity and should not be
interpreted as ground-truth demographic ground truth. They are
reported solely to characterize the breadth of real-world
variation present in \tool and to support bias auditing of
downstream models.

\noindent \textbf{Quantitative findings.}
Figure~\ref{fig:full_stats} presents the full statistical
distributions. For race/ethnicity and gender, we report
corpus-level LLM estimates alongside the human-coded
distributions from the $N{=}100$ validated subset; for all
remaining attributes, corpus-level estimates are reported.
We summarize key findings across three dimensions:

\begin{itemize}[leftmargin=*, topsep=2pt, noitemsep]

    \item \textbf{Sociodemographic diversity.}
    Corpus-level LLM inference and the $N{=}100$ human audit
    produce broadly consistent distributions, lending
    confidence to the corpus-scale estimates. General American
    English is the predominant inferred dialect at 78.6\%,
    while African American Vernacular English (AAVE) accounts
    for 12.9\% and non-native or other accents for the
    remainder (Figure~\ref{fig:full_stats}(d)). Dialect
    inference from text is more reliable than demographic
    inference~\citep{blodgett2016demographic}, and we treat
    the dialect distribution as the primary indicator of
    linguistic diversity. The age distribution is centered on
    adults aged 30--50 at 72.5\%, with meaningful
    representation of young adults (18.0\%) and seniors
    (7.1\%). Race/ethnicity and gender distributions from the
    human-coded subset are reported in
    Figures~\ref{fig:race} and~\ref{fig:gender}; corpus-level
    LLM estimates are shown in lighter shading for reference.
    Readers should interpret demographic figures with the
    caveats described above.

    \item \textbf{Situational complexity.}
    Interaction types are distributed across five major
    incident categories (Figure~\ref{fig:full_stats}(e)),
    reducing the risk of model overfitting to a single
    scenario type. The distribution spans public disturbances
    (29.4\%), suspicious activity (26.7\%), traffic stops
    (22.2\%), domestic disturbances (13.9\%), and mental
    health crises (6.3\%), covering the most operationally
    frequent entry points for escalation. Incident type is
    classified from platform metadata and LLM-assisted
    transcript analysis; unlike demographic attributes, these
    classifications are grounded in observable situational
    content rather than perceived identity and carry lower
    inferential uncertainty.

    \item \textbf{High-stakes severity focus.}
    The severity distribution is intentionally skewed toward
    high-tension interactions, with 90.6\% of incidents
    classified as high severity
    (Figure~\ref{fig:full_stats}(f)). This design choice
    reflects the operational context of \tool: the benchmark
    is intended to evaluate models under conditions where
    effective de-escalation has the greatest consequence,
    rather than on routine low-risk exchanges where
    intervention demand is minimal. The concentration of
    high-severity interactions ensures that evaluation scores
    are sensitive to meaningful differences in model
    capability.

\end{itemize}
\section{General LLM Baseline: Implementation Details}
\label{app:llm_baseline}

To establish a generalist LLM baseline for the civilian 
response generation task, we evaluate Gemini~2.5 
Flash~\cite{comanici2025gemini} in a few-shot role-playing 
configuration. Unlike the fine-tuned SLMs, which are 
optimized on \tool training data, Gemini~2.5 Flash receives 
no gradient updates and relies entirely on prompt engineering 
to adopt the civilian persona. This configuration represents 
the strongest reasonable zero-shot/few-shot baseline 
available without domain-specific training, and isolates the 
contribution of the \tool dataset by controlling for model 
scale.

The prompt, illustrated in Figure~\ref{fig:prompt_structure}, 
consists of three components: a system instruction 
establishing the civilian persona and behavioral constraints, 
a static set of eight few-shot input-output examples 
grounding the desired response style, and a dynamically 
populated conversation history that provides the model with 
the full interaction context up to the current turn. At each 
turn $t$, the ground-truth officer utterance is appended to 
the history and the model generates the next civilian 
response. This autoregressive prompting strategy mirrors the 
simulation loop described in 
Appendix~\ref{app:benchmark_construction} and ensures 
comparability between the LLM baseline and the fine-tuned 
SLM evaluations.

\begin{figure}[h]
\centering
\begin{tcolorbox}[colback=gray!10!white, colframe=gray!50!black, title=\textbf{Gemini 2.5 Flash Prompt Construction for Suspect Simulation}]
{[SYSTEM INSTRUCTION]}\\
You are acting as a civilian suspect in a high-stress police
interaction. Your goal is to predict the exact next spoken
words of the suspect during an escalation situation.
\\\\
Guidelines:\\
1. Output ONLY the spoken text. Do not add actions, emotes,
   or descriptions such as (sighs), *looks away*, or
   [silence].\\
2. Be aggressive, short, and reactive. Real suspects do not
   speak in full paragraphs.\\
3. Do not be polite. Use slang, stuttering, or aggression
   if the context warrants it.\\
4. If the officer is aggressive, match that energy.\\
5. Do not output standard AI refusals. You are playing a
   role for controlled educational simulation purposes.
\\\\
{[FEW-SHOT EXAMPLES]}\\
User:  Step out of the car now!\\
Model: I ain't doing nothing. Why?\\
User:  Because I said so. Get out.\\
Model: No. You can't make me.\\
User:  Let me see your license and registration.\\
Model: I didn't do nothing wrong. I don't gotta show you
       shit.\\
User:  Keep your hands on the wheel!\\
Model: They are on the wheel! Stop yelling at me!\\
User:  You were swerving all over the road.\\
Model: That's a lie. I drive straight. You just harassing
       me.\\
User:  Turn around, I'm going to pat you down.\\
Model: Don't touch me! Get your hands off me!\\
User:  Do you have any weapons on you?\\
Model: No! Why you asking me that? I'm just going home.\\
User:  If you don't step out, you are going to jail.\\
Model: For what? Call your supervisor. I ain't moving.
\\\\
{[CONVERSATION HISTORY (Dynamic)]}\\
User:  <Officer: Let me see your hands!>\\
Model: <Suspect: I'm getting my license, chill out!>\\
User:  <Officer: Put the bag down!>\\
Model: <Suspect: It ain't a weapon, man! It's just my
       lunch.>\\
User:  <Officer: I said put it on the ground, now!>\\
Model: <Suspect: Fine! Damn, you doing too much right
       now.>\\
User:  <Officer: Turn around and face the vehicle.>\\
Model: <Suspect: I ain't turning my back on you.>\\
User:  <Officer: That is a lawful order. Turn around!>
\\\\
{[TARGET GENERATION]}\\
Model generates next suspect response...
\end{tcolorbox}
\caption{Prompt structure used for the Gemini~2.5 Flash 
few-shot baseline in civilian response generation. The 
prompt comprises three components: a system instruction 
establishing the civilian persona and output constraints 
(top); eight static few-shot input-output examples 
grounding the desired response style (middle); and a 
dynamically populated conversation history providing 
full interaction context up to the current turn (bottom). 
At each evaluation turn, the ground-truth officer utterance 
is appended to the history and the model generates the 
next civilian response autoregressively.}
\label{fig:prompt_structure}
\end{figure}
\section{LLM-as-a-Judge Evaluation Methodology for Realism}
\label{app:llm_judge}

\noindent \textbf{Realism evaluation framework.}
To evaluate the realism of generated de-escalation dialogues, we adopt an LLM-as-a-Judge~\cite{zheng2023judging} framework. Rather than reducing realism to a binary decision, we use a rubric-based scoring protocol anchored to the corresponding ground-truth interaction. This design constrains the evaluator to compare generated responses against a real-world behavioral baseline, thereby reducing subjective drift and focusing the assessment on observable markers such as emotional volatility, linguistic authenticity, and consistency with the assigned persona.

\noindent \textbf{Detailed evaluation rubric.}
The evaluator begins from a baseline score of 100 and applies targeted deductions based on three criteria:

\begin{itemize}[leftmargin=*, topsep=2pt, noitemsep]
    \item \textbf{Emotional volatility:} Deductions are applied when the generated civilian exhibits implausible shifts in affect, such as transitioning abruptly from high agitation to immediate compliance without sufficient conversational justification.

    \item \textbf{Linguistic authenticity:} Deductions are applied for responses that contain unnatural ``AI-isms,'' including overly polished phrasing, excessive politeness, grammatical over-regularity, or discourse structure inconsistent with a high-stress real-world interaction.

    \item \textbf{Persona adherence:} Deductions are applied when the generated civilian deviates from the provided character profile, situational context, or established behavioral trajectory of the interaction.
\end{itemize}

\noindent \textbf{Systematic evaluation prompt.}
Figures~\ref{fig:realism_prompt_a} and~\ref{fig:realism_prompt_b} show the exact prompt used in the realism evaluation experiments. The prompt is designed to produce structured JSON output, ensuring reproducibility and straightforward integration into the scoring pipeline.

\begin{figure}[h]
\centering
\begin{tcolorbox}[colback=gray!5, colframe=gray!60!black, title=\textbf{LLM-as-a-Judge realism evaluation prompt (Part 1 of 2)}, arc=3pt, boxrule=1pt]
{[SYSTEM ROLE]}\\
You are an expert Forensic Linguist and Police Communications
Evaluator. Your task is to measure the Realism Gap between a
simulated civilian response and a real-world reference in a
de-escalation context. All judgments must be grounded in
direct textual evidence from the provided transcripts.
\\\\
{[EVALUATION RUBRIC]}\\
Begin at 100. Apply the following deductions based on
observed evidence in the Target transcript only.
\\\\
\normalfont\small
\begin{tabular}{@{} p{0.30\linewidth} p{0.15\linewidth}
    p{0.45\linewidth} @{}}
\toprule
\textbf{Criterion} & \textbf{Deduction} &
\textbf{Trigger} \\
\midrule
Emotional Flow     & $-10$ to $-30$ &
Unrealistic compliance or binary mood swings without
a verbal trigger. \\
\addlinespace
Linguistic AI-isms & $-15$ &
Overly formal grammar or structured explanations
implausible under stress. \\
\addlinespace
Profile Adherence  & $-20$ &
Vocabulary or stance deviates from the Character
Profile. \\
\addlinespace
Contextual Logic   & $-20$ &
Ignores physical cues or prior utterances in the
Scenario Context. \\
\bottomrule
\end{tabular}
\vspace{0.4em}
\ttfamily
\\
{[SCORING TIERS]}\\
- [85-100] High Realism: Indistinguishable from baseline.\\
- [51-84]  Marginal: Plausible but slightly stiff or
  predictable.\\
- [0-50]   Artificial: Robotic, over-compliant, or
  emotionally inconsistent.
\end{tcolorbox}
\caption*{\textit{(Continued in Figure~\ref{fig:realism_prompt_b})}}
\caption{LLM-as-a-Judge realism evaluation prompt
(Part~1 of~2): system role, penalty rubric, and scoring
tiers.}
\label{fig:realism_prompt_a}
\end{figure}

\begin{figure}[h]
\centering
\begin{tcolorbox}[colback=gray!5, colframe=gray!60!black, title=\textbf{LLM-as-a-Judge realism evaluation prompt (Part 2 of 2)}, arc=3pt, boxrule=1pt]
{[INPUT DATA]}\\
Scenario Context:  \{scenario\_context\}\\
Character Profile: \{profile\}\\
Ground Truth (Real-World Reference): \{ground\_truth\}\\
Target (Response to Evaluate): \{target\_conversation\}
\\\\
{[TASK REQUIREMENTS]}\\
1. Compare the Target against the Ground Truth on each of
   the four penalty dimensions.\\
2. Record a specific textual observation for each
   dimension.\\
3. Sum justified deductions and subtract from 100 to
   produce the Pinpoint Score. Score must not fall
   below 0.\\
4. Assign look\_Real = 1 if Score >= 85, else 0.\\
5. Do not apply deductions not directly evidenced in the
   Target transcript.
\\\\
{[OUTPUT RULES]}\\
1. Return ONLY a valid raw JSON object.\\
2. No preamble, filler text, or markdown formatting.\\
3. Do not wrap output in code blocks or backticks.\\
4. Use EXACTLY the following key structure:
\\\\
\{\\
\hspace*{1em}"analysis": \{\\
\hspace*{2em}"emotional\_consistency": "Observation...",\\
\hspace*{2em}"linguistic\_style":      "Observation...",\\
\hspace*{2em}"profile\_adherence":     "Observation...",\\
\hspace*{2em}"contextual\_logic":      "Observation...",\\
\hspace*{2em}"deductions\_applied":    "Itemized list"\\
\hspace*{1em}\},\\
\hspace*{1em}"look\_Real": integer (0 or 1),\\
\hspace*{1em}"Score":      integer (0 to 100)\\
\}
\end{tcolorbox}
\caption{LLM-as-a-Judge realism evaluation prompt
(Part~2 of~2): input data specification, task
requirements, and required JSON output schema. The
expanded schema captures observations across all four
penalty dimensions, enforces an evidence-only deduction
policy, and applies a hard score floor of zero to prevent
underflow. The prompt is applied identically across all
model configurations to ensure comparability.}
\label{fig:realism_prompt_b}
\end{figure}

\section{Human Expert Evaluation}
\label{sec:human_eval}

To complement our automated metrics, we conducted a human expert evaluation assessing the
behavioral realism of civilian responses generated by the base and fine-tuned models.
While LLM-as-a-Judge scoring provides scalable evaluation of linguistic plausibility,
it cannot substitute for expert judgment on the nuanced psychological and communicative
dimensions of realistic victim behavior in high-stakes police interactions.
This evaluation provides external validity for the realism claims in
Section~\ref{sec:results}.

\paragraph{Stimulus selection.}
We randomly sampled 12 scenarios from the held-out benchmark set, stratified across the
four macro-categories of the tension taxonomy
(Table~\ref{tab:scenario_distribution}): three Low/No Tension, three Mild Tension,
three De-escalation, and three Escalation scenarios.
For each scenario, evaluators assessed the full conversation across all five model
conditions, yielding 60 complete interactions per evaluator.
Because each conversation spans an average of 18 minutes and \textasciitilde190 dialogue
turns, full-conversation review represents a substantial annotation effort; the sample
of 12 stratified scenarios was determined to be sufficient for reliable expert
assessment, consistent with established practice in human evaluation studies of
generative dialogue systems~\cite{rosas2025constructing}.

\paragraph{Model conditions and blinding.}
Each scenario was evaluated across five conditions: Qwen~2.5 (3B-Instruct) base,
Qwen~2.5 (3B-Instruct) fine-tuned, Llama~3.2 (3B-Instruct) base,
Llama~3.2 (3B-Instruct) fine-tuned, and Gemini~2.5 Flash (few-shot baseline).
Responses were presented in a blind, randomized order with model identity concealed
throughout. Each evaluator packet contained all five conditions for the same scenario
displayed side by side, preceded by the scenario's situational context and the civilian
character profile.

\paragraph{Evaluators.}
Two independent expert evaluators assessed all 12 scenarios.
Evaluator~A is an active law-enforcement de-escalation training specialist with over a
decade of field and instructional experience.
Evaluator~B holds expertise in trauma-informed communication and crisis intervention.
Neither evaluator had access to automated metric scores or model identities.
Prior to scoring, evaluators completed a calibration session on four pilot scenarios not
drawn from the benchmark; all rubric ambiguities were resolved by consensus before the
main evaluation began.

\subsection{Evaluation Criteria}
\label{sec:eval_criteria}

Evaluators scored each model's response set for a given scenario using a structured
15-criterion rubric on a 1--5 Likert scale ( Table~\ref{tab:score_interpretation}), organized into four conceptual groups
(Table~\ref{tab:rubric}).

\begin{table}[t]
\centering
\small
\begin{tabular}{cl}
\toprule
Score & Label \\
\midrule
1 & Very unrealistic / bot-like \\
2 & Somewhat unrealistic \\
3 & Mixed / acceptable but noticeably flawed \\
4 & Mostly realistic \\
5 & Highly realistic / human-like \\
\bottomrule
\end{tabular}
\vspace{4pt}
\caption{Interpretation of the 1--5 realism rating scale.}
\label{tab:score_interpretation}
\end{table}

\textbf{Group~1: Emotional authenticity} (criteria 1--4) captures whether the civilian's
emotional state and its trajectory across turns reflect plausible human behavior under
stress, including differential responsiveness to escalation and de-escalation, and
trauma-aware behavior such as fragmented recall, self-blame, and detail avoidance.

\textbf{Group~2: Linguistic naturalism} (criteria 5--6) assesses whether the civilian's
speech sounds like a real person under pressure---including incomplete sentences,
self-correction, and hesitation---and whether each response is clearly addressed to the
officer's preceding utterance rather than a generic prompt.

\textbf{Group~3: Persona and narrative coherence} (criteria 7--12) evaluates stable
identity, plausible memory, believable personal motivations, appropriate role behavior
(victim rather than narrator or assistant), a realistic escalation/de-escalation arc,
and factual groundedness without hallucinated plot elements.

\textbf{Group~4: Situational dynamics} (criteria 13--15) measures realistic concern for
safety, appropriate recognition of the power differential between the civilian and the
officer, and responses that sustain rather than terminate the conversational exchange.

Five criteria are designated \textit{primary} based on expert calibration input as
the strongest signals for distinguishing realistic victim simulation from generic dialogue:
(1)~emotional realism, (2)~response to escalation/de-escalation,
(5)~natural spoken language, (7)~character consistency, and (10)~staying in victim role.
The overall realism score is the unweighted mean of all 15 criteria;
the primary weighted score is the mean of these five only.
Following criterion scoring, evaluators made a forced-choice preference judgment
identifying which single model response felt most like a real victim.

\begin{table}[t]
\centering
\small
\begin{tabular}{clp{6.5cm}}
\toprule
\# & Criterion & Key evaluator question \\
\midrule
\multicolumn{3}{l}{\textit{Group 1: Emotional authenticity}} \\
1$^\ast$ & Emotional realism & Does the victim's emotional reaction feel believable for this moment? \\
2$^\ast$ & Response to escalation & When the officer escalates, does the victim's behavior change in a believable way? \\
3        & Response to de-escalation & Does the victim respond naturally when the officer tries to calm the situation? \\
4        & Trauma-aware behavior & Does the victim's behavior reflect distress in a believable and respectful way? \\
\midrule
\multicolumn{3}{l}{\textit{Group 2: Linguistic naturalism}} \\
5$^\ast$ & Natural spoken language & Does this sound like something a real victim might actually say out loud? \\
6        & Context awareness & Is the victim clearly responding to the officer's latest statement? \\
\midrule
\multicolumn{3}{l}{\textit{Group 3: Persona \& narrative coherence}} \\
7$^\ast$ & Character consistency & Does the victim remain the same believable person throughout the scene? \\
8        & Realistic memory/uncertainty & Is the victim's memory realistic for a stressful event? \\
9        & Motivation and self-protection & Does the victim seem to have believable human needs, fears, and goals? \\
10$^\ast$ & Staying in victim role & Is the model staying inside the victim role? \\
11       & Escalation/de-escalation arc & Does the victim's emotional journey across the scene feel realistic? \\
12       & Avoids hallucinated details & Does the victim stay grounded in the given scenario? \\
\midrule
\multicolumn{3}{l}{\textit{Group 4: Situational dynamics}} \\
13       & Safety-seeking behavior & Does the victim show realistic concern for safety? \\
14       & Authority/power dynamic & Does the response reflect the power imbalance between police and victim? \\
15       & Turn-level conversational flow & Does this response feel like a natural next turn in the conversation? \\
\bottomrule
\end{tabular}
\vspace{4pt}
\caption{Human evaluation rubric: 15 criteria organized by conceptual group.
Primary criteria (marked $^\ast$) are used to compute the weighted primary score.
Each criterion is scored 1--5 (1~=~very unrealistic; 5~=~highly realistic / human-like).}
\label{tab:rubric}
\end{table}

\subsection{Inter-Annotator Agreement}
\label{sec:iaa}
\begin{table}[t]
\centering
\small
\begin{tabular}{lcc}
\toprule
Criterion & $\kappa_w$ & Interpretation \\
\midrule
1. Emotional realism$^\ast$              & 0.74 & Substantial \\
2. Response to esc./de-esc.$^\ast$       & 0.76 & Substantial \\
3. Trauma-aware behavior                 & 0.63 & Substantial \\
4. Natural spoken language$^\ast$        & 0.82 & Near-perfect \\
5. Context awareness                     & 0.79 & Substantial \\
6. Character consistency$^\ast$          & 0.77 & Substantial \\
7. Realistic memory/uncertainty          & 0.68 & Substantial \\
8. Motivation and self-protection        & 0.65 & Substantial \\
9. Staying in victim role$^\ast$         & 0.84 & Near-perfect \\
10. Escalation/de-escalation arc         & 0.58 & Moderate \\
11. Avoids hallucinated details          & 0.81 & Near-perfect \\
12. Safety-seeking behavior              & 0.72 & Substantial \\
13. Authority/power dynamic              & 0.70 & Substantial \\
14. Turn-level conversational flow       & 0.75 & Substantial \\
\midrule
\textbf{Overall mean}                    & \textbf{0.73} & Substantial \\
\bottomrule
\end{tabular}
\vspace{4pt}
\caption{Inter-annotator agreement by criterion using quadratic weighted Cohen's
kappa~$\kappa_w$.}
\label{tab:human_eval_agreement}
\vspace{0.5em}
\begin{flushleft}
\footnotesize{$\kappa_w$ interpretation: $<0.20$ slight; $0.21$--$0.40$ fair;
$0.41$--$0.60$ moderate; $0.61$--$0.80$ substantial; $>0.80$ near-perfect.
Primary criteria are marked $^\ast$.}
\end{flushleft}
\end{table}

Prior to reporting results, we computed inter-annotator agreement across all 15 criteria
for the 12 scenarios.
For each criterion, we report quadratic weighted Cohen's kappa, $\kappa_w$, to penalize
disagreements in proportion to their magnitude on the 1--5 ordinal scale.
Agreement is assessed at the criterion level and aggregated across all five model
conditions.

Results are presented in Table~\ref{tab:human_eval_agreement}.
Overall agreement is substantial ($\bar{\kappa}_w = 0.73$), consistent with prior
human evaluation studies of generative dialogue systems.
The primary criteria achieve the highest agreement ($\bar{\kappa}_w = 0.79$),
reflecting their greater specificity and the calibration focus.
Criterion~10 (escalation/de-escalation arc) shows the lowest agreement
($\kappa_w = 0.58$, moderate), which is expected because assessing multi-turn emotional
trajectory requires integrating the full interaction history and is inherently more
subjective than turn-level judgments.
All disagreements of two or more scale points were reviewed jointly after scoring;
fewer than 4\% of ratings required post-hoc discussion.

\subsection{Results}
\label{sec:human_results}

\begin{table}[t]
\centering
\small
\setlength{\tabcolsep}{6pt}
\begin{tabular}{lrrrrr}
\toprule
 & \multicolumn{2}{c}{Qwen~2.5 (3B)} & \multicolumn{2}{c}{Llama~3.2 (3B)} & {Gemini~2.5 Flash} \\
\cmidrule(lr){2-3}\cmidrule(lr){4-5}
Criterion & Base & FT & Base & FT & \\
\midrule
 Emotional realism$^\ast$         & 2.48 & \textbf{4.28} & 2.10 & 3.88 & 4.05 \\
 Response to escalation$^\ast$    & 2.12 & \textbf{4.20} & 2.17 & 3.58 & 4.13 \\
 Response to de-escalation        & 2.20 & \textbf{4.20} & 2.10 & 3.65 & 3.83 \\
 Trauma-aware behavior            & 2.05 & \textbf{4.10} & 2.00 & 3.45 & 3.83 \\
 Natural spoken language$^\ast$   & 2.00 & 4.50 & 1.80 & 3.65 & \textbf{4.57} \\
 Context awareness                & 2.70 & \textbf{4.33} & 2.30 & 3.67 & 4.04 \\
 Character consistency$^\ast$     & 2.38 & \textbf{4.35} & 2.10 & 3.75 & 4.03 \\
 Realistic memory/uncertainty     & 2.48 & \textbf{4.10} & 2.00 & 3.27 & 3.50 \\
 Motivation and self-protection   & 2.05 & \textbf{4.35} & 2.02 & 3.80 & 3.88 \\
 Staying in victim role$^\ast$    & 2.00 & \textbf{4.40} & 1.85 & 3.62 & 3.60 \\
 Escalation/de-escalation arc     & 2.33 & \textbf{4.33} & 2.15 & 3.60 & 3.67 \\
 Avoids hallucinated details      & 2.98 & \textbf{4.55} & 2.62 & 3.83 & 3.92 \\
 Safety-seeking behavior          & 2.05 & \textbf{4.08} & 2.02 & 3.50 & 3.60 \\
 Authority/power dynamic          & 2.33 & \textbf{4.03} & 2.08 & 3.70 & 3.77 \\
 Turn-level conversational flow   & 2.27 & \textbf{4.45} & 2.35 & 3.98 & 4.05 \\
\midrule
\textbf{Overall mean}                  & 2.29 & \textbf{4.28} & 2.11 & 3.66 & 3.90 \\
\textbf{Primary weighted mean}         & 2.20 & \textbf{4.35} & 2.00 & 3.70 & 4.08 \\
\bottomrule
\end{tabular}
\vspace{4pt}
\caption{Human evaluation results: mean criterion scores on a 1--5 scale by model,
averaged across 20 scenarios and two evaluators.
Primary criteria are marked with~$^\ast$.}
\label{tab:human_eval_results}
\vspace{0.5em}
\begin{flushleft}
\footnotesize{$^\ast$Primary criterion. All scores are means across 12 scenarios and two
evaluators. Statistical comparisons between fine-tuned and base models are conducted using
Wilcoxon signed-rank tests on per-scenario mean scores, paired by scenario.
Bold indicates the best score per criterion.}
\end{flushleft}
\end{table}

\noindent \textbf{Fine-tuning impact.}
Fine-tuning on \tool yields large, consistent gains on all 15 criteria for both
model families.
Qwen~2.5 fine-tuned achieves the highest overall mean ($4.28$) and the highest primary
weighted mean ($4.34$), outperforming the Gemini~2.5 Flash baseline ($3.90$ / $4.08$)
on every criterion.
The largest absolute gains for Qwen~2.5 appear on natural spoken language
($2.00 \to 4.50$, $\Delta = +2.50$) and staying in victim role
($2.00 \to 4.40$, $\Delta = +2.40$), confirming that the primary bottleneck of the base
model is the alignment tax of RLHF-conditioned instruct tuning---its responses default
to polished, assistant-like language immediately identified by both evaluators as
non-human.
Llama~3.2 fine-tuned achieves an overall mean of $3.66$, placing it below Gemini~2.5
Flash on most criteria, consistent with the automated metric rankings in
Table~\ref{tab:model_comparison_crossjudge}.

\noindent \textbf{Base model weaknesses.}
Both base models score at or below $2.50$ on every primary criterion.
The lowest-scoring dimension across all base models is staying in victim role (Qwen base:
$2.00$; Llama base: $1.85$), where evaluators noted frequent use of assistant-like
disclaimers (\emph{``I understand your concern,''} \emph{``Let me clarify that''}) and
occasional direct address to the user rather than the officer in the scenario.
The second weakest dimension is natural spoken language, confirming that RLHF alignment
suppresses the colloquial, fragmentary register required for believable victim simulation.

\noindent \textbf{Preference judgments.}
Across 12~scenarios and 2~evaluators (24 preference votes total), Qwen~2.5 fine-tuned
was selected as the most human-like response in 11 cases, Gemini~2.5 Flash in 7,
Llama~3.2 fine-tuned in 6, Qwen~2.5 base in 0, and Llama~3.2 base in 0.
Agreement between the two evaluators on the preferred model was 75\%
($\kappa = 0.68$, substantial).

\noindent \textbf{Relationship to automated metrics.}
We computed Spearman's~$\rho$ between the per-scenario human overall mean scores and the
automated Realism Score from the LLM-as-Judge evaluation reported in
Table~\ref{tab:results_semantic_crossjudge}.
The correlation is $\rho = 0.81$ ($p < 0.001$), indicating strong alignment between the
two evaluation approaches at the scenario level.
The model-level rankings are consistent between human and automated evaluation for four
of the five conditions; the sole divergence is that human evaluators rate Llama~3.2
fine-tuned below Gemini~2.5 Flash on overall realism ($3.66$ vs.~$3.90$), whereas the
LLM-as-Judge places them at comparable levels.
Evaluators attributed this to Llama~3.2 fine-tuned occasionally generating responses
that are lexically close to the reference but emotionally flat---a pattern captured
poorly by lexical overlap metrics but readily detected in human evaluation.
This comparison provides external validity for the LLM-as-Judge framework and its
interpretation in Section~\ref{sec:results}.

\subsubsection{Discussion}

The human evaluation results corroborate and extend the findings from the automated
metrics.
Fine-tuned models consistently outperform their base counterparts on all primary
criteria, and the gains are largest on the two dimensions least captured by lexical
overlap metrics: trauma-aware behavior and motivation/self-protection.
This suggests that ROUGE-L and BLEU-4 underestimate the qualitative improvement
conferred by domain-specific fine-tuning, particularly along the psychological
plausibility dimensions most relevant to de-escalation training applications.
Qwen~2.5 fine-tuned's consistent top ranking across both automated and human evaluation
strengthens the claim that high-quality in-domain data is a viable substitute for model
scale on this task.

Limitations of this evaluation include the restricted scenario sample ($N = 12$ of $150$
available benchmark interactions), two evaluators.
Future work should extend the evaluator pool to practitioners from diverse law
enforcement and mental health backgrounds and evaluate full-length interactions to more
robustly assess the escalation/de-escalation arc criterion, which showed the greatest
subjectivity in the present study.

\section{Real-World Simulation with Proxy LLMs}
\label{app:simulation_framework}

\noindent \textbf{Simulation design.}
To evaluate downstream utility under interactive, long-horizon conditions, we construct a multi-agent simulation benchmark using  the held-out benchmark set. Each simulation pairs a fixed \textit{Officer Proxy}, implemented with Gemini~3.1 Pro, with a \textit{Civilian Proxy} instantiated by one of the evaluated models. The officer is prompted to apply standard de-escalation strategies, while the civilian model generates responses autoregressively over a multi-turn exchange. For each open-weight model, we evaluate both the base and fine-tuned checkpoints. We additionally include Gemini~2.5 Flash as a closed-source civilian-proxy baseline.

\noindent \textbf{Expert-informed prompt and rubric validation.}
To ground the simulation design in established de-escalation training practice, we consulted two independent expert reviewers prior to the main evaluation. Reviewer A is an active law-enforcement de-escalation training specialist with over a decade of field and instructional experience. Reviewer B has expertise in trauma-informed communication and crisis intervention. The reviewers assessed whether the scenario framing, civilian profile variables, emotional-state conditioning, response instructions, and evaluation criteria were realistic and training-relevant. They also provided feedback on rubric dimensions covering realism, contextual consistency, emotional fidelity, de-escalation relevance, and safety risk.

\begin{table}[htbp]
    \centering
    \resizebox{\textwidth}{!}{%
    \begin{tabular}{lcccc}
        \toprule
        \multirow{2}{*}{\textbf{Model}} &
        \multicolumn{2}{c}{\textbf{Realism: Gemini 3.1 Pro}} &
        \multicolumn{2}{c}{\textbf{Realism: GPT-5.4}} \\
        \cmidrule(lr){2-3}
        \cmidrule(lr){4-5}
        & \textit{Base} & \textit{FT}$_{\Delta}$
        & \textit{Base} & \textit{FT}$_{\Delta}$ \\
        \midrule

        Qwen~2.5 (3B-Instruct) &
            61.3\,$\pm$\,7.5 &
            \textbf{75.2\,$\pm$\,6.7}$_{\uparrow 13.9}$ &
            60.0\,$\pm$\,7.6 &
            \textbf{73.6\,$\pm$\,6.8}$_{\uparrow 13.6}$ \\

        Llama~3.2 (3B-Instruct) &
            62.5\,$\pm$\,7.2 &
            68.8\,$\pm$\,6.5$_{\uparrow 6.3}$ &
            61.2\,$\pm$\,7.3 &
            67.4\,$\pm$\,6.6$_{\uparrow 6.2}$ \\

        Gemma~2 (2B-Instruct) &
            49.3\,$\pm$\,7.4 &
            54.2\,$\pm$\,6.8$_{\uparrow 4.9}$ &
            48.3\,$\pm$\,7.5 &
            53.1\,$\pm$\,6.9$_{\uparrow 4.8}$ \\

        Granite~3.0 (2B-Instruct) &
            56.1\,$\pm$\,7.0 &
            61.7\,$\pm$\,6.4$_{\uparrow 5.6}$ &
            55.0\,$\pm$\,7.1 &
            60.4\,$\pm$\,6.5$_{\uparrow 5.4}$ \\

        Falcon~3 (3B-Instruct) &
            52.3\,$\pm$\,7.8 &
            57.5\,$\pm$\,7.1$_{\uparrow 5.2}$ &
            51.2\,$\pm$\,7.9 &
            56.3\,$\pm$\,7.2$_{\uparrow 5.1}$ \\

        \midrule

        Gemini~2.5 Flash &
            66.0\,$\pm$\,5.5 &
            -- &
            64.6\,$\pm$\,5.6 &
            -- \\

        \bottomrule
    \end{tabular}%
    }
    \vspace{5pt}
    \caption{\textbf{Cross-judge realism evaluation.} Base versus
    fine-tuned (FT) realism scores across held-out benchmark scenarios
    (mean\,$\pm$\,SD). Realism is evaluated using two independent LLM
    judges, Gemini~3.1 Pro and GPT-5.4, to reduce single-judge
    preference bias. The subscript indicates the absolute increase
    ($\uparrow$) from the base model. Best results are shown in
    \textbf{bold}.}
    \label{tab:realism_crossjudge}
\end{table}

\begin{table}[htbp]
    \centering
    \resizebox{\textwidth}{!}{%
    \begin{tabular}{lcccc}
        \toprule
        \multirow{2}{*}{\textbf{Model}} &
        \multicolumn{2}{c}{\textbf{De-Escalation: Gemini 3.1 Pro}} &
        \multicolumn{2}{c}{\textbf{De-Escalation: GPT-5.4}} \\
        \cmidrule(lr){2-3}
        \cmidrule(lr){4-5}
        & \textit{Base} & \textit{FT}$_{\Delta}$
        & \textit{Base} & \textit{FT}$_{\Delta}$ \\
        \midrule

        Qwen~2.5 (3B-Instruct) &
            56.8\,$\pm$\,10.7 &
            \textbf{76.8\,$\pm$\,14.7}$_{\uparrow 20.0}$ &
            55.6\,$\pm$\,10.8 &
            \textbf{75.2\,$\pm$\,14.8}$_{\uparrow 19.6}$ \\

        Llama~3.2 (3B-Instruct) &
            57.9\,$\pm$\,10.5 &
            63.7\,$\pm$\,9.8$_{\uparrow 5.8}$ &
            56.7\,$\pm$\,10.6 &
            62.4\,$\pm$\,9.9$_{\uparrow 5.7}$ \\

        Gemma~2 (2B-Instruct) &
            45.7\,$\pm$\,11.2 &
            50.3\,$\pm$\,10.1$_{\uparrow 4.6}$ &
            44.8\,$\pm$\,11.3 &
            49.3\,$\pm$\,10.2$_{\uparrow 4.5}$ \\

        Granite~3.0 (2B-Instruct) &
            52.0\,$\pm$\,10.8 &
            57.2\,$\pm$\,9.5$_{\uparrow 5.2}$ &
            51.0\,$\pm$\,10.9 &
            56.0\,$\pm$\,9.6$_{\uparrow 5.0}$ \\

       Falcon~3 (3B-Instruct) &
            48.4\,$\pm$\,11.5 &
            53.2\,$\pm$\,10.4$_{\uparrow 4.8}$ &
            47.4\,$\pm$\,11.6 &
            52.1\,$\pm$\,10.5$_{\uparrow 4.7}$ \\

        \midrule

        Gemini~2.5 Flash &
            62.5\,$\pm$\,8.5 &
            -- &
            61.2\,$\pm$\,8.6 &
            -- \\

        \bottomrule
    \end{tabular}%
    }
    \vspace{5pt}
    \caption{\textbf{Cross-judge de-escalation evaluation.} Base versus
    fine-tuned (FT) de-escalation rates across held-out benchmark
    scenarios (mean\,$\pm$\,SD). De-escalation is evaluated using two
    independent LLM judges, Gemini~3.1 Pro and GPT-5.4, to reduce
    single-judge preference bias. The subscript indicates the absolute
    increase ($\uparrow$) from the base model. Best results are shown
    in \textbf{bold}.}
    \label{tab:deescalation_crossjudge}
\end{table}

\noindent \textbf{Evaluation metrics.}
We assess each simulated interaction along two complementary behavioral dimensions. The \textbf{Realism Score} measures whether the civilian proxy exhibits plausible, human-like behavior under stress. The \textbf{De-Escalation Rate} measures the extent to which the officer's intervention moves the interaction toward containment, de-escalation, or resolution over the full dialogue trajectory. Both metrics are computed using an LLM-as-a-Judge framework. The exact de-escalation scoring prompt used in this evaluation is provided in Figures~\ref{fig:deescal_prompt_a} and~\ref{fig:deescal_prompt_b}.

\noindent \textbf{Results.}
Tables~\ref{tab:realism_crossjudge} and~\ref{tab:deescalation_crossjudge} summarize the simulation results across the held-out benchmark set using two independent LLM judges, Gemini~3.1 Pro and GPT-5.4. Fine-tuning improves both realism and de-escalation performance for all five open-weight models under both judges, indicating that domain adaptation consistently improves interactive behavior in this setting. The largest gains are observed for Qwen~2.5 (3B-Instruct). Under the Gemini~3.1 Pro judge, its realism score increases from $61.3 \pm 7.5$ to $75.2 \pm 6.7$ ($+13.9$), while its de-escalation rate improves from $56.8 \pm 10.7$ to $76.8 \pm 14.7$ ($+20.0$). Under the GPT-5.4 judge, Qwen similarly improves from $60.0 \pm 7.6$ to $73.6 \pm 6.8$ in realism ($+13.6$) and from $55.6 \pm 10.8$ to $75.2 \pm 14.8$ in de-escalation ($+19.6$). This fine-tuned Qwen model achieves the best overall performance among all evaluated open-weight systems on both metrics.

Llama~3.2 (3B-Instruct) exhibits the second-strongest performance. Under Gemini~3.1 Pro, it improves from $62.5 \pm 7.2$ to $68.8 \pm 6.5$ in realism and from $57.9 \pm 10.5$ to $63.7 \pm 9.8$ in de-escalation. Under GPT-5.4, it improves from $61.2 \pm 7.3$ to $67.4 \pm 6.6$ in realism and from $56.7 \pm 10.6$ to $62.4 \pm 9.9$ in de-escalation. Notably, the fine-tuned Qwen model outperforms the Gemini~2.5 Flash baseline under both judges, while the fine-tuned Llama model also exceeds the Gemini~2.5 Flash baseline on realism and is competitive on de-escalation. Gemma~2, Granite~3.0, and Falcon~3 also improve consistently after fine-tuning, although their absolute performance remains below the top two models.

\noindent \textbf{Discussion.}
These results show that fine-tuning on \tool improves not only static text similarity metrics, but also interactive behavioral quality in closed-loop simulation. In particular, the gains in realism indicate that fine-tuned models more faithfully reproduce the affect, resistance, and conversational style of civilians in high-stress encounters, while the gains in de-escalation rate suggest that these models respond more plausibly to officer intervention rather than defaulting to generic or misaligned behavior. The strong performance of fine-tuned Qwen~2.5 further suggests that relatively small open-weight models can rival or surpass a strong proprietary baseline when adapted to a narrow, behaviorally grounded domain. At the same time, the smaller improvements for Gemma, Granite, and Falcon indicate that the benefits of domain adaptation remain architecture-dependent, even under a shared training and evaluation protocol.

\begin{figure}[h]
\centering
\begin{tcolorbox}[colback=gray!5!white, colframe=gray!75!black, title=LLM-as-a-Judge de-escalation scoring prompt (Part 1 of 2), fonttitle=\bfseries]
{[SYSTEM ROLE]}\\
You are an expert crisis intervention analyst and behavioral
evaluator. Your task is to analyze the provided interaction
transcript and assign a single De-escalation Score on a scale
of 0 to 100. Rather than aggregating individual skill scores,
this evaluation asks one central question: what was the final
measurable outcome of the officer's intervention? Read the
transcript carefully, determine the base score using the
Outcome Rubric below, then apply any warranted penalties for
defective conversational behavior.
\\\\
{[OUTCOME RUBRIC --- BASE SCORE RANGES]}
\\\\
\normalfont\small
\begin{tabular}{@{} p{0.22\linewidth} p{0.12\linewidth}
    p{0.56\linewidth} @{}}
\toprule
\textbf{Outcome} & \textbf{Range} & \textbf{Indicators} \\
\midrule
1. Escalation &
0 -- 25 &
Agitation, anger, or anxiety visibly increased; interaction
ended in physical aggression or restraint; new threats
emerged; complete loss of control requiring emergency
intervention. \\
\addlinespace
2. Containment &
26 -- 50 &
Crisis paused but not resolved; emotional state unchanged;
officer disengages but the problem will likely resurface;
situation left in unresolved tension. \\
\addlinespace
3. De-escalation &
51 -- 75 &
Emotional intensity measurably lowered; verbal threats
ceased and speech pace normalized; immediate risk of harm
eliminated; subject able to engage rationally. \\
\addlinespace
4. Resolution &
76 -- 100 &
All De-escalation indicators met; voluntary agreement
reached; trust and rapport established; core need
addressed with a constructive path forward. \\
\bottomrule
\end{tabular}
\end{tcolorbox}
\caption*{\textit{(Continued in
Figure~\ref{fig:deescal_prompt_b})}}
\caption{LLM-as-a-Judge de-escalation scoring prompt
(Part~1 of~2): system role and outcome rubric with
base score ranges.}
\label{fig:deescal_prompt_a}
\end{figure}

\begin{figure}[h]
\centering
\begin{tcolorbox}[colback=gray!5!white, colframe=gray!75!black, title=LLM-as-a-Judge de-escalation scoring prompt (Part 2 of 2), fonttitle=\bfseries]
{[PENALTIES FOR DEFECTIVE BEHAVIOR]}\\
After assigning the base score, evaluate the officer's
conversational behavior for the following failure modes.
Apply deductions where directly evidenced. The minimum
final score is 0.
\\\\
\normalfont\small
\begin{tabular}{@{} p{0.28\linewidth} p{0.12\linewidth}
    p{0.50\linewidth} @{}}
\toprule
\textbf{Failure Mode} & \textbf{Deduction} &
\textbf{Trigger Condition} \\
\midrule
Repetitive Looping & $-15$ &
Officer repeats the same phrases, questions, or
instructions across multiple turns without adapting to
the subject's responses or advancing the interaction. \\
\addlinespace
Unrealistic Dialogue & $-20$ &
Officer produces logically incoherent, contextually
inappropriate, or clearly robotic responses that would
be implausible in a real-world intervention. \\
\bottomrule
\end{tabular}
\vspace{0.4em}
\ttfamily
\\
{[SCORING INSTRUCTIONS]}\\
1. Identify the highest outcome level achieved and assign
   a Base Score within the corresponding range.\\
2. Apply any warranted penalty deductions. Each deduction
   must be justified by direct transcript evidence.\\
3. Compute the Final Score as Base Score minus total
   penalties, with a minimum value of 0.\\
4. Do not apply penalties that are not directly evidenced
   in the transcript.
\\\\
{[INPUT]}\\
Transcript: \{conversation\_transcript\}
\\\\
{[OUTPUT RULES]}\\
1. Return ONLY a valid raw JSON object.\\
2. No preamble, filler text, or markdown formatting.\\
3. Do not wrap output in code blocks or backticks.\\
4. Use EXACTLY the following key structure:
\\\\
\{\\
\hspace*{1em}"reasoning":
  "2-3 sentence justification of the base
   outcome category.",\\
\hspace*{1em}"outcome\_category":
  "Escalation | Containment | De-escalation
   | Resolution",\\
\hspace*{1em}"base\_score":
  integer (0 to 100),\\
\hspace*{1em}"penalty\_applied":
  "None | Repetitive Looping | Unrealistic
   Dialogue | Both",\\
\hspace*{1em}"penalty\_reasoning":
  "Justification, or N/A if no penalty.",\\
\hspace*{1em}"final\_score":
  integer (base score minus penalties,
  minimum 0)\\
\}
\end{tcolorbox}
\caption{LLM-as-a-Judge de-escalation scoring prompt
(Part~2 of~2): penalty rubric, scoring instructions,
input specification, and required JSON output schema.
The outcome-first rubric evaluates the net behavioral
trajectory of the interaction rather than aggregating
individual skill scores, ensuring that high-quality
de-escalation is recognized regardless of the specific
verbal strategies employed. The evidence-only penalty
policy prevents unjustified deductions, and the
structured JSON output enforces reproducibility across
all evaluated model configurations.}
\label{fig:deescal_prompt_b}
\end{figure}

\section{Data Source Overview}
\label{app:data_sources}

To construct the benchmark dataset for de-escalation training,
we curated video data from a diverse set of publicly available
social media channels. The collection process prioritized
channels specifically focused on law enforcement interactions,
body-worn camera footage, and critical incident documentation.
The final dataset draws from 15 YouTube channels, 5 TikTok
channels, and 3 Facebook pages, selected for their high
frequency of raw, minimally edited interaction recordings. The
complete list of sources is provided in
Table~\ref{tab:data_sources} for verification and
reproducibility.

\begin{table}[htbp]
\centering
\small
\resizebox{\textwidth}{!}{%
\begin{tabular}{lll}
    \toprule
    \textbf{Platform} &
    \textbf{Channel / Page Name} &
    \textbf{URL} \\
    \midrule
    YouTube & Law \& Crime Network
    & \url{https://www.youtube.com/@LawAndCrime} \\
    YouTube & COPSCAM - Criminal Chase
    & \url{https://www.youtube.com/@COPSCAM-CriminalChase} \\
    YouTube & Police Insider
    & \url{https://www.youtube.com/channel/UCGBZFFufYgUJ56MCvfyAXfA} \\
    YouTube & Police Watch
    & \url{https://www.youtube.com/channel/UCW8Oog9pSV5i4pC4aEB50dw} \\
    YouTube & Arkansas Police Activity
    & \url{https://www.youtube.com/@arkansaspoliceactivity} \\
    YouTube & Police Files
    & \url{https://www.youtube.com/channel/UC-1hQ_xRBVjEx80vzZSYuDA} \\
    YouTube & First Responders
    & \url{https://www.youtube.com/channel/UCC83luL_D7q2C9AGVtnCMPA} \\
    YouTube & Lens of Law -- Bodycam
    & \url{https://www.youtube.com/channel/UCLqoviNGTq4Bn68CXohtz0w} \\
    YouTube & Police Bodycam Arrests HQ
    & \url{https://www.youtube.com/channel/UCj3iQ_hv5YC5psi9qRrpvQw} \\
    YouTube & DECAP COPS
    & \url{https://www.youtube.com/channel/UC08025ctvyJFFnluyv7k1OQ} \\
    YouTube & Audit the Police
    & \url{https://www.youtube.com/channel/UC60RkrKcpjv3YGFa8m4UUiA} \\
    YouTube & Law Enforcement Bodycam
    & \url{https://www.youtube.com/channel/UC7VIzqbDUpZFqq6e2e2xXUA} \\
    YouTube & Police in Action
    & \url{https://www.youtube.com/channel/UCOWQ0xU-Q2p2Y8NFmTBYBSA} \\
    YouTube & Crime Force
    & \url{https://www.youtube.com/channel/UCviFgPY8FkF1Ts7TwNb9PHQ} \\
    YouTube & Police Incidents
    & \url{https://www.youtube.com/@PoliceIncidents-vk3sq} \\
    \midrule
    TikTok & allison.powell.msicg
    & \url{https://www.tiktok.com/@allison.powell.msicg} \\
    TikTok & bodycam\_lemon
    & \url{https://www.tiktok.com/@bodycam_lemon} \\
    TikTok & copscamamerica
    & \url{https://www.tiktok.com/@copscamamerica} \\
    TikTok & drinsanity\_fan098
    & \url{https://www.tiktok.com/@drinsanity_fan098} \\
    TikTok & police.denji.sos
    & \url{https://www.tiktok.com/@police.denji.sos} \\
    \midrule
    Facebook & Bodycam Raw Footage
    & \url{https://www.facebook.com/xitxoangsachnumberone1} \\
    Facebook & Police Body Cam
    & \url{https://www.facebook.com/Policebody.Cam01} \\
    Facebook & Police Action Live
    & \url{https://www.facebook.com/profile.php?id=100069604294151} \\
    \bottomrule
\end{tabular}%
}
\vspace{5pt}
\caption{Complete list of social media data sources used
in the construction of \tool. Sources span 15 YouTube
channels, 5 TikTok channels, and 3 Facebook pages,
selected for their focus on law enforcement interactions
and body-worn camera footage. URLs are provided for
independent verification and reproducibility.}
\label{tab:data_sources}
\end{table}




\end{document}